\documentclass{article}

\PassOptionsToPackage{numbers, compress}{natbib}
\usepackage[main,final]{neurips_2026}

\usepackage[utf8]{inputenc}
\usepackage[T1]{fontenc}
\usepackage{hyperref}
\usepackage{url}
\usepackage{booktabs}
\usepackage{amsfonts}
\usepackage{nicefrac}
\usepackage[nopatch=showhyphens]{microtype}
\usepackage[table]{xcolor}
\usepackage{graphicx}
\usepackage{subcaption}
\usepackage{placeins}
\usepackage{wrapfig}
\usepackage{needspace}
\usepackage{longtable}
\usepackage{enumitem}
\usepackage{siunitx}
\usepackage{makecell}
\usepackage{caption}
\usepackage{multirow,colortbl}
\usepackage{algorithm}
\usepackage{algorithmic}
\usepackage{amsmath}
\usepackage{amssymb}
\usepackage{mathtools}
\usepackage{amsthm}
\usepackage[most]{tcolorbox}
\usepackage[textsize=tiny]{todonotes}
\usepackage[capitalize,noabbrev]{cleveref}

\definecolor{FigBlue}{HTML}{69B7EE}
\definecolor{FigBlueDark}{HTML}{337BAA}
\definecolor{FigYellow}{HTML}{FFE77D}
\definecolor{FigPurple}{HTML}{8B81BD}
\definecolor{FigPurpleDark}{HTML}{665989}
\definecolor{FigPeach}{HTML}{F8CBBA}
\definecolor{FigPeachDark}{HTML}{BB8068}
\definecolor{FigInk}{HTML}{385567}
\definecolor{FigGainGreen}{HTML}{45966C}

\newcommand{\best}[1]{\textbf{#1}}

\newcommand{\metricgain}[1]{\textsuperscript{\textcolor{FigGainGreen}{\scriptsize #1}}}

\hypersetup{
  colorlinks=true,
  linkcolor=blue,
  citecolor=blue,
  urlcolor=blue
}


\theoremstyle{plain}
\newtheorem{theorem}{Theorem}[section]
\newtheorem{proposition}[theorem]{Proposition}

\newtheorem{corollary}[theorem]{Corollary}
\theoremstyle{definition}
\newtheorem{definition}[theorem]{Definition}

\theoremstyle{remark}

\title{TabXLLM: Unlocking Zero- and Few-Shot Tabular Prediction in LLMs via Structural-Prior Reinforcement Learning}

\author{%
  Pengxiang Cai\quad Zihao Gao\quad Wanchen Lian\quad Guocong Li\quad
  Jintai Chen\thanks{Corresponding author: Jintai Chen
  (\href{mailto:jintaiCHEN@hkust-gz.edu.cn}{\texttt{jintaiCHEN@hkust-gz.edu.cn}}).} \\[2pt
]
  AI Thrust, Information Hub \\
  The Hong Kong University of Science and Technology (Guangzhou)
}

\begin{document}
\maketitle

\begin{abstract}
Tabular prediction has long been dominated by gradient-boosted decision trees and specialized deep tabular models, while large language models (LLMs) remain difficult to make competitive despite their cross-task adaptability and transparent reasoning traces. We address this gap by incorporating tabular structural priors into LLM post-training. Specifically, we propose Permutation Relative Policy Optimization (PRPO), which operationalizes column-permutation invariance through label-preserving column permutations and two-level advantage estimation. This design converts sparse outcome rewards into denser and more stable optimization signals. Using PRPO, we develop \textbf{TabXLLM}, an \textbf{8B} LLM for tabular prediction. Extensive experiments on 139 OpenML datasets show that TabXLLM reaches a genuinely competitive regime against strong specialized tabular baselines. It achieves strong fully supervised performance, dominates zero-shot settings, and performs on par with 32-shot strong baselines. Moreover, it substantially outperforms much larger general-purpose and reasoning LLMs, including up to a \textbf{53.17\%} improvement over DeepSeek-R1 (\textbf{685B}). These results show that structural-prior RL post-training is an effective route for making LLMs competitive in tabular prediction.

\par\noindent\textbf{Project:} \url{https://hkustgz-ml4health-lab.github.io/TabXLLM/}\\
\textbf{Code:} \url{https://github.com/HKUSTGZ-ML4Health-Lab/TabXLLM}\\
\textbf{Model:} \url{https://huggingface.co/conanca1/TabXLLM-8B}\\
\textbf{Datasets:} \url{https://huggingface.co/datasets/conanca1/TabXLLM}
\end{abstract}

\begin{figure*}[t]
    \centering
    \includegraphics[width=1\textwidth]{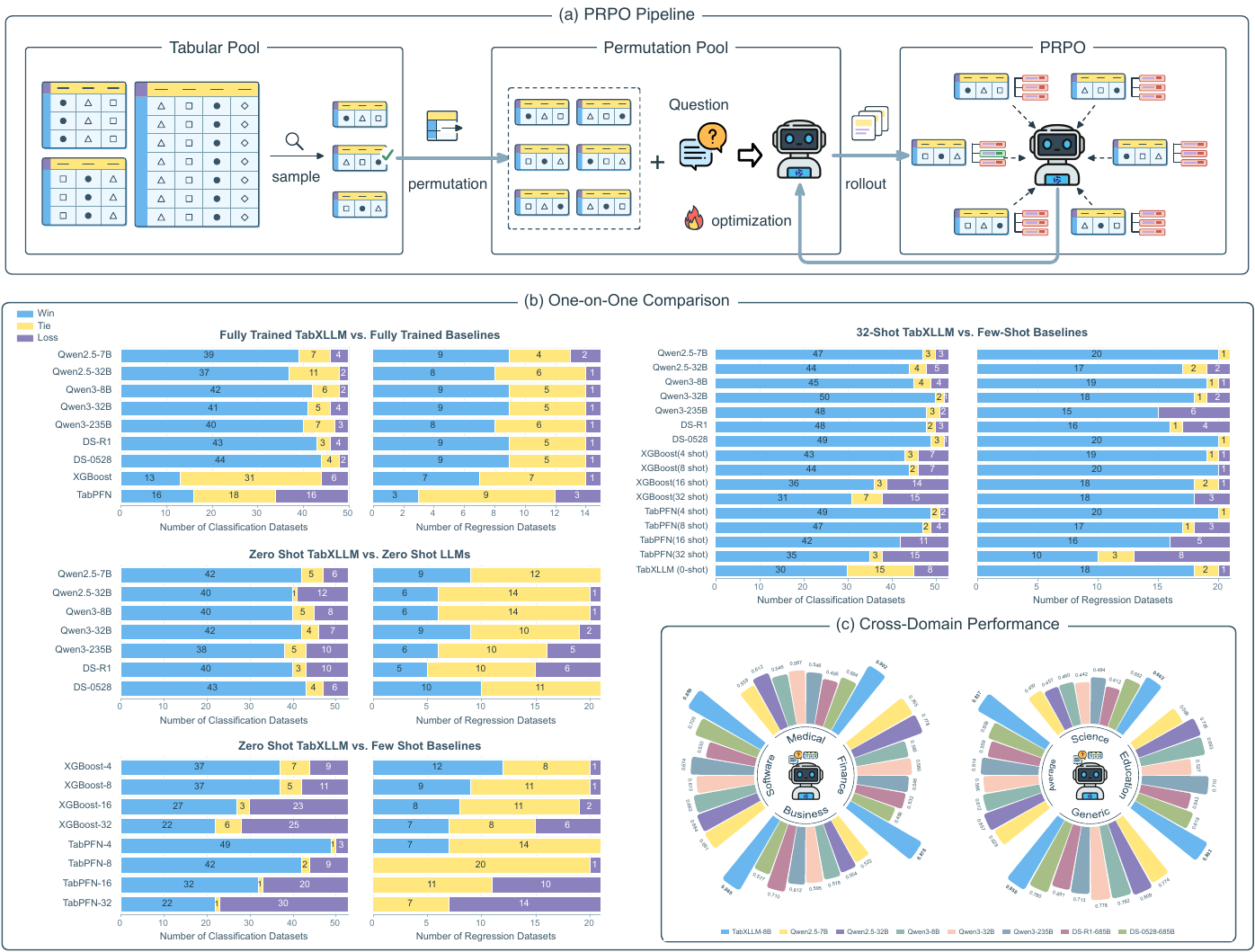}
    \caption{\textbf{(a)} We collect 139 datasets to construct a PRPO-compatible reinforcement learning dataset.
Each training sample is permuted into multiple variants, paired with a prediction question, and fed into TabXLLM.
The rollouts generate both intra-permutation and inter-permutation rewards, providing effective optimization signals for TabXLLM. \textbf{(b)} Win--Tie--Loss comparison between TabXLLM and other models under fully trained, few-shot, and zero-shot settings. \textbf{(c)} Cross-domain performance across seven domains, illustrating the gains brought by PRPO post-training over unadapted general-purpose LLMs on the full benchmark.}
    \label{fig:overview}
\end{figure*}

\section{Introduction}
Tabular prediction is a central task in machine learning with wide-ranging applications in healthcare, finance, and recommendation systems. Recent advances, including decision tree-based models~\citep{chen2016xgboost,prokhorenkova2018catboost} and deep learning models~\citep{hollmann2023tabpfn,hollmann2025accurate,chen2022danets,chen2024can,gorishniy2021revisiting}, have delivered strong performance on benchmark datasets. However, most of these approaches still require task-specific supervision, provide limited cross-task transfer, and offer limited interpretability~\citep{ke2017lightgbm, prokhorenkova2018catboost}, thereby impeding their deployment in real-world scenarios.

Large language models (LLMs) present a new paradigm for tabular prediction. Beyond producing predictive outcomes, LLMs can produce reasoning chains, thereby enhancing transparency and trust. Their pretraining on large-scale, multi-domain corpora also endows them with strong cross-task generalization ~\citep{brown2020language,wei2022chainofthought,wang2023selfconsistency}, enabling rapid adaptation to new tasks under few-shot and zero-shot conditions. However, this potential has not yet been fully realized for tabular data.
We argue that a key obstacle is the structure gap: standard language pretraining and generic instruction tuning do not directly encode the invariances and supervision patterns that tabular prediction relies on. Therefore, bringing tabular structural priors into LLM post-training is crucial for making LLMs competitive with specialized tabular predictors.

Recently, reinforcement learning (RL)~\citep{schulman2017proximal,ouyang2022training,bai2022training, rafailov2023direct,shao2024deepseekmath,deepseek-ai2025deepseekr1} has become a key approach for enhancing the reasoning ability of LLMs. Models such as DeepSeek-R1~\citep{shao2024deepseekmath,deepseek-ai2025deepseekr1} demonstrate that Group Relative Policy Optimization (GRPO) can substantially improve long-chain reasoning by leveraging group-relative advantage estimation. Yet, as widely observed when LLMs are extended beyond text (\textit{e.g.}, visual LLMs)~\citep{yao2025r1sharevl}, tabular LLMs struggle to replicate these significant gains due to the sparse-reward problem, stemming from the lack of process-level guidance and the scarcity of positive rewards during RL rollouts, as evidenced by DeepSeek-R1 (685B) achieving only about 57\% accuracy on binary classification tasks. Unable to extract effective guidance from such sparse feedback, tabular LLMs fail to capture intrinsic tabular structures, leading to inefficient exploration and severely limiting their few-shot generalization capabilities.

To address this issue, we propose \textbf{Permutation Relative Policy Optimization (PRPO)}, a reinforcement learning method specifically designed for tabular prediction that strengthens reward signals by injecting structural priors into LLM post-training. As shown in \Cref{fig:overview}(a), PRPO leverages the structural prior that tabular semantics remain invariant under column permutations. For each sample, it generates multiple column-permuted variants and serializes their feature names and values into prediction instructions for the LLM. Advantages are then estimated both within and across permutations, providing richer supervisory signals from the same training sample. This mechanism converts sparse outcome-level feedback into denser learning signals while preserving reward fidelity. By operationalizing permutation invariance in this manner, PRPO mitigates inefficient exploration, stabilizes optimization, and markedly improves the generalization ability of LLMs in tabular prediction. Building upon PRPO, we develop \textbf{TabXLLM}, an LLM-based tabular predictor that achieves strong fully supervised, zero-shot, and few-shot performance while retaining transparent prediction traces. \Cref{fig:overview}(b) and \Cref{fig:overview}(c) present the overall performance comparison between TabXLLM and multiple baselines, highlighting the superior capability of our approach. Our contributions are summarized as follows:

(i) We present the first systematic study of using tabular structural priors for LLM post-training, with the goal of strengthening LLMs for tabular prediction. We also construct a reinforcement learning dataset with verifiable rewards to support this post-training process, providing a useful data foundation for future LLM-for-tabular research.
    
(ii) We propose Permutation Relative Policy Optimization (PRPO), a novel reinforcement learning strategy that exploits column-permutation invariance to convert sparse outcome-level rewards into dense learning signals, thereby stabilizing training and improving generalization under limited supervision.
    
(iii) We validate that TabXLLM achieves strong performance under full supervision and competitive results in few-shot learning. In the zero-shot setting, it substantially outperforms other zero-shot methods, and its zero-shot performance already approaches the 32-shot performance of strong tabular baselines such as XGBoost and TabPFN-v2. Moreover, it substantially outperforms general-purpose LLMs that are up to two orders of magnitude larger while maintaining transparent reasoning traces.

\begin{figure*}[t]
    \centering
    \includegraphics[width=1\textwidth]{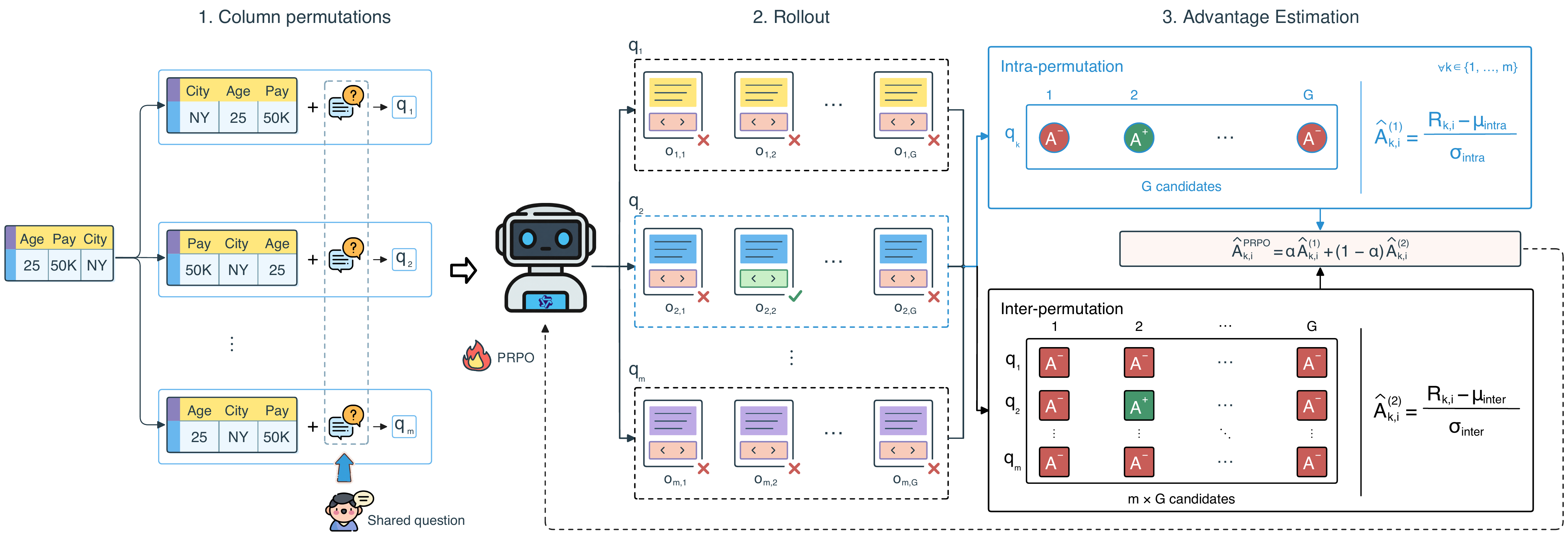}
    \caption{\textbf{Overview of PRPO.} Given a tabular sample, PRPO first generates multiple column-permuted variants that preserve the original label. Each permuted sample is then serialized, paired with the task question \smash{$Q$}, and passed to TabXLLM for rollout to produce candidate reasoning outputs \smash{$\{o_{i,j}\}$}. Rewards are computed via rule-based evaluation against verifiable ground-truth labels. Next, intra-permutation advantages \smash{$\{A_{i,j}^{(1)}\}$} are estimated within each permutation group, while inter-permutation advantages \smash{$\{A_{i,j}^{(2)}\}$} are aggregated across permutations. Finally, the two levels of advantages are integrated into \smash{$\{A_{i,j}\}$}, where \smash{$A_{i,j} = \alpha A_{i,j}^{(1)} + (1-\alpha) A_{i,j}^{(2)}$}, enabling permutation-aware reinforcement learning fine-tuning.}
    \label{fig:prpo overview}
\end{figure*}

\section{Method}
\label{method}
We strengthen LLMs for tabular prediction by incorporating tabular structural priors into LLM post-training. Previous LLM-based tabular prediction methods have mainly relied on supervised fine-tuning~\citep{hegselmann2023tabllm,gardner2024large} or prompt engineering~\citep{wei2022chainofthought, wang2023selfconsistency, brown2020language}. Although these methods bring certain performance improvements, they still struggle to make LLMs genuinely competitive with specialized tabular models.
In contrast, we propose Permutation Relative Policy Optimization (PRPO), which strengthens LLMs for tabular prediction through structural-prior RL post-training. The development of TabXLLM comprises two key stages: (1) \textbf{Tabular serialization}. LLMs cannot directly process structured tabular data, as they are primarily trained on unstructured text. To address this, we serialize tabular data into a natural language format suitable for LLM input. Based on this serialization, we constructed a reinforcement learning dataset, enabling reinforcement learning with verifiable rewards. (2) \textbf{PRPO fine-tuning}. PRPO encodes the column-permutation invariance of tabular semantics and thus transforms sparse rewards into dense learning signals.

\subsection{Tabular Serialization}
\label{subsec:tab ser}

We serialize tabular data into concise and consistent textual representations so that they can serve as effective inputs for our LLM.
We adopt a text-template-based~\citep{hegselmann2023tabllm} tabular serialization approach. Specifically, we define a function $serialize(F, x)$, where $F$ denotes the set of column names and $x$ represents the corresponding feature values. This function converts each feature--value pair in the table into a fixed-format natural language sentence such as \textit{``The \texttt{[feature]} is \texttt{[value]}.''} and concatenates all feature descriptions sequentially according to the column order, thereby producing a coherent textual representation of the entire row.
In addition to feature serialization, the LLM also receives a task-specific prediction query $Q$. When the serialized features $serialize(F, x)$ are combined with the task instruction $Q$, they form the complete LLM input $(serialize(F, x), Q)$, which guides the model's reasoning and prediction process. The detailed design and examples of the serialization template are provided in the Appendix~\ref{appendix:text template tabr1}. Utilizing this method, we constructed a reinforcement learning dataset tailored for tabular prediction tasks, which serves as the data foundation for our subsequent PRPO training.

\subsection{PRPO Fine-tuning}
\paragraph{Preliminary.} Recently, reinforcement learning (RL) has significantly advanced the long-chain logical reasoning capabilities of LLMs. In particular, the emergence of Group Relative Policy Optimization (GRPO) has established itself as a standard paradigm for LLM post-training~\citep{shao2024deepseekmath,deepseek-ai2025deepseekr1}. GRPO stabilizes reinforcement learning for LLMs by normalizing rewards within groups. 
Given an input $x$, the policy $\pi_\theta$ generates $G$ candidate outputs $\{o_1, \dots, o_G\}$, each assigned a scalar reward $\{R_1, \dots, R_G\}$. 
Relative advantages are computed as
\begin{equation}
\hat{A}_i = \frac{R_i - \mu_R}{\sigma_R}, \quad
\mu_R = \tfrac{1}{G}\sum_{j=1}^G R_j, \quad
\sigma_R = \sqrt{\tfrac{1}{G}\sum_{j=1}^G (R_j - \mu_R)^2}.
\end{equation}

The policy is updated using a PPO-style clipped objective with KL regularization:
\begin{equation}
L^{\text{GRPO}}(\theta) =
\mathbb{E}_{x,\,o_i \sim \pi_\theta}\!\Big[\min\!\big(r_i(\theta) \hat{A}_i, \text{clip}(r_i(\theta), 1{-}\epsilon, 1{+}\epsilon)\hat{A}_i \big)\Big] - \beta \, D_{\text{KL}}\!\big[\pi_\theta \| \pi_{\text{ref}}\big],
\end{equation}
where
\begin{equation}
r_i(\theta) = \frac{\pi_\theta(o_i|x)}{\pi_{\text{ref}}(o_i|x)}.
\end{equation}

While GRPO has achieved remarkable success in mathematical reasoning, it struggles to replicate these significant gains in tabular prediction. This stems from the absence of process-level guidance and the sparse reward problem, which provide limited effective feedback for policy learning. As shown in \Cref{fig:prpo_grpo_comparison}, GRPO suffers from sparse reward signals during the early stages of training, leading to slow convergence.

\paragraph{PRPO.} To address this issue, we propose \textbf{Permutation Relative Policy Optimization}, as illustrated in~\Cref{fig:prpo overview}, which exploits the column-order invariance of tabular data to construct a two-level advantage estimation and densify reward signals. 

Formally, a training tabular example with a default column order can be represented as
\begin{equation}
T = \{x_1, x_2, \dots, x_n, y\},    
\end{equation}
where $x_i$ denotes a feature together with its name and the value and $y$ denotes the label together with its name and the value.
We define a \textit{permutation} $\pi \in S$, which is a reordering of the feature index set $\{1,2,\dots,n\}$.  
When applied to input $T$, the feature-permuted result is expressed as
\begin{equation}
\pi(T) = \{x_{\pi(1)}, x_{\pi(2)}, \dots, x_{\pi(n)}, y\}.
\end{equation}
The set of all possible permutations forms the symmetric group $S$, with cardinality $|S| = n!$.  
In practice, we sample $m$ permutations from $S$ to construct a set of column-permuted variants:
\begin{equation}
\{t_1, t_2, \dots, t_m\}, \quad t_k = \pi_k(T), \quad \pi_k \in S.
\end{equation}
For any feature-permuted sample $t_k$, we first serialize it and combined with a task-specific question $Q$ (following the procedure in \Cref{subsec:tab ser}) to obtain a natural language sequence $q_k$, and then feed it to the policy model (denoted as $\pi_\theta$). The policy model subsequently generates $G$ candidate outputs $\{o_{k,1}, \dots, o_{k,G}\}$ by sampling from its distribution $o_{k,i} \sim \pi_\theta(\cdot|q_k)$.
From each output sequence $o_{k,i}$, we first extract a final answer $\hat{y}_{k,i}$ and then compute its rule-based reward $R(\hat{y}_{k,i})$. The reward function $R(\cdot)$ is defined as follows. For classification tasks, the reward for an answer $\hat{y}_{k,i}$ is:
\begin{equation}
\label{eq:reward_classification}
R(\hat{y}_{k,i}) = 
\begin{cases} 
    1.0 & \text{ if } \hat{y}_{k,i} = y^* \\ 
    0.1 & \text{ if } \hat{y}_{k,i} \neq y^*, \ \text{but correctly formatted}\\
    0.0 & \text{otherwise}
\end{cases}
\end{equation}
Where $y^*$ is the ground-truth value.

For regression tasks, where $y^*$ is the ground-truth value, the reward is determined by the relative error $e_{k,i} = |\hat{y}_{k,i} - y^*| / |y^*|$:
\begin{equation}
\label{eq:reward_regression}
R(\hat{y}_{k,i}) =
\begin{cases}
    1.0 & \text{if } \ \hat{y}_{k,i} \ \text{ is valid and } e_{k,i} \leq 0.1 \\
    0.5 & \text{if } \ \hat{y}_{k,i} \ \text{ is valid and } 0.1 < e_{k,i} \leq 0.5 \\
    0.1 & \text{if } \ \hat{y}_{k,i} \ \text{ is valid and } e_{k,i} > 0.5 \\
    0.0 & \text{otherwise}
\end{cases}
\end{equation}
On this basis, we further define a two-level advantage estimation: intra-permutation advantages and inter-permutation advantages.

\textbf{Intra-permutation advantages.} 
Intra-permutation advantages are computed within the permutation:
    \begin{equation}
    \hat{A}^{(1)}_{k,i} = \frac{R(\hat{y}_{k,i}) - \mu_k}{\sigma_k},
    \end{equation}
where $\mu_k = \tfrac{1}{G}\sum_{i=1}^G R(\hat{y}_{k,i})$ and $\sigma_k$ is the standard deviation.
    
\textbf{Inter-permutation advantages.}  
    All candidates across permutations are pooled into a single global group. Inter-permutation advantages are then computed as:
    \begin{equation}
    \hat{A}^{(2)}_{k,i} = \frac{R(\hat{y}_{k,i}) - \mu_{\text{global}}}{\sigma_{\text{global}}},
    \end{equation}
    where $\mu_{\text{global}}$ and $\sigma_{\text{global}}$ are computed over all $\{q_1, \dots, q_m\}$.

The final PRPO advantage integrates both levels:
\begin{equation}
\hat{A}^{\text{PRPO}}_{k,i} = \alpha \cdot \hat{A}^{(1)}_{k,i} + (1-\alpha) \cdot \hat{A}^{(2)}_{k,i},
\end{equation}
where $\alpha \in [0,1]$ balances local and global signals.

The final PRPO objective, which extends GRPO by incorporating the two-level advantages, is denoted by:
\begin{equation}
\resizebox{0.98\linewidth}{!}{$\displaystyle
L^{\text{PRPO}}(\theta) =
\mathbb{E}_{x}\Bigg[\sum_{k=1}^m \sum_{i=1}^G \min\Big(r_{k,i}(\theta)\hat{A}^{\text{PRPO}}_{k,i}, \text{clip}(r_{k,i}(\theta),1{-}\epsilon,1{+}\epsilon)\hat{A}^{\text{PRPO}}_{k,i}\Big)\Bigg] - \beta \, D_{\text{KL}}\!\big[\pi_\theta(\cdot|q_k) \| \pi_{\text{ref}}(\cdot|q_k)\big]
$}
\end{equation}
where
\begin{equation}
r_{k,i}(\theta) = \frac{\pi_\theta(o_{k,i}|q_k)}{\pi_{\text{ref}}(o_{k,i}|q_k)}.
\end{equation}

\subsection{Theoretical Analysis}
  \label{sec:theory}

  We analyze PRPO in Appendix~\ref{appendix:prpo_proof}. First, \Cref{prop:advantage_variance} characterizes the variance of PRPO's two-level advantage estimator. Second, \Cref{thm:reward_densification} and \Cref{prop:prpo_dominates} quantify correct-answer coverage under independent rollouts. Finally, \Cref{prop:clipped_prpo_gradient} gives the clipped surrogate's gradient, and \Cref{thm:convergence} gives a conditional stationarity bound with an explicit gradient-bias term.

\definecolor{GroupBlue}{HTML}{FFF6CE}
\definecolor{GroupPurple}{HTML}{E2E0EE}
\definecolor{GroupGreen}{HTML}{D2E9FA}
\definecolor{PRPOBlue}{HTML}{69B7EE}
\definecolor{GainBlue}{HTML}{337BAA}
\definecolor{GroupBlueText}{HTML}{AA8B28}
\definecolor{GroupPurpleText}{HTML}{665989}
\definecolor{GroupGreenText}{HTML}{337BAA}

\section{Experiments}
\label{experiment}
To comprehensively and systematically evaluate TabXLLM's predictive performance, we focus on two key aspects: (1) its performance under full supervision, and (2) its cross-dataset zero-shot transfer and few-shot generalization capabilities after structural-prior RL post-training.

\subsection{Datasets and Experimental Setup}
We collect 139 datasets from OpenML, including 103 classification datasets and 36 regression datasets. These datasets span a broad range of real-world domains, such as healthcare, finance, software engineering, business, education, and science. Following the tabular serialization procedure, we process each dataset and define a task-specific prediction question, constructing a reinforcement learning dataset with verifiable rewards for the training and evaluation of TabXLLM.

We comprehensively evaluate the overall performance of TabXLLM through a two-stage experimental setup. In the first stage, 50 classification datasets and 15 regression datasets are used for fully supervised training and testing. TabXLLM is fine-tuned using PRPO on the training sets and evaluated on the test sets to assess its performance under the fully supervised tabular prediction setting. In the second stage, we construct \textbf{ZeroTabBench} and \textbf{FewTabBench} from the remaining 53 classification and 21 regression datasets from OpenML to evaluate cross-dataset generalization. Neither benchmark uses target-task training or validation data for model development. ZeroTabBench evaluates prediction without target-task examples, whereas FewTabBench provides 32 labeled target-task observations only at inference time as in-context demonstrations, without updating model parameters.

TabXLLM is initialized from Qwen3-8B-Base and post-trained using PRPO. Our implementation is based on the Verl framework~\citep{sheng2025hybridflow}. All experiments are conducted on a node equipped with 8 NVIDIA H100-80G GPUs. A comprehensive list of hyperparameters and training configurations is provided in Appendix~\ref{appendixA}.

\subsection{Baselines}
We compare TabXLLM against three categories of baselines: \textbf{(1) General LLMs.} We evaluate several LLMs on ZeroTabBench and FewTabBench using raw zero-shot inference and 32-shot prompting, respectively, including models with comparable parameter scales to TabXLLM-8B, such as Qwen2.5-7B, Qwen2.5-32B, Qwen3-8B, and Qwen3-32B, as well as large-scale reasoning-oriented LLMs including DeepSeek-R1-0528-685B, DeepSeek-R1-685B, and Qwen3-235B. For these LLMs, we design text-template prompts suitable for zero-shot generation. Detailed configurations of the text-template prompts are provided in Appendix~\ref{appendix:text template qwen deepseek}. \textbf{(2) Tabular LLMs.} We evaluate Tabula-8B on ZeroTabBench and FewTabBench, and include TabLLM as a supervised few-shot comparison. \textbf{(3) Strong Tabular Baselines.} We evaluate XGBoost, TabPFN v2, CatBoost, and LightGBM under full-supervision and supervised few-shot settings. The latter use labeled target-task samples for fitting and serve as supervised comparisons on the same tasks, rather than inference-only FewTabBench results. These models serve as strong, task-specific baselines known for their robustness and efficiency. Through this comprehensive comparison, we assess whether reinforcement learning with structural priors can make LLMs competitive with specialized tabular predictors across fully supervised, few-shot, and cross-dataset zero-shot transfer settings.

Complete per-dataset results are provided in Appendix~\ref{appendix:full_quantitative_results}.

\subsection{Main Results}
\label{subse:fully}

\paragraph{Fully Supervised Performance.}
In the fully supervised setting, TabXLLM is strongest on classification and remains competitive on regression compared with specialized tabular models.

\par\addvspace{10pt}
\noindent\begin{minipage}{\linewidth}
    \begin{minipage}[t]{0.48\linewidth}
        \vspace{0pt}
As shown in \Cref{table:classification_fully_trained}, on 50 classification datasets, TabXLLM achieves the highest mean accuracy of \textbf{0.8436}, outperforming TabPFN (0.8413), CatBoost (0.8398), LightGBM (0.8321), and XGBoost (0.8234).
Similarly, for regression tasks (\Cref{table:regression_fully_trained}), TabXLLM achieves an NMAE of \textbf{0.1499}, outperforming XGBoost (0.1682) and LightGBM (0.1607), while remaining competitive with CatBoost (0.1458) and TabPFN (0.1236).
These findings show that structural-prior RL post-training can move an 8B LLM into the performance range of strong specialized tabular models.
    \end{minipage}\hfill
    \begin{minipage}[t]{0.49\linewidth}
        \vspace{0pt}
        \centering
\captionsetup{hypcap=false}\captionof{table}{Fully supervised results: mean accuracy ($\uparrow$) on classification datasets and mean NMAE ($\downarrow$) on regression datasets.}
\label{table:fully_supervised_results}
\label{table:classification_fully_trained}
\label{table:regression_fully_trained}
\setlength{\tabcolsep}{3pt}
\begin{tabular*}{\linewidth}{@{\extracolsep{\fill}}lcc@{}}
\toprule
Model & Mean Acc. $\uparrow$ & Mean NMAE $\downarrow$ \\
\midrule
\rowcolor{GroupPurple} \multicolumn{3}{l}{\textit{Tabular Models}} \\
XGBoost & 0.8234 & 0.1682 \\
LightGBM & 0.8321 & 0.1607 \\
CatBoost & 0.8398 & 0.1458 \\
TabPFN & 0.8413 & \best{0.1236} \\
\midrule
\rowcolor{GroupGreen} \multicolumn{3}{l}{\textit{PRPO-Post-Trained LLM}} \\
TabXLLM-8B & \best{0.8436} & 0.1499 \\
\bottomrule
\end{tabular*}
    \end{minipage}
\end{minipage}
\par\addvspace{4pt}

\begin{table}[!htbp]
\centering
\caption{Results on ZeroTabBench and FewTabBench, with 32-shot supervised comparisons. (a) reports accuracy ($\uparrow$) on 53 classification tasks, and (b) reports NMAE ($\downarrow$) on 21 regression tasks. Zero-shot entries are evaluated on ZeroTabBench; 32-shot in-context entries are evaluated on FewTabBench. TabLLM and specialized tabular models instead use the 32 labeled target-task samples for fitting.}
\label{table:zero_shot_32_shot_results}
\begin{subtable}[t]{0.48\textwidth}
\centering
\caption{Classification / Mean Acc. $\uparrow$}
\label{table:classification_few_zero}
{\small
\begin{tabular}{lcc}
\toprule
Model & \makecell{Zero-shot/\\transfer} & 32-shot \\
\midrule
\rowcolor{GroupBlue} \multicolumn{3}{l}{\textit{General-Purpose LLMs}} \\
Qwen2.5-7B & 0.5804 & 0.6012 \\
Qwen2.5-32B & 0.5930 & 0.6042 \\
Qwen3-8B & 0.5765 & 0.6110 \\
Qwen3-32B & 0.5599 & 0.6020 \\
Qwen3-235B & 0.5870 & 0.5973 \\
DS-R1 & 0.5313 & 0.5895 \\
DS-0528 & 0.5519 & 0.5943 \\
\midrule
\rowcolor{GroupPurple} \multicolumn{3}{l}{\textit{Tabular LLMs and Tabular Models}} \\
TabLLM & -- & 0.7246 \\
Tabula-8B & 0.6468 & 0.7165 \\
XGBoost & -- & 0.7219 \\
LightGBM & -- & 0.7312 \\
CatBoost & -- & 0.7396 \\
TabPFN & -- & 0.6974 \\
\midrule
\rowcolor{GroupGreen} \multicolumn{3}{l}{\textit{PRPO-Post-Trained LLM}} \\
\textbf{TabXLLM-8B} & \best{0.7021} & \best{0.7542} \\
\bottomrule
\end{tabular}
}
\end{subtable}
\hfill
\begin{subtable}[t]{0.48\textwidth}
\centering
\caption{Regression / Mean NMAE $\downarrow$}
\label{table:regression_few_zero}
{\small
\begin{tabular}{lcc}
\toprule
Model & \makecell{Zero-shot/\\transfer} & 32-shot \\
\midrule
\rowcolor{GroupBlue} \multicolumn{3}{l}{\textit{General-Purpose LLMs}} \\
Qwen2.5-7B & 0.2021 & 0.1961 \\
Qwen2.5-32B & 0.3146 & 0.2754 \\
Qwen3-8B & 0.2409 & 0.2173 \\
Qwen3-32B & 0.2370 & 0.2060 \\
Qwen3-235B & 0.2544 & 0.2190 \\
DS-R1 & 0.2329 & 0.2247 \\
DS-0528 & 2.3435 & 1.0681 \\
\midrule
\rowcolor{GroupPurple} \multicolumn{3}{l}{\textit{Tabular LLMs and Tabular Models}} \\
TabLLM & -- & -- \\
Tabula-8B & -- & -- \\
XGBoost & -- & 0.1744 \\
LightGBM & -- & 0.1246 \\
CatBoost & -- & 0.1187 \\
TabPFN & -- & 0.1119 \\
\midrule
\rowcolor{GroupGreen} \multicolumn{3}{l}{\textit{PRPO-Post-Trained LLM}} \\
\textbf{TabXLLM-8B} & \best{0.1614} & \best{0.1110} \\
\bottomrule
\end{tabular}
}
\end{subtable}
\end{table}

\paragraph{Generalization on ZeroTabBench and FewTabBench.}
As shown in \Cref{table:classification_few_zero}, on the 53 classification tasks in ZeroTabBench, TabXLLM's zero-shot accuracy (\textbf{0.7021}) substantially outperforms general-purpose LLMs even with 32-shot prompting, and already exceeds 32-shot TabPFN (0.6974).
On FewTabBench, TabXLLM's 32-shot performance further improves to \textbf{0.7542}, surpassing all 32-shot baselines, including CatBoost (0.7396), LightGBM (0.7312), TabLLM (0.7246), XGBoost (0.7219), and Tabula-8B (0.7165).

The comparison also suggests that model scale alone does not determine tabular prediction quality. Within the Qwen3 family, increasing model size from 8B to 235B does not improve 32-shot classification accuracy, while TabXLLM achieves stronger results with only 8B parameters. The larger DeepSeek reasoning models likewise remain below TabXLLM. These observations highlight the importance of aligning post-training with tabular structure, rather than relying solely on general-purpose model capacity or reasoning ability. Notably, TabXLLM's zero-shot classification accuracy exceeds that of the strongest 32-shot general-purpose LLM by 9.11 percentage points. This result further illustrates the value of reinforcement learning with structural priors for transferring tabular prediction capabilities to previously unseen tasks, even without labeled target-task demonstrations at inference time.

\FloatBarrier
\Needspace{255pt}
\begin{wrapfigure}{r}{0.60\textwidth}
    \vspace{-6pt}
    \centering
    \includegraphics[width=\linewidth]{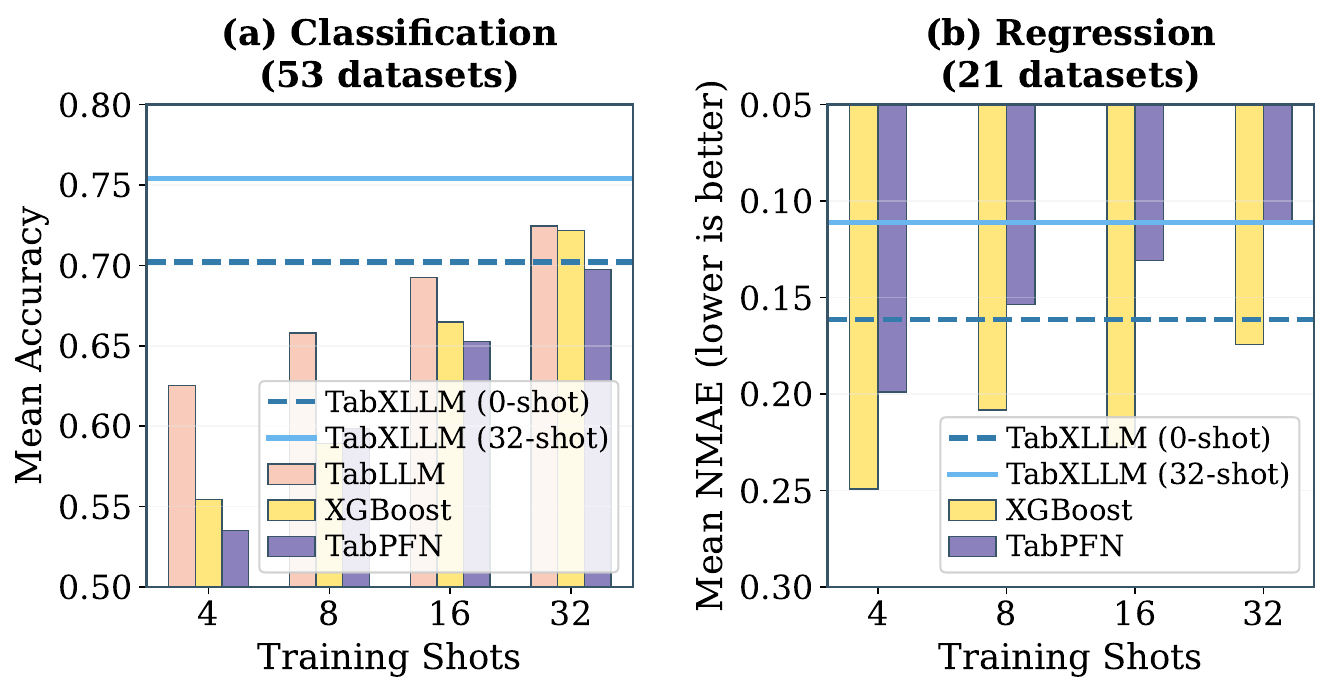}
    \caption{TabXLLM on ZeroTabBench and FewTabBench versus supervised few-shot baselines: (a) classification and (b) regression. Dashed and solid horizontal lines denote TabXLLM's ZeroTabBench and FewTabBench results, respectively.}
    \label{fig:zeroshot_vs_fewshot}
    \vspace{-6pt}
\end{wrapfigure}
As shown in \Cref{table:regression_few_zero}, similar trends are observed on the 21 regression tasks in ZeroTabBench, where TabXLLM's zero-shot NMAE (\textbf{0.1614}) is better than 32-shot XGBoost (0.1744) and all general-purpose LLMs even with 32-shot prompting, while TabXLLM's result on FewTabBench (\textbf{0.1110}) is the strongest among all compared methods.
We also validate PRPO on additional open-weight base models, including Llama-3.1-8B-Instruct, Mistral-7B-Instruct-v0.3, and DeepSeek-R1-0528-Qwen3-8B. As summarized in \Cref{table:cross_model_validation}, PRPO improves all reported metrics on all evaluated base models, indicating that the gain is not specific to Qwen3-8B-Base.

\begin{table}[!htbp]
\centering
\caption{Cross-model validation of PRPO across four base LLMs. ZeroTabBench entries use raw inference for base models and cross-dataset transfer for PRPO-post-trained models. Colored superscripts report direction-adjusted gains after PRPO post-training.}
\label{table:cross_model_validation}
\resizebox{\textwidth}{!}{
\begin{tabular}{lcccc}
\toprule
Model & \makecell{Full cls.\\Acc. $\uparrow$} & \makecell{ZeroTabBench\\cls. Acc. $\uparrow$} & \makecell{Full reg.\\NMAE $\downarrow$} & \makecell{ZeroTabBench\\reg. NMAE $\downarrow$} \\
\midrule
Qwen3-8B-Base & 0.499 & 0.508 & 2628.906 & $8.89{\times}10^{16}$ \\
TabXLLM-8B & 0.844\metricgain{$\uparrow$+0.345} & 0.702\metricgain{$\uparrow$+0.194} & 0.150\metricgain{$\downarrow$+2628.756} & 0.161\metricgain{$\downarrow$+8.89e16} \\
Llama-3.1-8B-Instruct & 0.470 & 0.501 & 0.559 & 8.060 \\
Llama-3.1-8B-Instruct+PRPO & 0.823\metricgain{$\uparrow$+0.353} & 0.608\metricgain{$\uparrow$+0.107} & 0.155\metricgain{$\downarrow$+0.404} & 0.222\metricgain{$\downarrow$+7.838} \\
Mistral-7B-Instruct-v0.3 & 0.207 & 0.208 & 21.446 & 842.030 \\
Mistral-7B-Instruct-v0.3+PRPO & 0.790\metricgain{$\uparrow$+0.583} & 0.605\metricgain{$\uparrow$+0.397} & 0.169\metricgain{$\downarrow$+21.277} & 0.671\metricgain{$\downarrow$+841.359} \\
DeepSeek-R1-0528-Qwen3-8B & 0.601 & 0.564 & 0.507 & 0.264 \\
DeepSeek-R1-0528-Qwen3-8B+PRPO & 0.762\metricgain{$\uparrow$+0.161} & 0.618\metricgain{$\uparrow$+0.054} & 0.223\metricgain{$\downarrow$+0.284} & 0.247\metricgain{$\downarrow$+0.017} \\
\bottomrule
\end{tabular}
}
\end{table}

\FloatBarrier
\Needspace{185pt}
\begin{wrapfigure}{r}{0.55\textwidth}
    \vspace{-6pt}
    \centering
    \includegraphics[width=\linewidth]{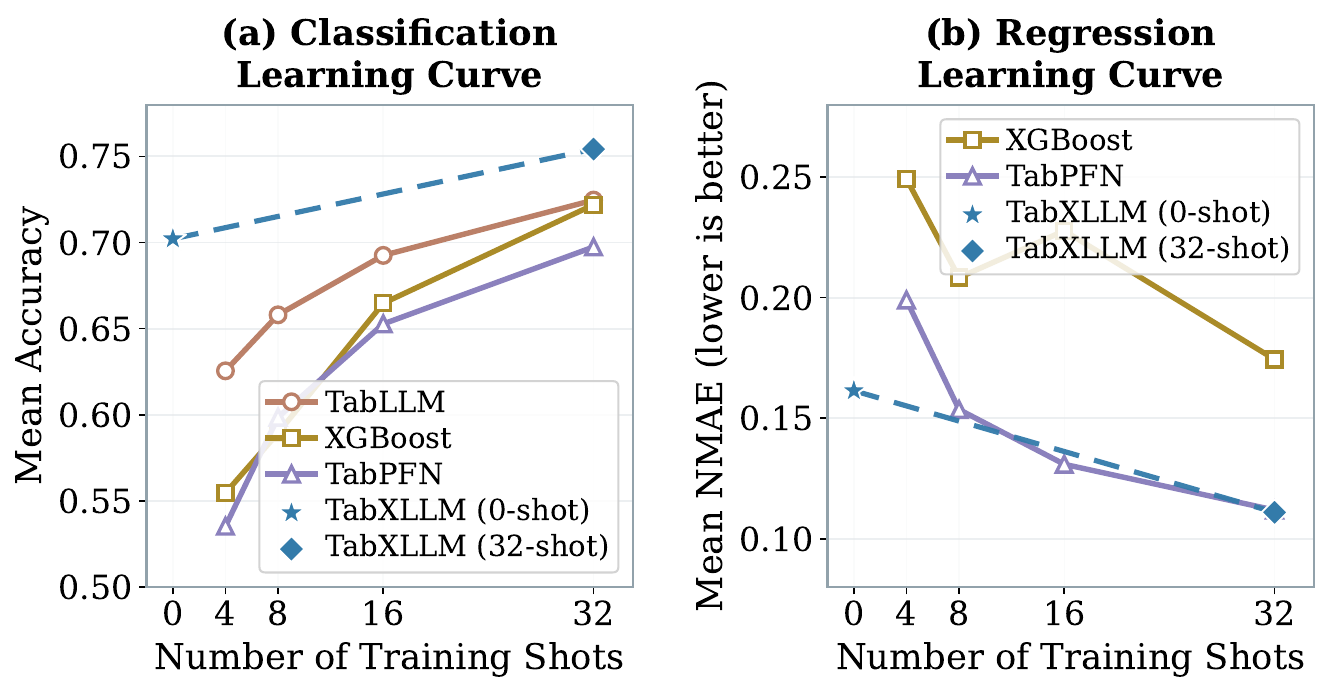}
        \caption{Few-shot comparisons on (a) classification and (b) regression tasks. TabXLLM's stars and 32-shot points denote ZeroTabBench and FewTabBench results, respectively.}
    \label{fig:learning_curve}
    \vspace{-6pt}
\end{wrapfigure}
\paragraph{Visual Analysis of Learning Dynamics.}
As shown in \Cref{fig:zeroshot_vs_fewshot}, TabXLLM's performance on ZeroTabBench and FewTabBench provides a direct reference against supervised few-shot baselines across classification and regression tasks.
\Cref{fig:learning_curve} further reveals contrasting scaling dynamics: while baselines rely on increasing labeled data for monotonic improvement, TabXLLM's ZeroTabBench classification accuracy already exceeds the 16-shot results of competitors.
This performance gap confirms that PRPO provides robust tabular post-training priors, significantly mitigating the reliance on task-specific supervision.

\paragraph{Per-Dataset Analysis.} \Cref{fig:scatter_combined} provides a fine-grained comparison of TabXLLM against baselines on the 53 classification tasks shared by ZeroTabBench and FewTabBench. On ZeroTabBench (top row), TabXLLM uses no target-task examples, while all baselines are trained on 32 labeled examples per dataset. Even so, zero-shot TabXLLM achieves a 9/19/25 (Win/Tie/Loss) record against 32-shot TabLLM, winning mostly on datasets where TabLLM struggles (accuracy $<$70\%). Against 32-shot XGBoost (22/6/25) and TabPFN (22/1/30), it wins on approximately 40\% of datasets, particularly on more challenging tasks. On FewTabBench (bottom row), providing just 32 in-context examples dramatically shifts the balance in our favor. Against TabLLM, our record improves to 20/24/9, matching or exceeding it on 83\% of datasets and trailing on only~9. Similarly, we achieve 31/7/15 against XGBoost and 35/3/15 against TabPFN. Notably, the performance gain from ZeroTabBench to FewTabBench is most pronounced on datasets where we previously trailed, indicating that in-context examples effectively help TabXLLM adapt to diverse data distributions. These results highlight two key findings: (1) PRPO training provides a strong cross-dataset zero-shot foundation that generalizes to unseen datasets; and (2) TabXLLM effectively leverages in-context examples to achieve state-of-the-art results without any gradient-based fine-tuning.

\par\addvspace{10pt}
\noindent\begin{minipage}{\linewidth}
    \begin{minipage}[t]{0.52\linewidth}
        \vspace{0pt}
        \paragraph{Data Efficiency.}
As illustrated in \Cref{fig:data_efficiency}, we quantify data efficiency by estimating the number of labeled samples each supervised baseline requires to match TabXLLM's ZeroTabBench performance via linear interpolation (or extrapolation beyond 32 shots) from their 4/8/16/32-shot results.
On the 53 classification tasks in ZeroTabBench (a), reaching TabXLLM's accuracy (70.2\%) requires TabLLM 21 samples, XGBoost 26 samples, and TabPFN 34 samples per dataset for few-shot training.
On the 21 regression tasks in ZeroTabBench (b), matching TabXLLM's NMAE (0.1614) requires XGBoost 41 samples, while TabPFN needs only 7 samples.
PRPO training thus provides data efficiency equivalent to 7--41 labeled samples per dataset, demonstrating practical value for data-scarce scenarios.
    \end{minipage}\hfill
    \begin{minipage}[t]{0.44\linewidth}
        \vspace{-6pt}
        \centering
        \includegraphics[width=0.842\linewidth]{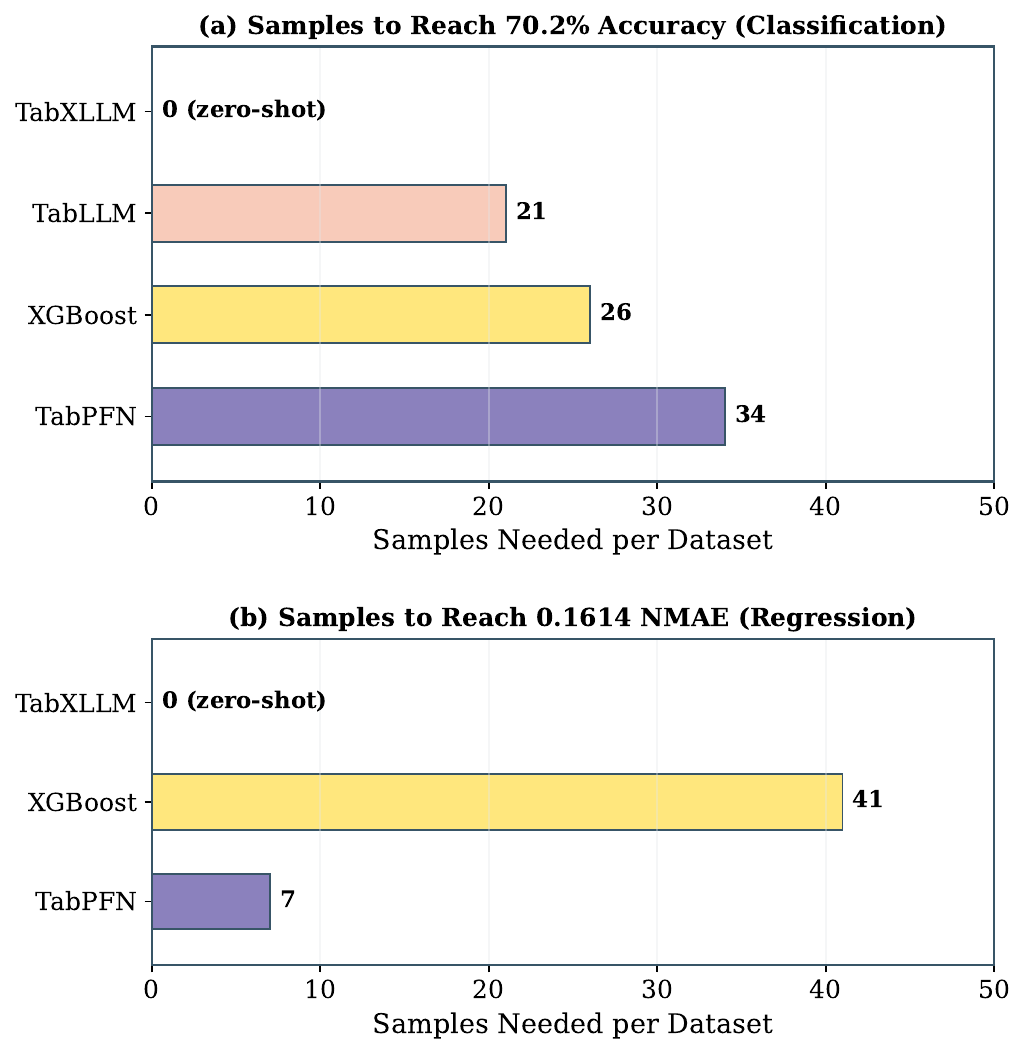}
        \captionsetup{hypcap=false}\captionof{figure}{Samples to match TabXLLM: (a) classification; (b) regression.}
        \label{fig:data_efficiency}
    \end{minipage}
\end{minipage}
\par\addvspace{10pt}


\par\addvspace{10pt}
\noindent\begin{minipage}{\linewidth}
    \centering
    \includegraphics[width=0.80\linewidth]{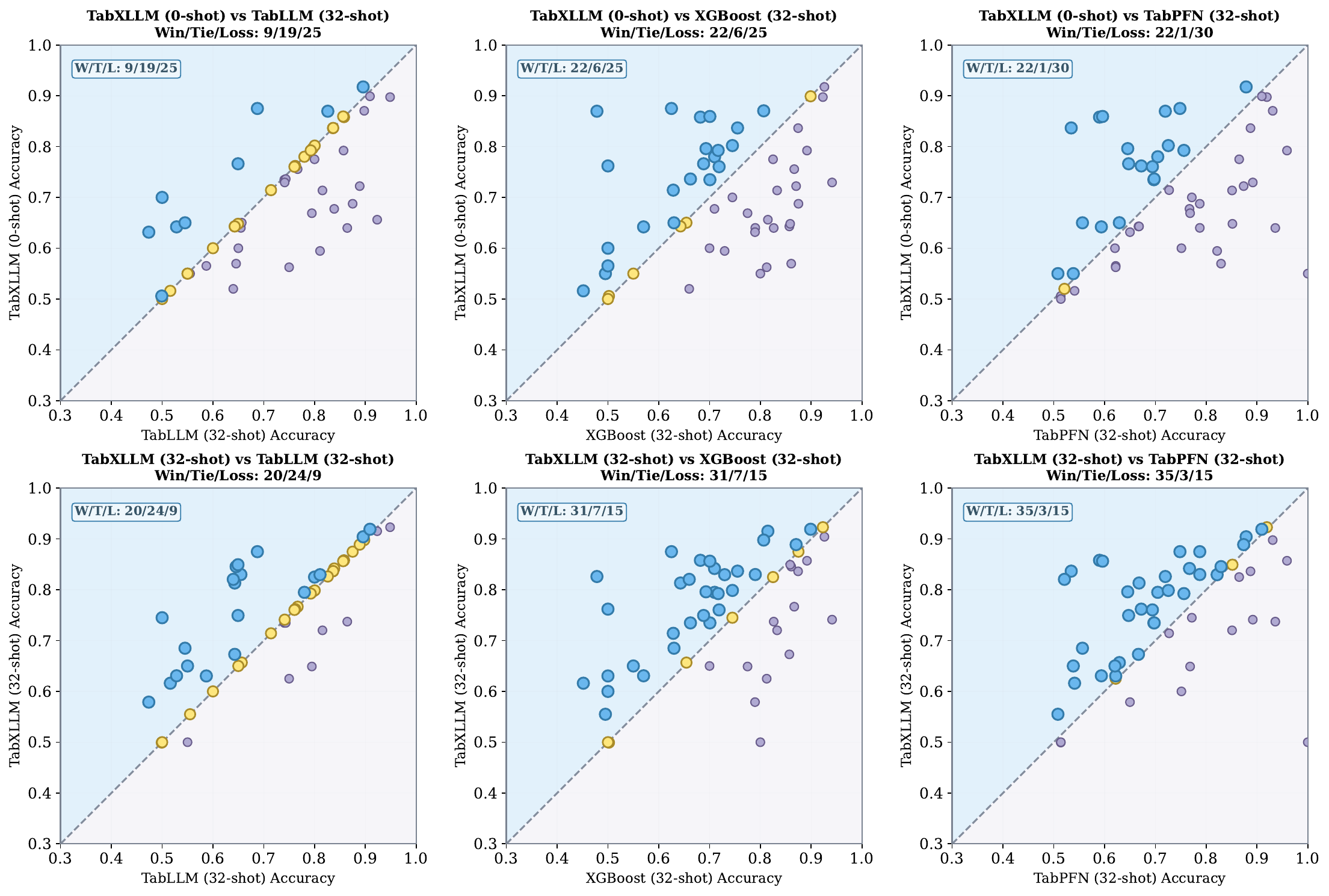}
    \captionsetup{hypcap=false}\captionof{figure}{TabXLLM on ZeroTabBench (top) and FewTabBench (bottom) versus 32-shot supervised baselines across 53 classification tasks. Points above the diagonal indicate TabXLLM wins; counts report Win/Tie/Loss statistics.}
    \label{fig:scatter_combined}
\end{minipage}
\par\addvspace{10pt}

ZeroTabBench and FewTabBench further reveal complementary roles for transferred knowledge and inference-time observations. With 32 labeled demonstrations, classification accuracy increases from 0.7021 to 0.7542, while regression NMAE decreases from 0.1614 to 0.1110, a relative reduction of 31.2\%. These improvements require no parameter updates. Thus, PRPO post-training provides a transferable starting point while preserving the model's ability to use target-task observations when available, supporting both prediction without target examples and adaptation through in-context demonstrations. The same post-trained model therefore supports both benchmarks by changing only the inference context.




\FloatBarrier
\par\addvspace{10pt}
\noindent\begin{minipage}[b]{\linewidth}
    \begin{minipage}[t]{0.58\linewidth}
        \vspace{0pt}
        \paragraph{Ablation Study.}
Finally, we isolate where the gain of PRPO comes from in \Cref{fig:prpo_grpo_comparison}.
We use standard GRPO with a single serialization view and $G{=}5$ rollouts as the reference baseline (A0), and then compare five variants under the same total rollout budget of 20 rollouts per original sample: GRPO with more rollouts (A1, $m{=}1,G{=}20$), GRPO with random column-order training (A2, $m{=}1,G{=}20$), permutation-aware training with only intra-permutation advantages (A3, $m{=}4,G{=}5$), permutation-aware training with only inter-permutation advantages (A4, $m{=}4,G{=}5$), and full PRPO with both advantage levels (A5, $m{=}4,G{=}5$).
This design separates the effects of rollout budget, random column ordering, permutation augmentation, and the two-level advantage estimator.
Increasing the rollout budget alone brings only a small gain over standard GRPO (+0.0061 classification accuracy and -0.0028 regression NMAE), indicating that PRPO's improvement is not simply due to more sampled outputs.
Random column-order training also remains far below full PRPO.
The intra-only and inter-only variants form a stronger intermediate tier, showing that both local within-permutation comparison and global cross-permutation comparison are useful.
Full PRPO is consistently best, improving over A2/A3/A4 by +0.0780/+0.0520/+0.0470 in classification accuracy and by -0.0151/-0.0113/-0.0091 in regression NMAE.
    \end{minipage}\hfill
    \begin{minipage}[t]{0.39\linewidth}
        \vspace{0pt}
        \centering
        \includegraphics[width=\linewidth]{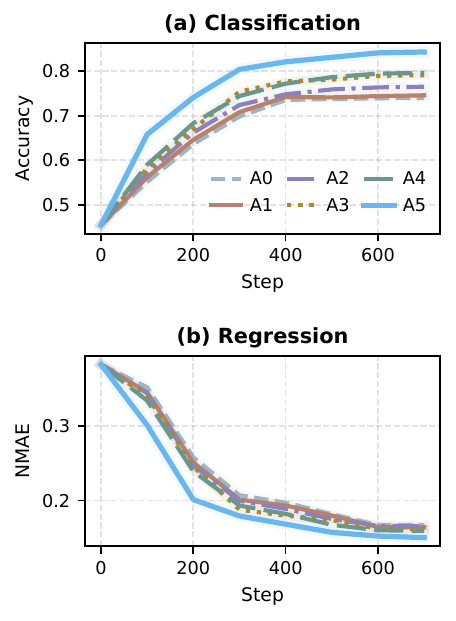}
        \captionsetup{hypcap=false}\captionof{figure}{Attribution ablation on classification and regression datasets. A0 is standard GRPO; A1--A5 isolate rollout count, random column order, intra-only, inter-only, and full PRPO under matched rollout budgets.}
    \label{fig:prpo_grpo_comparison}
    \end{minipage}
\end{minipage}
\par\addvspace{10pt}

\section{Related Work}
\textbf{Traditional tabular models}~\citep{arik2021tabnet,huang2020tabtransformer,gorishniy2021revisiting,popov2019neural,somepalli2021saint,borisov2024deep} such as gradient-boosted decision trees (XGBoost, LightGBM)~\citep{chen2016xgboost,ke2017lightgbm} and specialized deep architectures (TabNet, TabTransformer)~\citep{arik2021tabnet, huang2020tabtransformer} achieve strong supervised performance but lack interpretability and cross-task generalization~\citep{hegselmann2023tabllm,zha2023tablegpt,gardner2024large,touvron2023llama,qwen2025qwen25}. Approaches like TabPFN~\citep{hollmann2023tabpfn,hollmann2025accurate} introduce pretrained priors for few-shot settings, yet remain task-specific. Some tabular methods described as zero-shot still condition predictions on labeled examples from the target task, and thus do not support zero-shot transfer without target training data~\citep{wu2025zeroshotmetalearningtabularprediction,breejen2025finetunedincontextlearningtransformers,kong2026tabfmzeroshotfoundationmodel,cheng2026enoughbinaryclassificationpriors}.

\textbf{Tabular LLMs}. TabLLM~\citep{hegselmann2023tabllm} adopts a closed-token prediction paradigm by serializing tabular data into natural-language prompts, enabling efficient zero- and few-shot modeling on structured data. It generally achieves higher accuracy than subsequent open-token generation models.
Tabula-8B~\citep{gardner2024large}, built upon Llama-3 8B, employs an open-token prediction framework with large-scale tabular transfer pretraining, offering greater generality but slightly lower precision. Therefore, we select TabLLM as the main comparison baseline in this work.

\textbf{Reasoning LLMs} (e.g., GPT-4~\citep{openai2024gpt4}, DeepSeek-R1~\citep{deepseek-ai2025deepseekr1}, Qwen3~\citep{bai2023qwen,qwen2025qwen25,yang2025qwen3}) demonstrate strong multi-step reasoning with reinforcement learning methods such as GRPO, but their ability does not naturally transfer to tabular data due to a modality gap and sparse outcome-level rewards.


\section{Conclusions}

We propose Permutation Relative Policy Optimization, a structural-prior RL post-training method for strengthening LLMs on tabular prediction. PRPO encodes column-permutation invariance to transform sparse outcome rewards into denser learning signals, improving optimization stability and cross-dataset generalization. Experiments show that TabXLLM becomes genuinely competitive with specialized tabular models, achieves strong fully supervised and few-shot performance, and substantially outperforms much larger general-purpose LLMs on ZeroTabBench while retaining transparent prediction traces. FewTabBench further demonstrates its ability to adapt through inference-time observations without parameter updates.

\section*{Acknowledgements}
This research is partly supported by the National Natural Science Foundation of China (No. 62606436), the Young Program of Guangdong Province (No. 2025QN12X202), the Guangdong Basic and Applied Basic Research Foundation (2025A1515110170, 2026A1515011793), and the Youth S\&T Talent Support Programme of Guangdong Provincial Association for Science and Technology (SKXRC2025467).





\FloatBarrier
\bibliographystyle{plainnat}
\bibliography{references}

\newpage
\appendix
\onecolumn
\flushbottom

\section{Theoretical Analysis of PRPO}\label{appendix:prpo_proof}

We analyze the two-level advantage mixture, correct-answer coverage under independent rollouts, and the gradient of the clipped PRPO surrogate. We also derive a conditional stationarity bound with an explicit gradient-bias term.

\subsection{Notation and Preliminaries}


We first establish the notation used throughout this appendix:
\begin{tcolorbox}[title=\textbf{Notation}, colback=FigBlue!5, colframe=FigInk, toptitle=0.3mm,bottomtitle=0.3mm,top=1.7mm,bottom=1.7mm]
    \vspace{0.5em}
        $T = \{x_1, x_2, \dots, x_n, y\}$: A tabular sample with $n$ features and label $y$.
        
    \vspace{1em}
        $S$: The symmetric group of all permutations on $\{1, 2, \dots, n\}$, with $|S| = n!$.
        
    \vspace{1em}
        $\pi_k \in S$: The $k$-th sampled permutation, $k \in \{1, \dots, m\}$.
        
    \vspace{1em}
        $t_k = \pi_k(T)$: The $k$-th permuted variant of $T$.
        
    \vspace{1em}
        $q_k$: The serialized natural language representation of $t_k$.
        
    \vspace{1em}
        $o_{k,i}$: The $i$-th output sampled from $\pi_\theta(\cdot|q_k)$, $i \in \{1, \dots, G\}$.
        
    \vspace{1em}
        $R_{k,i} = R(\hat{y}_{k,i})$: The reward for output $o_{k,i}$.
        
    \vspace{1em}
        $\pi_\theta$: The policy model with parameters $\theta$.
        
    \vspace{1em}
        $\pi_{\text{ref}}$: The reference policy (frozen).
    
\end{tcolorbox}

\subsection{Column-Permutation Invariance}

\begin{definition}[Column-Permutation Invariance]
A tabular prediction task is said to satisfy \textbf{column-permutation invariance} if for any permutation $\pi \in S$ and any tabular sample $T$, the ground-truth label $y^*$ remains unchanged:
\begin{equation}
f^*(T) = f^*(\pi(T)) = y^*,
\end{equation}
where $f^*$ denotes the true underlying function.
\end{definition}

\begin{proposition}[Reward Invariance Under Permutation]
\label{prop:reward_invariance}
Let $q$ and $q'$ serialize a tabular sample and a label-preserving column permutation. Assume that the policy induces the same distribution of extracted predictions and formatting validity under both prompts. Then the expected reward is invariant:
\begin{equation}
\mathbb{E}_{o\sim\pi_\theta(\cdot\mid q)}[R(o)]
=\mathbb{E}_{o\sim\pi_\theta(\cdot\mid q')}[R(o)].
\end{equation}
\end{proposition}

\begin{proof}
Write $z(o)$ for the extracted prediction and formatting-validity information, and write the rule-based reward as $r(z(o),y^*)$. Since $y^*$ is unchanged and the law $P_\theta$ of $z(o)$ is the same for both prompts,
\begin{equation}
\mathbb{E}_{o\sim\pi_\theta(\cdot\mid q)}[R(o)]
=\int r(z,y^*)\,dP_\theta(z\mid q)
=\int r(z,y^*)\,dP_\theta(z\mid q')
=\mathbb{E}_{o\sim\pi_\theta(\cdot\mid q')}[R(o)].
\end{equation}
This argument applies to both the classification and regression rewards.
\end{proof}

\Needspace*{100pt}
\subsection{Variance of the Two-Level Advantage Mixture}

We characterize the variance of the scalar advantage mixture. Throughout this analysis, advantages are well-defined with finite second moments; zero-variance groups are assigned zero advantages.

\begin{proposition}[Variance of the Two-Level Advantage Mixture]
\label{prop:advantage_variance}
For a fixed output index, let $X=\hat A^{(1)}_{k,i}$ and $Y=\hat A^{(2)}_{k,i}$. Denote $v_1=\operatorname{Var}(X)$, $v_2=\operatorname{Var}(Y)$, and $C=\operatorname{Cov}(X,Y)$. For any fixed $\alpha\in[0,1]$,
\begin{equation}
V(\alpha):=\operatorname{Var}(\alpha X+(1-\alpha)Y)
=\alpha^2v_1+(1-\alpha)^2v_2+2\alpha(1-\alpha)C.
\end{equation}
\end{proposition}

\begin{proof}
Let $X_c=X-\mathbb E[X]$ and $Y_c=Y-\mathbb E[Y]$. Expanding the squared centered mixture gives
\begin{equation}
V(\alpha)=\mathbb E\big[(\alpha X_c+(1-\alpha)Y_c)^2\big]
=\alpha^2\mathbb E[X_c^2]+(1-\alpha)^2\mathbb E[Y_c^2]+2\alpha(1-\alpha)\mathbb E[X_cY_c].
\end{equation}
Substituting $\mathbb E[X_c^2]=v_1$, $\mathbb E[Y_c^2]=v_2$, and $\mathbb E[X_cY_c]=C$ proves the identity.
\end{proof}

\subsection{Reward Densification}

\begin{theorem}[Reward Densification via Permutation]
\label{thm:reward_densification}
Condition on a tabular sample. Assume that the correctness indicators of all rollouts are independent and have a common success probability $p_{\text{success}}\in(0,1)$ across the considered serializations. For GRPO with $G$ rollouts, the probability of at least one correct answer is:
\begin{equation}
P_{\text{GRPO}} = 1 - (1 - p_{\text{success}})^G.
\end{equation}
For PRPO with $m$ permutations and $G$ rollouts each, this probability becomes:
\begin{equation}
P_{\text{PRPO}} = 1 - (1 - p_{\text{success}})^{mG}.
\end{equation}
Thus, $P_{\text{PRPO}} > P_{\text{GRPO}}$ for $m > 1$ under these assumptions.
\end{theorem}

\begin{proof}
The probability of receiving no correct answer in GRPO is $(1 - p_{\text{success}})^G$, since each of the $G$ rollouts independently fails with probability $1 - p_{\text{success}}$.

For PRPO, we have $m$ permutations, each with $G$ rollouts, giving a total of $mG$ independent trials. The probability of all trials failing is $(1 - p_{\text{success}})^{mG}$.

Therefore:
\begin{align}
P_{\text{PRPO}} - P_{\text{GRPO}} &= (1 - p_{\text{success}})^G - (1 - p_{\text{success}})^{mG} \\
&= (1 - p_{\text{success}})^G \left[1 - (1 - p_{\text{success}})^{(m-1)G}\right] \\
&> 0, \quad \text{for } m > 1 \text{ and } p_{\text{success}} \in (0, 1).
\end{align}
\end{proof}

\begin{corollary}[Reward Signal Coverage]
\label{cor:sample_efficiency}
PRPO evaluates $mG$ rollouts across $m$ label-preserving permutations. Relative to GRPO with $G$ rollouts, the improvement in correct-answer coverage under the assumptions of Theorem~\ref{thm:reward_densification} is:
\begin{equation}
\frac{P_{\text{PRPO}}}{P_{\text{GRPO}}} = \frac{1 - (1 - p_{\text{success}})^{mG}}{1 - (1 - p_{\text{success}})^G}.
\end{equation}
Under the assumptions of Theorem~\ref{thm:reward_densification}, PRPO with $mG$ rollouts achieves higher correct-answer coverage than GRPO with $G$ rollouts for $m>1$ and $p_{\text{success}}\in(0,1)$. Correct-answer coverage measures the probability of observing at least one correct answer. Valid but incorrect answers may still receive a formatting reward.
\end{corollary}

\subsection{Gradient of the Clipped PRPO Surrogate}

We analyze the surrogate actually differentiated during a policy update.
Condition on a fixed rollout batch $\mathcal B$ containing $N$ outputs,
and treat its computed advantages $A_j=\hat A_j^{\mathrm{PRPO}}$ as
constants during differentiation. Let $r_j(\theta)$ be the probability
ratio with a fixed denominator, and let $K_{\mathcal B}(\theta)$ denote
the differentiable regularization term used for this batch. Define
\begin{equation}
\widehat L_{\mathcal B}(\theta)
=\sum_{j=1}^{N}
\min\!\left\{r_j(\theta)A_j,
\operatorname{clip}(r_j(\theta),1-\epsilon,1+\epsilon)A_j\right\}
-\beta K_{\mathcal B}(\theta).
\end{equation}
This uses the sum convention of the paper; for a consistently averaged
objective, the corresponding gradient terms are averaged as well.

\begin{proposition}[Gradient of the Clipped Surrogate]
\label{prop:clipped_prpo_gradient}
Suppose the output probabilities are positive and differentiable,
$K_{\mathcal B}$ is differentiable, and the probability-ratio
denominators are positive and independent of $\theta$. Away from clipping thresholds,
\begin{equation}
\nabla\widehat L_{\mathcal B}(\theta)
=\sum_{j=1}^{N}
I_j(\theta)r_j(\theta)A_j
\nabla\log\pi_\theta(o_j\mid q_j)
-\beta\nabla K_{\mathcal B}(\theta),
\end{equation}
where
\begin{equation}
I_j(\theta)=
\begin{cases}
1, & A_j>0\ \text{and}\ r_j(\theta)<1+\epsilon,\\
1, & A_j<0\ \text{and}\ r_j(\theta)>1-\epsilon,\\
0, & \text{otherwise}.
\end{cases}
\end{equation}
\end{proposition}

\begin{proof}
For fixed $A>0$, the clipped contribution is
$A\min\{r,1+\epsilon\}$, so its derivative with respect to $r$
is $A$ below the upper threshold and zero above it.
For $A<0$, the contribution is $A\max\{r,1-\epsilon\}$,
whose derivative is zero below the lower threshold and $A$ above it.
For $A=0$, the contribution is identically zero.
Since the ratio denominator is fixed,
$\nabla r_j=r_j\nabla\log\pi_\theta(o_j\mid q_j)$.
Summing these derivatives and differentiating the regularizer gives
the result. At a clipping threshold, one may use a generalized
gradient; the displayed equality is asserted only away from thresholds.
\end{proof}

\subsection{Conditional Stationarity Bound with Gradient Bias}

We analyze stochastic ascent of the expected-reward objective with an explicit gradient-bias term.
Let $J(\theta)=\mathbb E_{q,o\sim\pi_\theta(\cdot\mid q)}[R(o)]$,
where $q$ follows a fixed distribution over training samples and column
permutations. Let $\mathcal F_t$ contain the optimization history and
current parameters, before sampling the next stochastic update.
For the update $\theta_{t+1}=\theta_t+\eta g_t$, define
\begin{equation}
b_t=\mathbb E[g_t\mid\mathcal F_t]-\nabla J(\theta_t).
\end{equation}

\begin{theorem}[Conditional Stationarity Bound]
\label{thm:convergence}
Assume that $J$ is $L$-smooth, $J(\theta)\le J_{\max}<\infty$,
the initial parameter $\theta_0$ is fixed, and
$\mathbb E[\|g_t\|^2\mid\mathcal F_t]\le M^2$ for every step.
For a horizon $T$ and constant step size $\eta=c/\sqrt T$, $c>0$,
\begin{equation}
\min_{0\le t<T}\mathbb E\|\nabla J(\theta_t)\|^2
\le
\frac{2(J_{\max}-J(\theta_0))}{c\sqrt T}
+\frac{LcM^2}{\sqrt T}
+\frac1T\sum_{t=0}^{T-1}\mathbb E\|b_t\|^2.
\end{equation}
\end{theorem}

\begin{proof}
The lower smoothness bound gives
\begin{align}
\mathbb E[J(\theta_{t+1})\mid\mathcal F_t]
&\ge J(\theta_t)
+\eta\langle\nabla J(\theta_t),\nabla J(\theta_t)+b_t\rangle
-\frac{L\eta^2}{2}
  \mathbb E[\|g_t\|^2\mid\mathcal F_t]\\
&\ge J(\theta_t)
+\frac{\eta}{2}\|\nabla J(\theta_t)\|^2
-\frac{\eta}{2}\|b_t\|^2
-\frac{L\eta^2M^2}{2},
\end{align}
where the second inequality uses
$\langle u,v\rangle\ge-(\|u\|^2+\|v\|^2)/2$.
Taking expectations, summing over $t=0,\ldots,T-1$, and using
$\mathbb E J(\theta_T)\le J_{\max}$ yields
\begin{equation}
\frac1T\sum_{t=0}^{T-1}\mathbb E\|\nabla J(\theta_t)\|^2
\le
\frac{2(J_{\max}-J(\theta_0))}{\eta T}
+L\eta M^2
+\frac1T\sum_{t=0}^{T-1}\mathbb E\|b_t\|^2.
\end{equation}
The minimum is no larger than this average. Substituting
$\eta=c/\sqrt T$ proves the result.
\end{proof}

If the average squared bias vanishes, the bound gives a vanishing
expected squared-gradient criterion. If it is bounded by a nonzero
$\delta^2$, the bound is $O(T^{-1/2})+\delta^2$.

\subsection{Comparison with GRPO}

\begin{proposition}[PRPO Improves Reward Coverage in Sparse Reward Settings]
\label{prop:prpo_dominates}
Under the independent common-success-probability model of Theorem~\ref{thm:reward_densification}, PRPO increases the probability of observing at least one correct answer when using $mG$ rollouts instead of $G$. Specifically:
\begin{equation}
\frac{P_{\text{PRPO}}}{P_{\text{GRPO}}} =
\frac{1 - (1 - p_{\text{success}})^{mG}}{1 - (1 - p_{\text{success}})^G}
> 1,
\end{equation}
for $m>1$ and $p_{\text{success}}\in(0,1)$.
\end{proposition}

\begin{proof}
The ratio follows from Theorem~\ref{thm:reward_densification}. For $m>1$ and $p_{\text{success}}\in(0,1)$, $(1-p_{\text{success}})^{mG}<(1-p_{\text{success}})^G$, so the ratio exceeds one.
\end{proof}

\flushbottom
\FloatBarrier
\section{PRPO Fine-tuning Setting}\label{appendixA}
The PRPO training procedure is summarized in \Cref{alg:prpo}, and the hyperparameters and training configurations of PRPO for TabXLLM are provided in \Cref{tab:prpo-hyperparams}.

\begin{algorithm}[H]
\caption{PRPO Fine-Tuning}
\label{alg:prpo}
\begin{algorithmic}[1]
\REQUIRE Dataset $D$; policy $\pi_\theta$; permutations $m$; rollouts $G$
\ENSURE Updated parameters $\theta$
\WHILE{not converged}
  \STATE Sample minibatch $\mathcal{B} \subset D$
  \FORALL{$T \in \mathcal{B}$}
    \STATE Generate $m$ permuted variants $\{t_1, \dots, t_m\}$
    \FOR{$k=1$ to $m$}
      \STATE Sample $G$ outputs $\{o_{k,i}\}_{i=1}^G \sim \pi_\theta(\cdot|t_k)$
      \STATE Compute rewards $R_{k,i}$ and intra-perm. $\hat{A}^{(1)}_{k,i}$
    \ENDFOR
    \STATE Compute inter-perm. advantages $\hat{A}^{(2)}_{k,i}$
    \STATE Aggregate: $\hat{A}^{\text{PRPO}}_{k,i} = \alpha \hat{A}^{(1)}_{k,i} + (1-\alpha) \hat{A}^{(2)}_{k,i}$
    \STATE Compute $\mathcal{L}^{\text{PRPO}}$ with PPO clipping
  \ENDFOR
  \STATE Update $\theta \leftarrow \theta + \eta \nabla_\theta \mathcal{L}^{\text{PRPO}}$
\ENDWHILE
\end{algorithmic}
\end{algorithm}

\begingroup
\setlength{\intextsep}{6pt plus 1pt minus 1pt}
\begin{table}[ht]
\centering
\renewcommand{\arraystretch}{0.92}
\caption{Key hyperparameters and environment settings for PRPO fine-tuning on \textbf{TabXLLM}.}
\label{tab:prpo-hyperparams}
\begin{tabular}{ll}
\toprule
\textbf{Parameter} & \textbf{Value} \\
\midrule
Base model & Qwen3-8B-Base \\
Training batch size & 128 \\
PPO mini-batch size & 32 \\
Micro-batch per GPU & 4 \\
Max prompt length & 5120 \\
Max response length & 1024 \\
Learning rate & $1\times10^{-6}$ \\
KL loss coefficient $\beta$ & 0.001 \\
number of permutations $m$ & 4 \\
intra-permutation advantages coefficient $\alpha$ & 0.1 \\
inter-permutation advantages coefficient $1-\alpha$ & 0.9 \\
Entropy coefficient & 0 \\
Rollout parallel size & 2 \\
Number of rollouts per sample $G$ & 5 \\
GPU memory utilization & 0.6 \\
Number of GPUs per node & 8 \\
Number of nodes & 1 \\
Total training epochs & 30 \\
CUDA version & 12.6 \\
\bottomrule
\end{tabular}

\end{table}
\endgroup

\section{Hyperparameter Tuning of XGBoost and TabPFN v2}
\label{appendix:baseline_settings}
In this section, we introduce the hyperparameter tuning and experimental setup of the two baseline models used in our experiments, XGBoost and TabPFN.

\subsection{TabPFN}
For TabPFN, we employed the official \textit{TabPFN Extensions} implementation and extended it to handle heterogeneous OpenML datasets containing missing or non-numeric features. 
Missing values were imputed using the most frequent strategy.
The TabPFN model used its default architecture and priors without additional hyperparameter tuning. 

\Needspace*{120pt}
\subsection{XGBoost}
We adopted the \texttt{XGBoost} classifier for both binary and multi-class classification tasks. 
Hyperparameters were optimized using the \texttt{Optuna} framework with a random search strategy (\texttt{RandomSampler}), minimizing the log-loss on the training set. 
Label encoding was applied to categorical targets. 
For categorical features, a one-hot encoder was used, while numerical or string-valued columns were coerced to numeric values. 
Each optimization trial trained a model using early stopping and evaluated the log-loss to guide parameter selection. 
We used the same hyperparameter search space as~\citep{shwartz-ziv2021tabular}, which is provided in~\Cref{tab:xgboost_hyperparams}.

\begingroup
\setlength{\intextsep}{6pt plus 1pt minus 1pt}
\begin{table}[H]
\centering
\renewcommand{\arraystretch}{0.92}
\caption{Hyperparameter search space for XGBoost.}
\label{tab:xgboost_hyperparams}
\begin{tabular}{lllll}
\toprule
\textbf{baseline} & \textbf{name} & \textbf{type} & \textbf{log} & \textbf{range} \\
\midrule
\multirow{10}{*}{XGBoost} 
 & learning\_rate & float & $[e^{-7}, 1]$ & yes \\
 & max\_depth & int & $[1, 10]$ & -- \\
 & subsample & float & $[0.2, 1]$ & -- \\
 & colsample\_bytree & float & $[0.2, 1]$ & -- \\
 & colsample\_bylevel & float & $[0.2, 1]$ & -- \\
 & min\_child\_weight & float & $[e^{-16}, e^{5}]$ & yes \\
 & alpha & float & $[e^{-16}, e^{2}]$ & yes \\
 & lambda & float & $[e^{-16}, e^{2}]$ & yes \\
 & gamma & float & $[e^{-16}, e^{2}]$ & yes \\
 & n\_estimators & int & $[100, 4000]$ & -- \\
\bottomrule
\end{tabular}
\end{table}
\endgroup




\raggedbottom
\section{Data}\label{appendixB}
\subsection{Data Distribution}
\begin{figure}[H]
    \centering
    \includegraphics[width=0.81\linewidth]{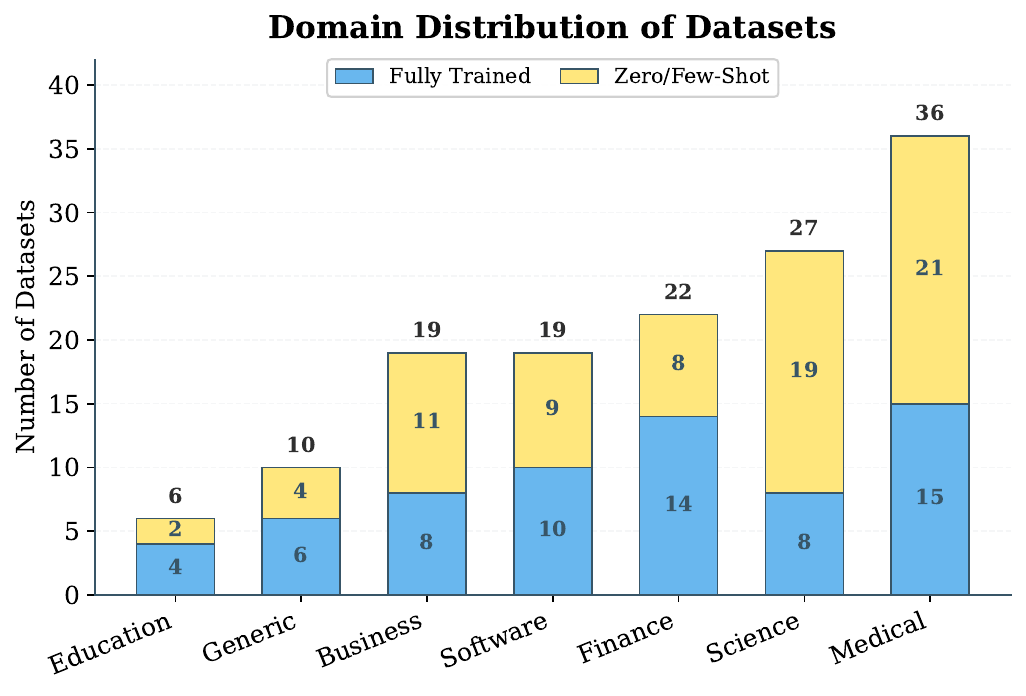}
    \caption{Domain distribution of fully trained datasets and the tasks shared by ZeroTabBench and FewTabBench.}
    \label{fig:domain_distribution}
\end{figure}

To enable LLMs to acquire robust tabular reasoning capabilities through PRPO, the collection of diverse and sufficiently large tabular datasets is essential.
We collected 139 public datasets from OpenML, covering a wide range of application domains.
To comprehensively evaluate the capability of TabXLLM, datasets from different domains are divided into two groups: one subset is used for fully trained experiments to assess supervised performance, while the remaining 53 classification and 21 regression datasets form ZeroTabBench and FewTabBench for evaluating cross-dataset generalization without and with inference-time observations, respectively.
The overall domain distribution of these datasets is illustrated in \Cref{fig:domain_distribution}.
\vspace{-6pt}

\begingroup
\setlength{\parskip}{2pt}
\tcbset{before skip=2pt,after skip=3pt,boxsep=0.3mm}
\vspace{4pt}
\noindent\begin{minipage}{\linewidth}
\subsection{Text Template for Qwen and DeepSeek-R1 Series Models}
\label{appendix:text template qwen deepseek}
In the ZeroTabBench experiments presented in \Cref{table:classification_few_zero} and \Cref{table:regression_few_zero}, the Qwen and DeepSeek-R1 series models are evaluated under raw zero-shot inference. The text templates for classification and regression tasks are shown below.
\vspace{4pt}
\begin{tcolorbox}[
  enhanced,
  unbreakable,
  before skip=2pt,
  colback=FigBlue!8,
  colframe=FigBlueDark,
  colbacktitle=FigBlueDark,
  coltitle=white,
  title={Example of the text template used for Qwen and DeepSeek-R1 evaluation on ZeroTabBench classification tasks},
  fonttitle=\bfseries\small,
  fontupper=\small,
  boxrule=0.9pt,
  arc=2mm,
  left=1.2mm,right=1.2mm,top=2.6780162pt,bottom=2.6791705pt,
  toptitle=0.4mm,bottomtitle=0.4mm,
  before title={\rule[-3.5pt]{0pt}{12pt}},after title={\rule[-3.5pt]{0pt}{12pt}}
]
\begin{tcolorbox}[
  unbreakable,
  colback=white,
  colframe=FigBlue!70!FigInk,
  colbacktitle=FigBlue!20,
  coltitle=FigBlueDark,
  title={System},
  fonttitle=\bfseries\small,
  fontupper=\small,
  boxrule=0.5pt,
  arc=1mm,
  left=1mm,right=1mm,top=2.0414130pt,bottom=1.0920107pt,
  toptitle=0.4mm,bottomtitle=0.4mm,
  before title=\vphantom{System},after title=\vphantom{System},before skip=0.5899237pt,after skip=0.5899237pt
]
\textbf{Role:} You are a helpful assistant.
\end{tcolorbox}

\begin{tcolorbox}[
  unbreakable,
  colback=white,
  colframe=FigBlue!70!FigInk,
  colbacktitle=FigBlue!20,
  coltitle=FigBlueDark,
  title={User},
  fonttitle=\bfseries\small,
  fontupper=\small,
  boxrule=0.5pt,
  arc=1mm,
  left=1mm,right=1mm,top=2.0407305pt,bottom=1.1370107pt,
  toptitle=0.4mm,bottomtitle=0.4mm,
  before title=\vphantom{System},after title=\vphantom{System},before skip=4.4339521pt,after skip=4.4339521pt
]
\textbf{Input:} The Married is Yes. The Education is Graduate. The LoanAmount is 44.0. The CoapplicantIncome is 0.0.\\
The Gender is Male. The ApplicantIncome is 5649. The Property\_Area is Urban. The Dependents is 1.\\
The Self\_Employed is No. The Loan\_Amount\_Term is 360.0. The Credit\_History is 1.0.

\vspace{3.5408021pt}
\textbf{Task:} {\color{FigPeachDark}\textbf{Will the applicant's loan be approved? yes or no?}}

\vspace{3.4328021pt}
\textbf{Output Format:} Return the final answer in {\color{FigPurpleDark}\texttt{<answer> </answer>}} tags.
\end{tcolorbox}
\end{tcolorbox}

\vspace{4pt}
\begin{tcolorbox}[
  enhanced,
  unbreakable,
  before skip=2pt,
  colback=FigBlue!8,
  colframe=FigBlueDark,
  colbacktitle=FigBlueDark,
  coltitle=white,
  title={Example of the text template used for Qwen and DeepSeek-R1 evaluation on ZeroTabBench regression tasks},
  fonttitle=\bfseries\small,
  fontupper=\small,
  boxrule=0.9pt,
  arc=2mm,
  left=1.2mm,right=1.2mm,top=2.6781321pt,bottom=2.6791560pt,
  toptitle=0.4mm,bottomtitle=0.4mm,
  before title={\rule[-3.5pt]{0pt}{12pt}},after title={\rule[-3.5pt]{0pt}{12pt}}
]
\begin{tcolorbox}[
  unbreakable,
  colback=white,
  colframe=FigBlue!70!FigInk,
  colbacktitle=FigBlue!20,
  coltitle=FigBlueDark,
  title={System},
  fonttitle=\bfseries\small,
  fontupper=\small,
  boxrule=0.5pt,
  arc=1mm,
  left=1mm,right=1mm,top=2.0413985pt,bottom=1.0919962pt,
  toptitle=0.4mm,bottomtitle=0.4mm,
  before title=\vphantom{System},after title=\vphantom{System},before skip=0.5899237pt,after skip=0.5899237pt
]
\textbf{Role:} You are a helpful assistant.
\end{tcolorbox}

\begin{tcolorbox}[
  unbreakable,
  colback=white,
  colframe=FigBlue!70!FigInk,
  colbacktitle=FigBlue!20,
  coltitle=FigBlueDark,
  title={User},
  fonttitle=\bfseries\small,
  fontupper=\small,
  boxrule=0.5pt,
  arc=1mm,
  left=1mm,right=1mm,top=2.0408465pt,bottom=1.1369962pt,
  toptitle=0.4mm,bottomtitle=0.4mm,
  before title=\vphantom{System},after title=\vphantom{System},before skip=4.4338071pt,after skip=4.4338071pt
]
\textbf{Input:} The City is Hyderabad. The Year is 2022. The Population\_Under\_Poverty is 29.1.\\
The Malnutrition\_Rate is 15.1. The Food\_Insecurity is 21.2. The Access\_to\_Clean\_Water is 83.5.\\
The Food\_Production\_Index is 96.5.

\vspace{3.5407876pt}
\textbf{Task:} {\color{FigPeachDark}\textbf{What is the percentage of children who are underweight in that city?}}

\vspace{3.4327876pt}
\textbf{Output Format:} Return the final answer in {\color{FigPurpleDark}\texttt{<answer> </answer>}} tags.
\end{tcolorbox}
\end{tcolorbox}


\end{minipage}
\par

\noindent\begin{minipage}{\linewidth}
\subsection{Text Template for TabXLLM}
\label{appendix:text template tabr1}
In this subsection, we introduce the detailed design of the serialization template.
The template is designed to convert tabular data into structured textual inputs that can be effectively processed by the LLMs.
Each feature--value pair in the table is expressed as a short, human-readable clause (e.g., \textit{``The [feature] is [value].''}) and concatenated in a fixed column order to form a complete instance representation.
The prompt consists of two components: (1) a system instruction that defines the model's role as a tabular prediction expert and guides its internal reasoning, and (2) a user message that includes the serialized feature descriptions and a task-specific prediction question.

We further wrap the model's reasoning and final prediction with $<think>$ and $<answer>$ tags, respectively, to clearly separate the reasoning process from the final output.
This design enables TabXLLM to perform effective reasoning over tabular data and provides verifiable reward signals for reinforcement learning.
It serves as the data foundation of PRPO, facilitating TabXLLM's learning of tabular reasoning patterns. The text templates for classification and regression tasks are shown below.
\begin{tcolorbox}[
  enhanced,
  unbreakable,
  before skip=2pt,
  colback=FigBlue!8,
  colframe=FigBlueDark,
  colbacktitle=FigBlueDark,
  coltitle=white,
  title={Example of the text template used for a classification task in TabXLLM},
  fonttitle=\bfseries\small,
  fontupper=\small,
  boxrule=0.9pt,
  arc=2mm,
  left=1.2mm,right=1.2mm,top=3.2021878pt,bottom=3.2035328pt,
  toptitle=0.4mm,bottomtitle=0.4mm,
  before title={\rule[-3.5pt]{0pt}{12pt}},after title={\rule[-3.5pt]{0pt}{12pt}}
]
\begin{tcolorbox}[
  unbreakable,
  colback=white,
  colframe=FigBlue!70!FigInk,
  colbacktitle=FigBlue!20,
  coltitle=FigBlueDark,
  title={System},
  fonttitle=\bfseries\small,
  fontupper=\small,
  boxrule=0.5pt,
  arc=1mm,
  left=1mm,right=1mm,top=2.5656449pt,bottom=1.6163731pt,
  toptitle=0.4mm,bottomtitle=0.4mm,
  before title=\vphantom{System},after title=\vphantom{System},before skip=1.3088719pt,after skip=1.3088719pt
]
\textbf{Role:} You are an expert in tabular prediction. When addressing tabular prediction tasks, you first conduct analysis and reasoning in your mind before presenting the prediction results to the user.
\end{tcolorbox}

\begin{tcolorbox}[
  unbreakable,
  colback=white,
  colframe=FigBlue!70!FigInk,
  colbacktitle=FigBlue!20,
  coltitle=FigBlueDark,
  title={User},
  fonttitle=\bfseries\small,
  fontupper=\small,
  boxrule=0.5pt,
  arc=1mm,
  left=1mm,right=1mm,top=2.5649322pt,bottom=1.6613731pt,
  toptitle=0.4mm,bottomtitle=0.4mm,
  before title=\vphantom{System},after title=\vphantom{System},before skip=4.9584974pt,after skip=4.9584974pt
]
\textbf{Input:} The Married is Yes. The Education is Graduate. The LoanAmount is 44.0. The CoapplicantIncome is 0.0.\\
The Gender is Male. The ApplicantIncome is 5649. The Property\_Area is Urban. The Dependents is 1.\\
The Self\_Employed is No. The Loan\_Amount\_Term is 360.0. The Credit\_History is 1.0.

\vspace{4.0651625pt}
\textbf{Task:} {\color{FigPeachDark}\textbf{Will the applicant's loan be approved? yes or no?}}

\vspace{3.9571524pt}
\textbf{Output Format:} Show your analysis and reasoning in {\color{FigPurpleDark}\texttt{<think> </think>}} tags. And return the final answer in {\color{FigPurpleDark}\texttt{<answer> </answer>}} tags.
\end{tcolorbox}
\end{tcolorbox}

\begin{tcolorbox}[
  enhanced,
  unbreakable,
  before skip=2pt,
  colback=FigBlue!8,
  colframe=FigBlueDark,
  colbacktitle=FigBlueDark,
  coltitle=white,
  title={Example of the text template used for a regression task in TabXLLM},
  fonttitle=\bfseries\small,
  fontupper=\small,
  boxrule=0.9pt,
  arc=2mm,
  left=1.2mm,right=1.2mm,top=3.2023027pt,bottom=3.2035172pt,
  toptitle=0.4mm,bottomtitle=0.4mm,
  before title={\rule[-3.5pt]{0pt}{12pt}},after title={\rule[-3.5pt]{0pt}{12pt}}
]
\begin{tcolorbox}[
  unbreakable,
  colback=white,
  colframe=FigBlue!70!FigInk,
  colbacktitle=FigBlue!20,
  coltitle=FigBlueDark,
  title={System},
  fonttitle=\bfseries\small,
  fontupper=\small,
  boxrule=0.5pt,
  arc=1mm,
  left=1mm,right=1mm,top=2.5656293pt,bottom=1.6163575pt,
  toptitle=0.4mm,bottomtitle=0.4mm,
  before title=\vphantom{System},after title=\vphantom{System},before skip=1.3088719pt,after skip=1.3088719pt
]
\textbf{Role:} You are an expert in tabular prediction. When addressing tabular prediction tasks, you first conduct analysis and reasoning in your mind before presenting the prediction results to the user.
\end{tcolorbox}

\begin{tcolorbox}[
  unbreakable,
  colback=white,
  colframe=FigBlue!70!FigInk,
  colbacktitle=FigBlue!20,
  coltitle=FigBlueDark,
  title={User},
  fonttitle=\bfseries\small,
  fontupper=\small,
  boxrule=0.5pt,
  arc=1mm,
  left=1mm,right=1mm,top=2.5650471pt,bottom=1.6613575pt,
  toptitle=0.4mm,bottomtitle=0.4mm,
  before title=\vphantom{System},after title=\vphantom{System},before skip=4.9583513pt,after skip=4.9583513pt
]
\textbf{Input:} The City is Hyderabad. The Year is 2022. The Population\_Under\_Poverty is 29.1.\\
The Malnutrition\_Rate is 15.1. The Food\_Insecurity is 21.2. The Access\_to\_Clean\_Water is 83.5.\\
The Food\_Production\_Index is 96.5.

\vspace{4.0651468pt}
\textbf{Task:} {\color{FigPeachDark}\textbf{What is the percentage of children who are underweight in that city?}}

\vspace{3.9571468pt}
\textbf{Output Format:} Show your analysis and reasoning in {\color{FigPurpleDark}\texttt{<think> </think>}} tags. And return the final answer in {\color{FigPurpleDark}\texttt{<answer> </answer>}} tags.
\end{tcolorbox}
\end{tcolorbox}


\vspace{4pt}

\end{minipage}
\par

\endgroup

\section{More Results}
\label{appendix:full_tables}

\subsection{Math Benchmark Diagnostics}
\label{appendix:math_diagnostics}
Evaluating on 10 math benchmarks~\citep{cobbe2021training,hendrycks2021measuring,shao2024deepseekmath,patel2021are,amini2019mathqa,ling2017program,wang2024mmlupro,chen2023theoremqa} without math-specific fine-tuning, TabXLLM improves over Qwen3-8B-Base by 6.80 points on average, as summarized in \Cref{table:math_benchmark}.

\begin{table}[!htbp]
\centering
\caption{Performance comparison on mathematical reasoning benchmarks. TabXLLM is trained only on tabular data with PRPO and shows consistent improvements over the base model, suggesting that the post-training effect is not limited to table-format adaptation.}
\label{table:math_benchmark}
\begin{tabular}{lccc}
\toprule
Benchmark & Qwen3-8B-Base & TabXLLM-8B & $\Delta$ \\
\midrule
\rowcolor{FigBlue!10} \multicolumn{4}{l}{\textit{Grade School Math}} \\
GSM8K & 82.11 & \textbf{90.30} & \textcolor{GainBlue}{$\uparrow$8.19} \\
SVAMP & 79.33 & \textbf{91.33} & \textcolor{GainBlue}{$\uparrow$12.00} \\
\midrule
\rowcolor{FigBlue!10} \multicolumn{4}{l}{\textit{Competition Math}} \\
MATH & 49.42 & \textbf{56.30} & \textcolor{GainBlue}{$\uparrow$6.88} \\
AIME & 2.22 & \textbf{7.78} & \textcolor{GainBlue}{$\uparrow$5.56} \\
AMC & 30.12 & \textbf{45.78} & \textcolor{GainBlue}{$\uparrow$15.66} \\
\midrule
\rowcolor{FigBlue!10} \multicolumn{4}{l}{\textit{Math Word Problems}} \\
AQuA & 44.88 & \textbf{53.15} & \textcolor{GainBlue}{$\uparrow$8.27} \\
MathQA & 46.10 & \textbf{48.27} & \textcolor{GainBlue}{$\uparrow$2.17} \\
\midrule
\rowcolor{FigBlue!10} \multicolumn{4}{l}{\textit{Advanced Math}} \\
MMLU-Pro Math & 35.75 & \textbf{38.05} & \textcolor{GainBlue}{$\uparrow$2.30} \\
TheoremQA & 23.50 & \textbf{28.12} & \textcolor{GainBlue}{$\uparrow$4.62} \\
NuminaMath-CoT & 32.80 & \textbf{35.10} & \textcolor{GainBlue}{$\uparrow$2.30} \\
\midrule
\textbf{Average} & 42.62 & \textbf{49.42} & \textcolor{GainBlue}{$\uparrow$6.80} \\
\bottomrule
\end{tabular}
\end{table}

The math benchmark evaluation uses the same prompt for Qwen3-8B-Base and TabXLLM, and the prompt already asks both models to solve each problem step by step. Thus, the comparison controls the answer format to a large extent. We further inspect representative cases where the two models produce similar solution structures and differ mainly in numerical interpretation or computation.

One representative example is the combinatorics problem asking for the sum of the elements of all 15 two-element subsets of $\{1,2,3,4,5,6\}$. Both models identify that each element appears in exactly five subsets, but the base model incorrectly computes $5(1+2+3+4+5+6)$ as 90, while TabXLLM computes $5 \times 21 = 105$. This type of case suggests that the math benchmark gains are not explained solely by a more structured answer format.

\subsection{Full Quantitative Results}
\label{appendix:full_quantitative_results}
Complete per-dataset results for the original benchmark comparison are detailed in \Cref{full_table:classification_fully_trained}, \Cref{full_table:classification_few_zero}, \Cref{full_table:regression_fully_trained} and \Cref{full_table:regression_few_zero}.

\definecolor{AppStripe}{HTML}{FBFDFF}
\definecolor{AppSummary}{HTML}{F0F8FD}
\definecolor{AppAccent}{HTML}{337BAA}
\definecolor{AppRankPeak}{HTML}{69B7EE}
\definecolor{AppRankOne}{HTML}{96CDF3}
\definecolor{AppRankTwo}{HTML}{B4DBF6}
\definecolor{AppRankThree}{HTML}{CFE8FA}
\definecolor{AppRankFour}{HTML}{E6F3FC}
\definecolor{AppRankFive}{HTML}{F4FAFE}

\noindent\textbf{Fully trained regression.} Lower NMAE indicates better performance.
\begin{table}[H]
\centering
\caption{Per-dataset NMAE ($\downarrow$) comparison for the original benchmark methods on regression datasets under the fully trained setting. Darker blue cells indicate lower NMAE. General-purpose LLM baselines are evaluated with raw zero-shot inference, whereas TabXLLM is trained on each target training set with PRPO.}
\label{full_table:regression_fully_trained}
\begingroup
\fontsize{7.053401}{7.758742}\selectfont
\setlength{\tabcolsep}{1.481214pt}
\renewcommand{\arraystretch}{1.32}
\setlength{\heavyrulewidth}{0.564272pt}
\setlength{\lightrulewidth}{0.352670pt}
\setlength{\aboverulesep}{1.214723pt}
\setlength{\belowrulesep}{1.973944pt}
\begin{tabular}{lcccccccccc}
\toprule
\textbf{dataset} & \textbf{Qwen2.5-7B} & \textbf{Qwen2.5-32B} & \textbf{Qwen3-8B} & \textbf{Qwen3-32B} & \textbf{Qwen3-235B} & \textbf{DS-R1} & \textbf{DS-0528} & \textbf{XGBoost} & \textbf{TabPFN} & \textcolor{AppAccent}{\textbf{TabXLLM-8B}} \\
\midrule

\rowcolors{2}{AppStripe}{white}
1000 & 0.0367 & \cellcolor{AppRankOne} 0.0284 & \cellcolor{AppRankFour} 0.0350 & 0.0537 & 0.0495 & 0.0597 & 0.0555 & \cellcolor{AppRankThree} 0.0335 & 0.0651 & \cellcolor{AppRankTwo} 0.0325 \\
AAPL & 0.0105 & \cellcolor{AppRankThree} 0.0105 & \cellcolor{AppRankThree} 0.0105 & \cellcolor{AppRankThree} 0.0105 & \cellcolor{AppRankThree} 0.0105 & \cellcolor{AppRankFour} 0.0105 & 0.0111 & 0.0203 & \cellcolor{AppRankOne} 0.0018 & \cellcolor{AppRankTwo} 0.0090 \\
Carpe & 0.1591 & 4.3618 & 0.1507 & \cellcolor{AppRankFour} 0.1495 & 0.1586 & 0.1508 & 0.1556 & \cellcolor{AppRankTwo} 0.1217 & \cellcolor{AppRankOne} 0.1133 & \cellcolor{AppRankThree} 0.1314 \\
chole & 0.4430 & \cellcolor{AppRankTwo} 0.1181 & 0.3795 & 0.2358 & 0.1576 & 0.1541 & 0.6597 & \cellcolor{AppRankThree} 0.1461 & \cellcolor{AppRankFour} 0.1467 & \cellcolor{AppRankOne} 0.1083 \\
coffe & 0.2138 & 0.2363 & \cellcolor{AppRankFour} 0.2087 & 0.2397 & 0.2745 & 0.4086 & 0.3108 & \cellcolor{AppRankTwo} 0.0016 & \cellcolor{AppRankThree} 0.0066 & \cellcolor{AppRankOne} 0.0000 \\
discl & \cellcolor{AppRankOne} 0.1189 & -- & 0.6316 & 0.6437 & \cellcolor{AppRankThree} 0.1668 & 0.6665 & 0.6249 & 0.2587 & \cellcolor{AppRankTwo} 0.1314 & \cellcolor{AppRankFour} 0.1705 \\
e1684 & 0.0018 & 0.0060 & \cellcolor{AppRankFour} 0.0013 & \cellcolor{AppRankTwo} 0.0001 & \cellcolor{AppRankThree} 0.0006 & \cellcolor{AppRankOne} 0.0000 & 0.0032 & 0.1797 & 0.1257 & 0.3468 \\
Goodr & 0.3344 & \cellcolor{AppRankFour} 0.1908 & 0.4979 & 0.4742 & 0.4872 & 0.4248 & 0.4379 & \cellcolor{AppRankOne} 0.0632 & \cellcolor{AppRankTwo} 0.0698 & \cellcolor{AppRankThree} 0.0736 \\
MIP & 0.5528 & 0.3731 & 0.4850 & 0.5280 & 0.5592 & 0.6728 & \cellcolor{AppRankFour} 0.3705 & \cellcolor{AppRankThree} 0.2786 & \cellcolor{AppRankOne} 0.1810 & \cellcolor{AppRankTwo} 0.2136 \\
munic & 1.7735 & 0.0017 & \cellcolor{AppRankThree} 0.0000 & \cellcolor{AppRankFour} 0.0000 & \cellcolor{AppRankOne} 0.0000 & \cellcolor{AppRankTwo} 0.0000 & 0.0003 & 0.0085 & 0.0009 & 0.0001 \\
nwtco & 0.3164 & 0.2903 & 0.3260 & 0.2833 & 0.2805 & \cellcolor{AppRankFour} 0.2425 & 0.2739 & \cellcolor{AppRankTwo} 0.2085 & \cellcolor{AppRankOne} 0.1890 & \cellcolor{AppRankThree} 0.2262 \\
pakis & 1.2080 & \cellcolor{AppRankFour} 1.1707 & 1.2080 & 1.3703 & 1.2051 & 1.2080 & 1.2080 & \cellcolor{AppRankThree} 0.4480 & \cellcolor{AppRankOne} 0.2503 & \cellcolor{AppRankTwo} 0.2606 \\
plasm & \cellcolor{AppRankFour} 0.2154 & 0.2187 & 0.2588 & 0.2505 & 0.2420 & 0.2216 & 0.2434 & \cellcolor{AppRankThree} 0.1762 & \cellcolor{AppRankOne} 0.1447 & \cellcolor{AppRankTwo} 0.1603 \\
tempe & 0.4185 & \cellcolor{AppRankThree} 0.3416 & 0.3671 & 0.3621 & \cellcolor{AppRankFour} 0.3459 & 0.3547 & 0.3767 & 0.3683 & \cellcolor{AppRankOne} 0.2564 & \cellcolor{AppRankTwo} 0.2623 \\
tumor & \cellcolor{AppRankThree} 0.2219 & 0.2285 & 0.2690 & 0.2425 & 0.2353 & \cellcolor{AppRankFour} 0.2262 & 0.2738 & \cellcolor{AppRankTwo} 0.2096 & \cellcolor{AppRankOne} 0.1708 & 0.2528 \\
\midrule
\rowcolor{AppSummary}
\textbf{Mean} & \textbf{0.4016} & \textbf{0.5412} & \textbf{0.3219} & \textbf{0.3229} & \cellcolor{AppRankFour}\textbf{0.2782} & \textbf{0.3201} & \textbf{0.3337} & \cellcolor{AppRankThree}\textbf{0.1682} & \cellcolor{AppRankOne}\textbf{0.1236} & \cellcolor{AppRankTwo}\textbf{0.1499} \\
\rowcolor{AppSummary}
\textbf{Mean Rank} & \textbf{6.27} & \cellcolor{AppRankFour}\textbf{5.07} & \textbf{6.40} & \textbf{6.20} & \textbf{5.60} & \textbf{5.80} & \textbf{7.13} & \cellcolor{AppRankThree}\textbf{4.13} & \cellcolor{AppRankOne}\textbf{2.93} & \cellcolor{AppRankTwo}\textbf{3.33} \\
\bottomrule
\end{tabular}
\endgroup
\end{table}

\noindent\textbf{Fully trained classification.} Higher accuracy indicates better performance.
\begin{table}[H]
\centering
\caption{Per-dataset accuracy ($\uparrow$) comparison for the original benchmark methods on classification datasets under the fully trained setting. Darker blue cells indicate higher accuracy. General-purpose LLM baselines are evaluated with raw zero-shot inference, whereas TabXLLM is trained on each target training set with PRPO.}
\label{full_table:classification_fully_trained}
\begingroup
\fontsize{7.5}{9.0}\selectfont
\setlength{\tabcolsep}{1.1pt}
\renewcommand{\arraystretch}{1.15}
\begin{tabular}{@{}lcccccccccc@{}}
\toprule
\textbf{Dataset} & \textbf{Qwen2.5-7B} & \textbf{Qwen2.5-32B} & \textbf{Qwen3-8B} & \textbf{Qwen3-32B} & \textbf{Qwen3-235B} & \textbf{DS-R1} & \textbf{DS-0528} & \textbf{XGBoost} & \textbf{TabPFN} & \textcolor{AppAccent}{\textbf{TabXLLM-8B}} \\
\midrule

\rowcolors{2}{AppStripe}{white}
ada & 0.7330 & 0.7943 & 0.7856 & \cellcolor{AppRankFour} 0.8118 & 0.7615 & \cellcolor{AppRankThree} 0.8249 & 0.8009 & \cellcolor{AppRankTwo} 0.8313 & \cellcolor{AppRankOne} 0.8553 & 0.7834 \\
Amazo & 0.0600 & 0.0600 & 0.0850 & \cellcolor{AppRankFour} 0.3650 & 0.2400 & \cellcolor{AppRankThree} 0.9200 & 0.0700 & \cellcolor{AppRankOne} 0.9425 & \cellcolor{AppRankOne} 0.9425 & \cellcolor{AppRankTwo} 0.9400 \\
arsen & \cellcolor{AppRankOne} 0.8750 & \cellcolor{AppRankThree} 0.8571 & \cellcolor{AppRankFour} 0.8036 & 0.6071 & 0.4107 & 0.0714 & 0.7500 & \cellcolor{AppRankThree} 0.8571 & \cellcolor{AppRankTwo} 0.8723 & \cellcolor{AppRankThree} 0.8571 \\
art & \cellcolor{AppRankOne} 0.9678 & \cellcolor{AppRankOne} 0.9678 & \cellcolor{AppRankOne} 0.9678 & \cellcolor{AppRankOne} 0.9678 & \cellcolor{AppRankFour} 0.9356 & 0.8663 & \cellcolor{AppRankThree} 0.9604 & \cellcolor{AppRankTwo} 0.9678 & \cellcolor{AppRankTwo} 0.9678 & \cellcolor{AppRankOne} 0.9678 \\
AVIDa & \cellcolor{AppRankOne} 0.8021 & \cellcolor{AppRankOne} 0.8021 & \cellcolor{AppRankOne} 0.8021 & \cellcolor{AppRankOne} 0.8021 & \cellcolor{AppRankThree} 0.7995 & \cellcolor{AppRankFour} 0.6791 & \cellcolor{AppRankOne} 0.8021 & \cellcolor{AppRankTwo} 0.8019 & \cellcolor{AppRankTwo} 0.8019 & \cellcolor{AppRankOne} 0.8021 \\
blood & 0.4000 & 0.6133 & 0.4400 & \cellcolor{AppRankThree} 0.6800 & \cellcolor{AppRankOne} 0.7600 & \cellcolor{AppRankFour} 0.6667 & 0.4400 & \cellcolor{AppRankOne} 0.7600 & \cellcolor{AppRankTwo} 0.7533 & \cellcolor{AppRankOne} 0.7600 \\
breas & 0.5172 & 0.5517 & 0.5862 & \cellcolor{AppRankFour} 0.6552 & 0.5172 & 0.6207 & 0.5517 & \cellcolor{AppRankThree} 0.7069 & \cellcolor{AppRankTwo} 0.7500 & \cellcolor{AppRankOne} 0.7586 \\
brstw & 0.9143 & 0.9143 & 0.8714 & \cellcolor{AppRankOne} 0.9857 & \cellcolor{AppRankTwo} 0.9714 & \cellcolor{AppRankFour} 0.9571 & 0.8143 & 0.9429 & \cellcolor{AppRankThree} 0.9614 & \cellcolor{AppRankFour} 0.9571 \\
bwin & 0.5283 & \cellcolor{AppRankFour} 0.5849 & 0.3585 & 0.3962 & 0.3962 & 0.3774 & 0.4717 & \cellcolor{AppRankTwo} 0.6698 & \cellcolor{AppRankOne} 0.6708 & \cellcolor{AppRankThree} 0.6604 \\
Click & 0.3750 & \cellcolor{AppRankThree} 0.8300 & \cellcolor{AppRankFour} 0.8050 & 0.4800 & 0.6050 & 0.5900 & 0.8000 & \cellcolor{AppRankOne} 0.8325 & \cellcolor{AppRankTwo} 0.8305 & \cellcolor{AppRankThree} 0.8300 \\
coil2 & \cellcolor{AppRankTwo} 0.9400 & 0.8210 & \cellcolor{AppRankThree} 0.9379 & \cellcolor{AppRankFour} 0.9247 & 0.9217 & 0.8891 & 0.8566 & \cellcolor{AppRankOne} 0.9405 & \cellcolor{AppRankOne} 0.9405 & \cellcolor{AppRankTwo} 0.9400 \\
confe & 0.4000 & 0.3600 & \cellcolor{AppRankThree} 0.6400 & 0.3600 & \cellcolor{AppRankFour} 0.5600 & \cellcolor{AppRankTwo} 0.8000 & 0.4800 & \cellcolor{AppRankOne} 0.8800 & \cellcolor{AppRankOne} 0.8800 & \cellcolor{AppRankOne} 0.8800 \\
datas & 0.4200 & 0.4400 & 0.3200 & 0.3900 & 0.4900 & \cellcolor{AppRankFour} 0.5200 & 0.3200 & \cellcolor{AppRankThree} 0.7000 & \cellcolor{AppRankTwo} 0.7045 & \cellcolor{AppRankOne} 0.7100 \\
datat & \cellcolor{AppRankThree} 0.8462 & \cellcolor{AppRankOne} 0.9231 & \cellcolor{AppRankThree} 0.8462 & 0.0769 & 0.5385 & 0.0769 & \cellcolor{AppRankFour} 0.7692 & \cellcolor{AppRankOne} 0.9231 & \cellcolor{AppRankTwo} 0.9115 & \cellcolor{AppRankOne} 0.9231 \\
depre & \cellcolor{AppRankOne} 0.8322 & \cellcolor{AppRankTwo} 0.7762 & \cellcolor{AppRankFour} 0.3007 & 0.2378 & 0.2797 & 0.2028 & \cellcolor{AppRankThree} 0.4336 & \cellcolor{AppRankOne} 0.8322 & \cellcolor{AppRankOne} 0.8322 & \cellcolor{AppRankOne} 0.8322 \\
dis & 0.9788 & \cellcolor{AppRankFour} 0.9815 & \cellcolor{AppRankFour} 0.9815 & 0.9762 & \cellcolor{AppRankFour} 0.9815 & 0.9762 & 0.9709 & \cellcolor{AppRankThree} 0.9841 & \cellcolor{AppRankOne} 0.9917 & \cellcolor{AppRankTwo} 0.9841 \\
doa & \cellcolor{AppRankThree} 0.5849 & 0.4151 & 0.4906 & 0.5283 & 0.4528 & 0.3585 & \cellcolor{AppRankFour} 0.5660 & \cellcolor{AppRankOne} 0.6604 & \cellcolor{AppRankTwo} 0.6321 & \cellcolor{AppRankOne} 0.6604 \\
Fraud & \cellcolor{AppRankThree} 0.8966 & \cellcolor{AppRankFour} 0.8942 & 0.7260 & 0.6635 & 0.4014 & 0.1971 & 0.3341 & \cellcolor{AppRankTwo} 0.8990 & \cellcolor{AppRankOne} 0.9012 & \cellcolor{AppRankTwo} 0.8990 \\
haber & 0.1935 & \cellcolor{AppRankThree} 0.6452 & 0.5484 & \cellcolor{AppRankFour} 0.5806 & \cellcolor{AppRankThree} 0.6452 & \cellcolor{AppRankFour} 0.5806 & 0.5484 & \cellcolor{AppRankTwo} 0.7419 & \cellcolor{AppRankOne} 0.7484 & \cellcolor{AppRankTwo} 0.7419 \\
imdb & 0.7875 & \cellcolor{AppRankThree} 0.8375 & 0.7875 & \cellcolor{AppRankTwo} 0.8500 & \cellcolor{AppRankOne} 0.8625 & \cellcolor{AppRankThree} 0.8375 & 0.8000 & 0.7500 & 0.7556 & \cellcolor{AppRankFour} 0.8250 \\
ipums & 0.4766 & 0.6061 & 0.7210 & 0.6262 & \cellcolor{AppRankFour} 0.7784 & 0.7063 & 0.4806 & \cellcolor{AppRankTwo} 0.8945 & \cellcolor{AppRankOne} 0.8951 & \cellcolor{AppRankThree} 0.8825 \\
iris & \cellcolor{AppRankOne} 1.0000 & \cellcolor{AppRankTwo} 0.9333 & \cellcolor{AppRankOne} 1.0000 & \cellcolor{AppRankOne} 1.0000 & \cellcolor{AppRankOne} 1.0000 & \cellcolor{AppRankOne} 1.0000 & \cellcolor{AppRankOne} 1.0000 & \cellcolor{AppRankThree} 0.6667 & \cellcolor{AppRankOne} 1.0000 & \cellcolor{AppRankTwo} 0.9333 \\
irish & 0.5600 & \cellcolor{AppRankThree} 0.9400 & \cellcolor{AppRankFour} 0.9000 & \cellcolor{AppRankOne} 1.0000 & \cellcolor{AppRankOne} 1.0000 & \cellcolor{AppRankOne} 1.0000 & 0.8400 & \cellcolor{AppRankOne} 1.0000 & \cellcolor{AppRankTwo} 0.9910 & \cellcolor{AppRankOne} 1.0000 \\
kc1 & 0.6445 & 0.5213 & 0.3886 & \cellcolor{AppRankFour} 0.7393 & 0.7346 & 0.6398 & 0.5592 & \cellcolor{AppRankTwo} 0.8460 & \cellcolor{AppRankOne} 0.8507 & \cellcolor{AppRankThree} 0.8436 \\
kc3 & 0.8261 & \cellcolor{AppRankOne} 0.9130 & \cellcolor{AppRankThree} 0.8913 & 0.7391 & 0.6739 & 0.5870 & 0.8696 & \cellcolor{AppRankTwo} 0.9022 & \cellcolor{AppRankFour} 0.8804 & \cellcolor{AppRankOne} 0.9130 \\
kick & \cellcolor{AppRankFour} 0.7850 & \cellcolor{AppRankThree} 0.8650 & 0.6850 & 0.5550 & 0.1750 & 0.1250 & 0.4950 & \cellcolor{AppRankOne} 0.8775 & \cellcolor{AppRankOne} 0.8775 & \cellcolor{AppRankTwo} 0.8750 \\
Loan & \cellcolor{AppRankFour} 0.7742 & \cellcolor{AppRankOne} 0.8226 & \cellcolor{AppRankThree} 0.7903 & \cellcolor{AppRankTwo} 0.8065 & \cellcolor{AppRankThree} 0.7903 & \cellcolor{AppRankThree} 0.7903 & 0.7097 & 0.6911 & 0.7724 & \cellcolor{AppRankOne} 0.8226 \\
Marke & 0.6473 & 0.7366 & 0.7679 & \cellcolor{AppRankFour} 0.7946 & 0.7723 & 0.6429 & 0.7902 & \cellcolor{AppRankThree} 0.8504 & \cellcolor{AppRankOne} 0.8891 & \cellcolor{AppRankTwo} 0.8750 \\
mc1 & 0.6610 & 0.7360 & 0.7276 & 0.6589 & 0.6051 & 0.8353 & \cellcolor{AppRankFour} 0.8680 & \cellcolor{AppRankTwo} 0.9926 & \cellcolor{AppRankOne} 0.9947 & \cellcolor{AppRankThree} 0.9916 \\
mc2 & \cellcolor{AppRankOne} 0.8235 & \cellcolor{AppRankFour} 0.6471 & 0.5294 & 0.5294 & 0.5294 & 0.5882 & 0.5882 & \cellcolor{AppRankThree} 0.6667 & 0.6061 & \cellcolor{AppRankTwo} 0.7059 \\
meta & 0.5472 & 0.5094 & \cellcolor{AppRankFour} 0.7925 & 0.6415 & 0.5472 & 0.0566 & 0.6415 & \cellcolor{AppRankThree} 0.8962 & \cellcolor{AppRankOne} 0.9208 & \cellcolor{AppRankTwo} 0.9057 \\
mw1 & 0.7073 & 0.6341 & 0.7073 & 0.6098 & \cellcolor{AppRankFour} 0.7561 & 0.5610 & \cellcolor{AppRankTwo} 0.9268 & \cellcolor{AppRankThree} 0.9259 & \cellcolor{AppRankOne} 0.9284 & \cellcolor{AppRankTwo} 0.9268 \\
pc1 & 0.7297 & 0.4955 & 0.3694 & 0.5045 & \cellcolor{AppRankFour} 0.7387 & 0.3874 & 0.4144 & \cellcolor{AppRankTwo} 0.9324 & \cellcolor{AppRankOne} 0.9423 & \cellcolor{AppRankThree} 0.9279 \\
pc2 & 0.7943 & \cellcolor{AppRankFour} 0.9893 & 0.8372 & 0.7048 & 0.8354 & 0.6261 & 0.8927 & \cellcolor{AppRankTwo} 0.9955 & \cellcolor{AppRankThree} 0.9954 & \cellcolor{AppRankOne} 0.9964 \\
pc3 & 0.5669 & 0.7261 & \cellcolor{AppRankFour} 0.7452 & 0.4904 & 0.6561 & 0.5032 & 0.6624 & \cellcolor{AppRankThree} 0.8978 & \cellcolor{AppRankOne} 0.8994 & \cellcolor{AppRankTwo} 0.8981 \\
pc4 & 0.6096 & \cellcolor{AppRankFour} 0.6644 & 0.6438 & 0.5068 & \cellcolor{AppRankFour} 0.6644 & 0.5205 & 0.6233 & \cellcolor{AppRankTwo} 0.8801 & \cellcolor{AppRankOne} 0.9199 & \cellcolor{AppRankThree} 0.8767 \\
plasm & 0.4375 & 0.2812 & 0.4688 & 0.4375 & 0.2500 & \cellcolor{AppRankThree} 0.5000 & 0.4375 & \cellcolor{AppRankOne} 0.5714 & \cellcolor{AppRankFour} 0.4762 & \cellcolor{AppRankTwo} 0.5625 \\
polis & \cellcolor{AppRankFour} 0.7377 & \cellcolor{AppRankThree} 0.7614 & 0.3672 & 0.4450 & 0.4129 & 0.5482 & 0.1997 & \cellcolor{AppRankTwo} 0.9306 & \cellcolor{AppRankOne} 0.9657 & \cellcolor{AppRankTwo} 0.9306 \\
polle & \cellcolor{AppRankOne} 0.5091 & 0.4286 & \cellcolor{AppRankThree} 0.4935 & 0.4753 & 0.4831 & 0.3766 & 0.4857 & \cellcolor{AppRankFour} 0.4870 & 0.4688 & \cellcolor{AppRankTwo} 0.5013 \\
profb & 0.0588 & 0.1029 & 0.0000 & 0.0294 & \cellcolor{AppRankFour} 0.2794 & 0.0735 & 0.0000 & \cellcolor{AppRankOne} 0.6741 & \cellcolor{AppRankTwo} 0.6630 & \cellcolor{AppRankThree} 0.6618 \\
quake & 0.5000 & 0.5183 & \cellcolor{AppRankOne} 0.5596 & 0.5229 & 0.5229 & 0.5183 & \cellcolor{AppRankFour} 0.5321 & \cellcolor{AppRankTwo} 0.5550 & \cellcolor{AppRankThree} 0.5532 & 0.5275 \\
regim & \cellcolor{AppRankOne} 1.0000 & \cellcolor{AppRankTwo} 0.9524 & \cellcolor{AppRankOne} 1.0000 & \cellcolor{AppRankFour} 0.9048 & \cellcolor{AppRankTwo} 0.9524 & 0.8571 & \cellcolor{AppRankOne} 1.0000 & 0.8049 & \cellcolor{AppRankThree} 0.9512 & \cellcolor{AppRankOne} 1.0000 \\
seism & 0.5869 & \cellcolor{AppRankFour} 0.6371 & 0.5637 & 0.4942 & 0.6062 & 0.3707 & 0.5830 & \cellcolor{AppRankTwo} 0.9342 & \cellcolor{AppRankThree} 0.9323 & \cellcolor{AppRankOne} 0.9344 \\
sf & \cellcolor{AppRankOne} 0.8800 & 0.8150 & \cellcolor{AppRankThree} 0.8750 & 0.8250 & 0.7600 & 0.4600 & \cellcolor{AppRankFour} 0.8450 & \cellcolor{AppRankTwo} 0.8775 & \cellcolor{AppRankTwo} 0.8775 & \cellcolor{AppRankOne} 0.8800 \\
solar & 0.4019 & 0.2336 & 0.1776 & 0.2056 & 0.3458 & 0.2617 & \cellcolor{AppRankFour} 0.6822 & \cellcolor{AppRankTwo} 0.8271 & \cellcolor{AppRankThree} 0.8037 & \cellcolor{AppRankOne} 0.8318 \\
Speed & 0.6301 & 0.4212 & 0.2780 & \cellcolor{AppRankThree} 0.6885 & 0.6420 & \cellcolor{AppRankFour} 0.6516 & 0.4654 & \cellcolor{AppRankTwo} 0.8353 & \cellcolor{AppRankOne} 0.8648 & \cellcolor{AppRankTwo} 0.8353 \\
taiwa & \cellcolor{AppRankThree} 0.9384 & \cellcolor{AppRankTwo} 0.9677 & 0.0323 & 0.0367 & 0.0367 & \cellcolor{AppRankFour} 0.1349 & 0.0499 & \cellcolor{AppRankTwo} 0.9677 & \cellcolor{AppRankOne} 0.9705 & \cellcolor{AppRankTwo} 0.9677 \\
tic & 0.6146 & 0.9271 & \cellcolor{AppRankTwo} 0.9896 & \cellcolor{AppRankTwo} 0.9896 & \cellcolor{AppRankOne} 1.0000 & \cellcolor{AppRankOne} 1.0000 & \cellcolor{AppRankFour} 0.9792 & 0.7083 & \cellcolor{AppRankThree} 0.9880 & \cellcolor{AppRankOne} 1.0000 \\
wilt & 0.0579 & 0.0661 & 0.0909 & 0.1178 & \cellcolor{AppRankFour} 0.2831 & 0.1756 & \cellcolor{AppRankThree} 0.2955 & \cellcolor{AppRankTwo} 0.9463 & \cellcolor{AppRankOne} 0.9889 & \cellcolor{AppRankTwo} 0.9463 \\
WMO & 0.4851 & 0.4692 & 0.5050 & \cellcolor{AppRankThree} 0.5129 & 0.5109 & \cellcolor{AppRankOne} 0.5308 & 0.4771 & \cellcolor{AppRankFour} 0.5114 & 0.4955 & \cellcolor{AppRankTwo} 0.5149 \\
\midrule
\rowcolor{AppSummary}
\textbf{Mean} & \textbf{0.6409} & \cellcolor{AppRankFour}\textbf{0.6678} & \textbf{0.6196} & \textbf{0.5986} & \textbf{0.6175} & \textbf{0.5608} & \textbf{0.6187} & \cellcolor{AppRankThree}\textbf{0.8234} & \cellcolor{AppRankTwo}\textbf{0.8413} & \cellcolor{AppRankOne}\textbf{0.8436} \\
\textbf{Rank} & \textbf{5.28} & \cellcolor{AppRankFour}\textbf{5.04} & \textbf{5.28} & \textbf{5.34} & \textbf{5.10} & \textbf{5.84} & \textbf{5.50} & \cellcolor{AppRankThree}\textbf{2.48} & \cellcolor{AppRankTwo}\textbf{2.18} & \cellcolor{AppRankOne}\textbf{2.08} \\
\bottomrule
\end{tabular}
\endgroup
\end{table}

\noindent\textbf{Zero-shot and few-shot classification.} Accuracy is reported on the shared unseen tasks.
\begin{table}[H]
\centering
\caption{Per-task accuracy ($\uparrow$) on ZeroTabBench and FewTabBench, with supervised few-shot comparisons. Darker blue cells indicate higher accuracy. General-purpose LLM columns and TabXLLM's zero-shot column report ZeroTabBench results; TabXLLM's 32-shot column reports FewTabBench results. \textcolor{FigBlueDark}{\textbf{TabLLM}}, \textcolor{FigBlueDark}{\textbf{TabPFN}} and \textcolor{FigBlueDark}{\textbf{XGBoost}} use labeled target-task samples for supervised few-shot fitting.}
\label{full_table:classification_few_zero}
\begingroup
\setlength{\tabcolsep}{1.6pt}
\renewcommand{\arraystretch}{1.02}
\resizebox{\textwidth}{!}{
\begin{tabular}{lccccccccccccccccccccc}
\toprule
\multirow{2}{*}{\vspace{-4pt}Dataset} & \multirow{2}{*}{\vspace{-4pt}Qwen-2.5-7B} & \multirow{2}{*}{\vspace{-4pt}Qwen-2.5-32B} & \multirow{2}{*}{\vspace{-4pt}Qwen3-8B} & \multirow{2}{*}{\vspace{-4pt}Qwen3-32B} & \multirow{2}{*}{\vspace{-4pt}Qwen3-235B} & \multirow{2}{*}{\vspace{-4pt}DS-R1} & \multirow{2}{*}{\vspace{-4pt}DS-0528} & \multicolumn{4}{c}{\textbf{TabLLM}} & \multicolumn{4}{c}{\textbf{XGBoost}} & \multicolumn{4}{c}{\textbf{TabPFN}} & \multicolumn{2}{c}{\textcolor{AppAccent}{\textbf{TabXLLM-8B}}} \\
\cmidrule(lr){9-12} \cmidrule(lr){13-16} \cmidrule(lr){17-20} \cmidrule(lr){21-22}
& &  &  &  &  &  &  & \textbf{4 shot} & \textbf{8 shot} & \textbf{16 shot} & \textbf{32 shot} & \textbf{4 shot} & \textbf{8 shot} & \textbf{16 shot} & \textbf{32 shot} & \textbf{4 shot} & \textbf{8 shot} & \textbf{16 shot} & \textbf{32 shot} & \textcolor{AppAccent}{\textbf{ZS transfer}} & \textcolor{AppAccent}{\textbf{32 shot}} \\
\midrule
\rowcolors{2}{AppStripe}{white}
adult & 0.6700 & 0.7350 & 0.7250 & \cellcolor{AppRankThree} 0.7600 & \cellcolor{AppRankFour} 0.7550 & \cellcolor{AppRankOne} 0.7900 & \cellcolor{AppRankThree} 0.7600 & \cellcolor{AppRankThree} 0.7600 & \cellcolor{AppRankThree} 0.7600 & \cellcolor{AppRankThree} 0.7600 & \cellcolor{AppRankTwo} 0.7800 & 0.6100 & 0.6900 & 0.6950 & 0.7100 & 0.4780 & 0.5497 & 0.6355 & 0.7047 & \cellcolor{AppRankTwo} 0.7800 & \cellcolor{AppRankPeak} 0.7950 \\
airli & 0.4400 & 0.3400 & \cellcolor{AppRankPeak} 0.5600 & 0.5250 & 0.5300 & \cellcolor{AppRankThree} 0.5450 & \cellcolor{AppRankTwo} 0.5500 & 0.4450 & 0.4450 & 0.5100 & \cellcolor{AppRankOne} 0.5550 & 0.4500 & 0.5300 & \cellcolor{AppRankFour} 0.5400 & 0.4950 & 0.4795 & 0.5022 & 0.5035 & 0.5082 & \cellcolor{AppRankTwo} 0.5500 & \cellcolor{AppRankOne} 0.5550 \\
analc & 0.1458 & 0.1667 & 0.5833 & 0.5417 & \cellcolor{AppRankTwo} 0.7917 & \cellcolor{AppRankOne} 0.8333 & 0.3542 & \cellcolor{AppRankPeak} 0.8750 & \cellcolor{AppRankPeak} 0.8750 & \cellcolor{AppRankPeak} 0.8750 & \cellcolor{AppRankPeak} 0.8750 & \cellcolor{AppRankOne} 0.8333 & \cellcolor{AppRankFour} 0.7500 & 0.4792 & \cellcolor{AppRankPeak} 0.8750 & 0.5400 & 0.6274 & 0.6747 & \cellcolor{AppRankThree} 0.7874 & 0.6875 & \cellcolor{AppRankPeak} 0.8750 \\
annea & 0.2444 & 0.3778 & 0.4111 & 0.2778 & 0.5556 & 0.4333 & 0.2889 & \cellcolor{AppRankTwo} 0.7667 & \cellcolor{AppRankTwo} 0.7667 & \cellcolor{AppRankTwo} 0.7667 & \cellcolor{AppRankTwo} 0.7667 & \cellcolor{AppRankFour} 0.6889 & 0.4111 & \cellcolor{AppRankOne} 0.7778 & \cellcolor{AppRankPeak} 0.8667 & 0.0000 & 0.0000 & 0.0000 & 0.0000 & \cellcolor{AppRankThree} 0.7556 & \cellcolor{AppRankTwo} 0.7667 \\
arrhy & \cellcolor{AppRankPeak} 0.6739 & 0.5870 & 0.5435 & 0.5435 & 0.5652 & \cellcolor{AppRankFour} 0.6087 & \cellcolor{AppRankOne} 0.6304 & 0.4348 & 0.5870 & 0.5435 & 0.5870 & 0.5652 & 0.4348 & 0.5435 & 0.5000 & 0.5220 & 0.5264 & \cellcolor{AppRankTwo} 0.6231 & \cellcolor{AppRankThree} 0.6220 & 0.5652 & \cellcolor{AppRankOne} 0.6304 \\
autoM & 0.5500 & 0.7000 & 0.7250 & 0.7000 & \cellcolor{AppRankTwo} 0.8500 & 0.8000 & 0.7000 & 0.5250 & 0.5250 & 0.7500 & 0.8000 & 0.5500 & 0.7500 & 0.7500 & \cellcolor{AppRankFour} 0.8250 & 0.7600 & \cellcolor{AppRankThree} 0.8337 & \cellcolor{AppRankOne} 0.8638 & \cellcolor{AppRankPeak} 0.8650 & 0.7750 & \cellcolor{AppRankFour} 0.8250 \\
Bank & \cellcolor{AppRankTwo} 0.7290 & \cellcolor{AppRankOne} 0.7480 & 0.4530 & 0.4050 & 0.4430 & 0.4110 & 0.3870 & \cellcolor{AppRankPeak} 0.7960 & \cellcolor{AppRankPeak} 0.7960 & \cellcolor{AppRankPeak} 0.7960 & \cellcolor{AppRankPeak} 0.7960 & 0.2450 & 0.4410 & 0.5500 & \cellcolor{AppRankThree} 0.6930 & 0.4263 & 0.5611 & 0.5682 & \cellcolor{AppRankFour} 0.6456 & \cellcolor{AppRankPeak} 0.7960 & \cellcolor{AppRankPeak} 0.7960 \\
blast & \cellcolor{AppRankThree} 0.7149 & \cellcolor{AppRankPeak} 0.7674 & 0.6000 & 0.6213 & 0.6099 & 0.6227 & 0.5489 & 0.4000 & \cellcolor{AppRankTwo} 0.7348 & \cellcolor{AppRankTwo} 0.7348 & \cellcolor{AppRankOne} 0.7404 & 0.4780 & 0.4227 & 0.6610 & \cellcolor{AppRankFour} 0.7007 & 0.5386 & 0.5505 & 0.6337 & 0.6971 & \cellcolor{AppRankTwo} 0.7348 & \cellcolor{AppRankTwo} 0.7348 \\
brazi & \cellcolor{AppRankFour} 0.6429 & 0.6190 & 0.3810 & 0.3571 & 0.3333 & 0.5238 & 0.2857 & \cellcolor{AppRankPeak} 0.7619 & \cellcolor{AppRankPeak} 0.7619 & \cellcolor{AppRankPeak} 0.7619 & \cellcolor{AppRankPeak} 0.7619 & \cellcolor{AppRankOne} 0.7381 & \cellcolor{AppRankTwo} 0.7143 & 0.5000 & 0.5000 & 0.5361 & 0.5880 & 0.5747 & \cellcolor{AppRankThree} 0.6723 & \cellcolor{AppRankPeak} 0.7619 & \cellcolor{AppRankPeak} 0.7619 \\
calif & 0.6200 & 0.3650 & 0.6700 & \cellcolor{AppRankFour} 0.7200 & \cellcolor{AppRankFour} 0.7200 & \cellcolor{AppRankOne} 0.7700 & 0.5050 & 0.5000 & 0.5000 & 0.5000 & 0.5000 & 0.5100 & 0.5950 & \cellcolor{AppRankThree} 0.7350 & \cellcolor{AppRankTwo} 0.7450 & 0.5493 & 0.6597 & 0.7027 & \cellcolor{AppRankPeak} 0.7718 & 0.7000 & \cellcolor{AppRankTwo} 0.7450 \\
chole & 0.4839 & 0.4516 & 0.4516 & 0.4516 & 0.4516 & 0.5161 & 0.4516 & 0.4516 & 0.4516 & 0.4516 & 0.5161 & \cellcolor{AppRankOne} 0.5484 & 0.3548 & 0.4194 & 0.4516 & \cellcolor{AppRankFour} 0.5262 & 0.5115 & \cellcolor{AppRankThree} 0.5344 & \cellcolor{AppRankTwo} 0.5410 & 0.5161 & \cellcolor{AppRankPeak} 0.6161 \\
churn & \cellcolor{AppRankTwo} 0.7840 & \cellcolor{AppRankOne} 0.7920 & \cellcolor{AppRankFour} 0.7480 & 0.5680 & 0.6900 & 0.5040 & 0.7340 & 0.1420 & 0.5800 & \cellcolor{AppRankPeak} 0.8580 & \cellcolor{AppRankPeak} 0.8580 & \cellcolor{AppRankThree} 0.7560 & 0.6460 & 0.1800 & 0.6820 & 0.4147 & 0.4433 & 0.5414 & 0.5901 & \cellcolor{AppRankPeak} 0.8580 & \cellcolor{AppRankPeak} 0.8580 \\
cleve & 0.6774 & 0.7097 & 0.4839 & 0.5484 & 0.5806 & 0.4839 & 0.5806 & 0.4516 & \cellcolor{AppRankThree} 0.7742 & 0.5484 & \cellcolor{AppRankOne} 0.8387 & 0.6452 & 0.5806 & \cellcolor{AppRankTwo} 0.8065 & 0.7097 & 0.5852 & 0.6361 & 0.7197 & \cellcolor{AppRankFour} 0.7672 & 0.6774 & \cellcolor{AppRankPeak} 0.8419 \\
colic & 0.3784 & 0.5405 & 0.6216 & 0.6216 & 0.6486 & \cellcolor{AppRankFour} 0.7568 & 0.5946 & 0.6216 & 0.6216 & 0.6216 & \cellcolor{AppRankTwo} 0.8108 & 0.4054 & 0.6486 & 0.7297 & 0.7297 & 0.5284 & 0.7108 & \cellcolor{AppRankThree} 0.7932 & \cellcolor{AppRankOne} 0.8216 & 0.5946 & \cellcolor{AppRankPeak} 0.8297 \\
commu & 0.5950 & 0.6300 & 0.6800 & 0.7000 & 0.6550 & 0.6100 & 0.6000 & 0.5750 & 0.4250 & 0.6550 & 0.6550 & 0.5950 & 0.7300 & \cellcolor{AppRankFour} 0.7600 & \cellcolor{AppRankOne} 0.7900 & 0.5935 & 0.7020 & \cellcolor{AppRankThree} 0.7644 & \cellcolor{AppRankTwo} 0.7875 & 0.6400 & \cellcolor{AppRankPeak} 0.8300 \\
compa & 0.3920 & 0.4792 & 0.5814 & \cellcolor{AppRankTwo} 0.6212 & \cellcolor{AppRankFour} 0.5947 & 0.5852 & \cellcolor{AppRankThree} 0.6023 & 0.4697 & 0.4830 & 0.4811 & 0.5284 & 0.5417 & 0.5303 & 0.4811 & 0.5701 & 0.5203 & 0.5413 & 0.5312 & 0.5937 & \cellcolor{AppRankPeak} 0.6420 & \cellcolor{AppRankOne} 0.6307 \\
dgf & 0.6082 & 0.4561 & 0.8216 & 0.7778 & 0.8129 & \cellcolor{AppRankThree} 0.8450 & 0.7398 & \cellcolor{AppRankFour} 0.8363 & \cellcolor{AppRankFour} 0.8363 & \cellcolor{AppRankFour} 0.8363 & \cellcolor{AppRankFour} 0.8363 & 0.1813 & 0.7368 & \cellcolor{AppRankPeak} 0.8947 & \cellcolor{AppRankTwo} 0.8743 & 0.4220 & 0.7726 & 0.8158 & \cellcolor{AppRankOne} 0.8873 & \cellcolor{AppRankFour} 0.8363 & \cellcolor{AppRankFour} 0.8363 \\
diabe & 0.6234 & \cellcolor{AppRankFour} 0.6623 & \cellcolor{AppRankThree} 0.6753 & 0.5974 & 0.6234 & 0.6234 & 0.6104 & 0.3506 & 0.4675 & 0.5455 & 0.6494 & \cellcolor{AppRankPeak} 0.7662 & 0.6494 & \cellcolor{AppRankThree} 0.6753 & \cellcolor{AppRankTwo} 0.6883 & 0.5630 & 0.6045 & 0.6162 & 0.6474 & \cellcolor{AppRankPeak} 0.7662 & \cellcolor{AppRankOne} 0.7494 \\
diffe & 0.7436 & 0.7949 & \cellcolor{AppRankPeak} 0.9744 & 0.8462 & \cellcolor{AppRankOne} 0.9487 & \cellcolor{AppRankOne} 0.9487 & 0.7436 & 0.7692 & 0.7179 & \cellcolor{AppRankFour} 0.8974 & \cellcolor{AppRankOne} 0.9487 & \cellcolor{AppRankFour} 0.8974 & \cellcolor{AppRankFour} 0.8974 & \cellcolor{AppRankFour} 0.8974 & \cellcolor{AppRankTwo} 0.9231 & 0.8195 & 0.8688 & 0.8818 & \cellcolor{AppRankThree} 0.9195 & \cellcolor{AppRankFour} 0.8974 & \cellcolor{AppRankTwo} 0.9231 \\
Emplo & 0.5408 & \cellcolor{AppRankPeak} 0.6609 & 0.5429 & 0.5880 & 0.5300 & 0.3562 & 0.4571 & 0.3433 & 0.3433 & \cellcolor{AppRankOne} 0.6567 & \cellcolor{AppRankOne} 0.6567 & \cellcolor{AppRankFour} 0.6416 & 0.5837 & 0.5794 & \cellcolor{AppRankTwo} 0.6545 & 0.5814 & 0.6404 & 0.6161 & 0.6291 & \cellcolor{AppRankThree} 0.6502 & \cellcolor{AppRankOne} 0.6567 \\
eye & 0.5007 & 0.5007 & 0.4993 & 0.4665 & 0.4599 & 0.4796 & \cellcolor{AppRankFour} 0.5033 & 0.5007 & 0.5007 & 0.5007 & 0.4993 & \cellcolor{AppRankOne} 0.5138 & \cellcolor{AppRankTwo} 0.5085 & 0.4915 & 0.5020 & 0.5007 & 0.5003 & \cellcolor{AppRankPeak} 0.5168 & \cellcolor{AppRankOne} 0.5138 & \cellcolor{AppRankThree} 0.5059 & 0.4993 \\
flags & \cellcolor{AppRankPeak} 0.7000 & 0.5500 & \cellcolor{AppRankThree} 0.6000 & 0.5000 & 0.4500 & 0.4000 & 0.4500 & 0.3500 & \cellcolor{AppRankOne} 0.6500 & \cellcolor{AppRankOne} 0.6500 & \cellcolor{AppRankOne} 0.6500 & 0.3000 & 0.3000 & 0.4000 & \cellcolor{AppRankPeak} 0.7000 & 0.4282 & 0.5128 & \cellcolor{AppRankFour} 0.5846 & \cellcolor{AppRankTwo} 0.6205 & \cellcolor{AppRankThree} 0.6000 & \cellcolor{AppRankOne} 0.6500 \\
hayes & 0.6429 & 0.5000 & 0.6429 & 0.6429 & 0.6429 & 0.4286 & \cellcolor{AppRankOne} 0.7857 & 0.6429 & 0.6429 & 0.6429 & 0.6429 & 0.3571 & \cellcolor{AppRankTwo} 0.7143 & \cellcolor{AppRankOne} 0.7857 & \cellcolor{AppRankPeak} 0.8571 & 0.5222 & 0.5556 & 0.5963 & \cellcolor{AppRankFour} 0.6667 & 0.6429 & \cellcolor{AppRankThree} 0.6729 \\
hepat & 0.2500 & 0.5000 & 0.4375 & 0.3125 & 0.5000 & 0.5000 & 0.3750 & \cellcolor{AppRankOne} 0.8125 & \cellcolor{AppRankOne} 0.8125 & \cellcolor{AppRankOne} 0.8125 & 0.6875 & 0.5625 & 0.6250 & 0.5625 & 0.6250 & \cellcolor{AppRankTwo} 0.7581 & \cellcolor{AppRankFour} 0.7387 & 0.7323 & \cellcolor{AppRankThree} 0.7484 & \cellcolor{AppRankPeak} 0.8750 & \cellcolor{AppRankPeak} 0.8750 \\
HMEQ & 0.6376 & \cellcolor{AppRankThree} 0.7886 & 0.4262 & 0.5101 & 0.5268 & 0.3792 & 0.7097 & \cellcolor{AppRankOne} 0.8003 & \cellcolor{AppRankOne} 0.8003 & \cellcolor{AppRankOne} 0.8003 & \cellcolor{AppRankOne} 0.8003 & \cellcolor{AppRankTwo} 0.7987 & 0.5923 & 0.5000 & \cellcolor{AppRankFour} 0.7450 & 0.4859 & 0.5800 & 0.6525 & 0.7255 & \cellcolor{AppRankPeak} 0.8020 & \cellcolor{AppRankTwo} 0.7987 \\
hypot & 0.4683 & 0.7063 & 0.4127 & 0.5344 & 0.4524 & 0.3704 & 0.3651 & \cellcolor{AppRankPeak} 0.9233 & \cellcolor{AppRankPeak} 0.9233 & \cellcolor{AppRankPeak} 0.9233 & \cellcolor{AppRankPeak} 0.9233 & \cellcolor{AppRankFour} 0.7593 & 0.5344 & \cellcolor{AppRankTwo} 0.8466 & \cellcolor{AppRankThree} 0.8148 & 0.0000 & 0.0000 & 0.0000 & 0.0000 & 0.6561 & \cellcolor{AppRankOne} 0.9153 \\
ibm & \cellcolor{AppRankTwo} 0.7755 & \cellcolor{AppRankOne} 0.7823 & 0.3401 & 0.3741 & 0.3878 & 0.2585 & 0.3469 & \cellcolor{AppRankPeak} 0.8367 & \cellcolor{AppRankPeak} 0.8367 & \cellcolor{AppRankPeak} 0.8367 & \cellcolor{AppRankPeak} 0.8367 & \cellcolor{AppRankFour} 0.6122 & 0.2585 & 0.6054 & \cellcolor{AppRankThree} 0.7551 & 0.5143 & 0.4810 & 0.5469 & 0.5344 & \cellcolor{AppRankPeak} 0.8367 & \cellcolor{AppRankPeak} 0.8367 \\
jungl & 0.4407 & 0.7415 & 0.6992 & 0.7161 & 0.7288 & 0.7076 & 0.6780 & 0.4407 & 0.4407 & 0.6229 & \cellcolor{AppRankOne} 0.8644 & 0.5932 & 0.6483 & \cellcolor{AppRankFour} 0.7585 & \cellcolor{AppRankTwo} 0.8263 & 0.6257 & 0.6711 & \cellcolor{AppRankThree} 0.7985 & \cellcolor{AppRankPeak} 0.9361 & 0.6398 & 0.7373 \\
kc2 & 0.4340 & 0.3774 & 0.2453 & 0.2642 & 0.5094 & 0.4528 & 0.3208 & \cellcolor{AppRankPeak} 0.7925 & \cellcolor{AppRankPeak} 0.7925 & \cellcolor{AppRankPeak} 0.7925 & \cellcolor{AppRankPeak} 0.7925 & 0.2830 & 0.5283 & 0.6226 & \cellcolor{AppRankTwo} 0.7170 & 0.6210 & \cellcolor{AppRankThree} 0.6714 & \cellcolor{AppRankFour} 0.6381 & \cellcolor{AppRankOne} 0.7562 & \cellcolor{AppRankPeak} 0.7925 & \cellcolor{AppRankPeak} 0.7925 \\
kdd & 0.6456 & 0.6203 & 0.5443 & 0.6076 & 0.6203 & 0.5190 & 0.6076 & 0.4557 & 0.4557 & 0.6456 & 0.6456 & 0.4430 & 0.3671 & \cellcolor{AppRankOne} 0.8481 & \cellcolor{AppRankPeak} 0.8608 & 0.5287 & 0.7102 & \cellcolor{AppRankFour} 0.7879 & \cellcolor{AppRankThree} 0.8293 & 0.5696 & \cellcolor{AppRankTwo} 0.8456 \\
lungc & \cellcolor{AppRankFour} 0.6522 & \cellcolor{AppRankTwo} 0.7826 & 0.3043 & 0.2609 & 0.3043 & 0.1739 & 0.4783 & \cellcolor{AppRankOne} 0.8261 & \cellcolor{AppRankOne} 0.8261 & \cellcolor{AppRankOne} 0.8261 & \cellcolor{AppRankOne} 0.8261 & 0.4783 & \cellcolor{AppRankTwo} 0.7826 & 0.5652 & 0.4783 & 0.4783 & 0.5696 & 0.6239 & \cellcolor{AppRankThree} 0.7196 & \cellcolor{AppRankPeak} 0.8696 & \cellcolor{AppRankOne} 0.8261 \\
NATIC & 0.6427 & 0.6293 & 0.6653 & 0.6467 & 0.6307 & 0.1893 & 0.6547 & 0.6493 & 0.6493 & 0.6493 & 0.6493 & 0.6787 & \cellcolor{AppRankPeak} 0.8640 & \cellcolor{AppRankFour} 0.8120 & \cellcolor{AppRankOne} 0.8587 & 0.5857 & 0.6758 & 0.7800 & \cellcolor{AppRankTwo} 0.8517 & 0.6480 & \cellcolor{AppRankThree} 0.8493 \\
newto & 0.4286 & 0.5000 & 0.5714 & 0.5714 & 0.5000 & 0.4286 & 0.5714 & 0.5000 & 0.3571 & 0.5000 & \cellcolor{AppRankThree} 0.6429 & 0.4286 & 0.5000 & \cellcolor{AppRankOne} 0.7143 & \cellcolor{AppRankThree} 0.6429 & 0.4821 & 0.5571 & \cellcolor{AppRankFour} 0.6107 & \cellcolor{AppRankTwo} 0.6679 & \cellcolor{AppRankThree} 0.6429 & \cellcolor{AppRankPeak} 0.8133 \\
no2 & 0.5000 & 0.4600 & 0.5000 & 0.4800 & 0.4800 & 0.5000 & 0.5000 & 0.5000 & 0.5000 & 0.5000 & \cellcolor{AppRankTwo} 0.6400 & 0.4800 & 0.4400 & 0.5200 & \cellcolor{AppRankOne} 0.6600 & 0.5100 & 0.5110 & \cellcolor{AppRankThree} 0.5280 & \cellcolor{AppRankFour} 0.5210 & 0.5200 & \cellcolor{AppRankPeak} 0.8205 \\
page & 0.5912 & \cellcolor{AppRankTwo} 0.9033 & 0.6642 & 0.7810 & \cellcolor{AppRankThree} 0.8978 & 0.8631 & 0.4380 & \cellcolor{AppRankThree} 0.8978 & \cellcolor{AppRankThree} 0.8978 & \cellcolor{AppRankThree} 0.8978 & \cellcolor{AppRankThree} 0.8978 & 0.2464 & \cellcolor{AppRankFour} 0.8869 & 0.6387 & 0.8066 & 0.7836 & 0.8520 & \cellcolor{AppRankOne} 0.9093 & \cellcolor{AppRankPeak} 0.9310 & 0.8704 & \cellcolor{AppRankThree} 0.8978 \\
phary & 0.5500 & 0.5500 & \cellcolor{AppRankFour} 0.6000 & 0.4500 & 0.3500 & 0.4500 & 0.5000 & \cellcolor{AppRankFour} 0.6000 & \cellcolor{AppRankFour} 0.6000 & \cellcolor{AppRankFour} 0.6000 & \cellcolor{AppRankFour} 0.6000 & \cellcolor{AppRankOne} 0.7500 & \cellcolor{AppRankFour} 0.6000 & \cellcolor{AppRankFour} 0.6000 & 0.5000 & \cellcolor{AppRankThree} 0.6154 & 0.5667 & \cellcolor{AppRankTwo} 0.7385 & \cellcolor{AppRankPeak} 0.7513 & \cellcolor{AppRankFour} 0.6000 & \cellcolor{AppRankFour} 0.6000 \\
pm10 & 0.4800 & 0.5000 & 0.5000 & 0.5000 & 0.5000 & 0.4800 & 0.5000 & 0.5000 & 0.5000 & 0.5000 & 0.5000 & 0.5000 & \cellcolor{AppRankPeak} 0.6200 & \cellcolor{AppRankOne} 0.5600 & 0.5000 & 0.4980 & \cellcolor{AppRankFour} 0.5020 & \cellcolor{AppRankTwo} 0.5200 & \cellcolor{AppRankThree} 0.5140 & 0.5000 & 0.5000 \\
PostP & 0.5298 & 0.6026 & 0.5894 & 0.5894 & 0.5629 & 0.6225 & 0.5894 & 0.6490 & 0.6490 & 0.6490 & \cellcolor{AppRankPeak} 0.7947 & \cellcolor{AppRankFour} 0.7417 & 0.4702 & \cellcolor{AppRankThree} 0.7682 & \cellcolor{AppRankOne} 0.7748 & 0.5934 & 0.6595 & 0.6890 & \cellcolor{AppRankTwo} 0.7684 & 0.6689 & 0.6490 \\
prnn & 0.1000 & 0.2000 & 0.5500 & \cellcolor{AppRankThree} 0.8000 & 0.4000 & 0.5000 & 0.3500 & 0.5000 & 0.5000 & 0.5000 & 0.5500 & 0.5500 & 0.5500 & \cellcolor{AppRankFour} 0.6500 & \cellcolor{AppRankThree} 0.8000 & 0.5175 & \cellcolor{AppRankTwo} 0.8025 & \cellcolor{AppRankOne} 0.9800 & \cellcolor{AppRankPeak} 1.0000 & 0.5500 & 0.5000 \\
road & 0.5350 & 0.3300 & 0.3500 & \cellcolor{AppRankTwo} 0.6400 & 0.5400 & 0.5550 & 0.4000 & 0.5100 & 0.5000 & \cellcolor{AppRankFour} 0.5850 & 0.5450 & 0.5000 & \cellcolor{AppRankTwo} 0.6400 & 0.5300 & \cellcolor{AppRankThree} 0.6300 & 0.5390 & 0.5545 & 0.5577 & 0.5565 & \cellcolor{AppRankOne} 0.6500 & \cellcolor{AppRankPeak} 0.6850 \\
segme & 0.8225 & 0.7186 & 0.4848 & 0.8528 & 0.8442 & 0.8355 & 0.8442 & \cellcolor{AppRankFour} 0.8571 & \cellcolor{AppRankFour} 0.8571 & \cellcolor{AppRankFour} 0.8571 & \cellcolor{AppRankFour} 0.8571 & 0.3247 & 0.6364 & \cellcolor{AppRankPeak} 0.9697 & \cellcolor{AppRankTwo} 0.8918 & 0.5929 & 0.6781 & \cellcolor{AppRankThree} 0.8833 & \cellcolor{AppRankOne} 0.9591 & 0.7922 & \cellcolor{AppRankFour} 0.8571 \\
Skin & 0.7043 & 0.8565 & 0.8652 & 0.8174 & 0.8652 & 0.8870 & 0.8565 & \cellcolor{AppRankOne} 0.9261 & \cellcolor{AppRankOne} 0.9261 & \cellcolor{AppRankOne} 0.9261 & \cellcolor{AppRankFour} 0.8957 & \cellcolor{AppRankTwo} 0.9174 & \cellcolor{AppRankOne} 0.9261 & \cellcolor{AppRankPeak} 0.9435 & \cellcolor{AppRankOne} 0.9261 & 0.6576 & 0.7172 & 0.7593 & 0.8787 & \cellcolor{AppRankTwo} 0.9174 & \cellcolor{AppRankThree} 0.9043 \\
spamb & 0.6052 & 0.7657 & 0.6161 & 0.5748 & 0.6920 & 0.5141 & 0.6703 & 0.6052 & 0.6052 & 0.7570 & \cellcolor{AppRankFour} 0.8156 & 0.6334 & 0.5597 & \cellcolor{AppRankThree} 0.8178 & \cellcolor{AppRankTwo} 0.8330 & 0.5544 & 0.7403 & \cellcolor{AppRankOne} 0.8347 & \cellcolor{AppRankPeak} 0.8510 & 0.7137 & 0.7202 \\
SPECT & \cellcolor{AppRankTwo} 0.7143 & 0.5143 & \cellcolor{AppRankPeak} 0.7429 & 0.6571 & \cellcolor{AppRankTwo} 0.7143 & \cellcolor{AppRankTwo} 0.7143 & \cellcolor{AppRankFour} 0.6857 & \cellcolor{AppRankTwo} 0.7143 & \cellcolor{AppRankTwo} 0.7143 & \cellcolor{AppRankFour} 0.6857 & \cellcolor{AppRankTwo} 0.7143 & 0.6286 & 0.6571 & 0.6571 & 0.6286 & 0.5800 & 0.6814 & \cellcolor{AppRankThree} 0.6986 & \cellcolor{AppRankOne} 0.7271 & \cellcolor{AppRankTwo} 0.7143 & \cellcolor{AppRankTwo} 0.7143 \\
spect & \cellcolor{AppRankOne} 0.8889 & 0.7778 & \cellcolor{AppRankOne} 0.8889 & 0.1852 & 0.6481 & 0.1667 & 0.6111 & \cellcolor{AppRankOne} 0.8889 & \cellcolor{AppRankOne} 0.8889 & \cellcolor{AppRankOne} 0.8889 & \cellcolor{AppRankOne} 0.8889 & 0.1481 & 0.3519 & \cellcolor{AppRankPeak} 0.9259 & \cellcolor{AppRankThree} 0.8704 & 0.5869 & 0.6963 & \cellcolor{AppRankFour} 0.8280 & \cellcolor{AppRankTwo} 0.8738 & 0.7222 & \cellcolor{AppRankOne} 0.8889 \\
telco & 0.6411 & \cellcolor{AppRankPeak} 0.7518 & 0.5702 & 0.5957 & 0.6241 & 0.6071 & 0.5901 & 0.2780 & \cellcolor{AppRankThree} 0.7348 & \cellcolor{AppRankThree} 0.7348 & \cellcolor{AppRankOne} 0.7433 & 0.5021 & 0.4936 & 0.6950 & 0.6624 & 0.5313 & 0.5566 & 0.6241 & \cellcolor{AppRankFour} 0.6977 & \cellcolor{AppRankTwo} 0.7362 & \cellcolor{AppRankThree} 0.7348 \\
Tour & 0.6562 & 0.5833 & 0.3958 & 0.3438 & 0.3854 & 0.3125 & 0.3438 & \cellcolor{AppRankPeak} 0.7604 & \cellcolor{AppRankPeak} 0.7604 & \cellcolor{AppRankPeak} 0.7604 & \cellcolor{AppRankPeak} 0.7604 & \cellcolor{AppRankTwo} 0.7292 & 0.4792 & \cellcolor{AppRankOne} 0.7396 & \cellcolor{AppRankThree} 0.7188 & 0.6236 & 0.6298 & 0.6812 & \cellcolor{AppRankFour} 0.6942 & \cellcolor{AppRankPeak} 0.7604 & \cellcolor{AppRankPeak} 0.7604 \\
triaz & 0.5263 & 0.5263 & 0.4737 & 0.4737 & 0.4211 & 0.4211 & 0.5263 & 0.5789 & 0.3684 & 0.5263 & 0.4737 & 0.5263 & \cellcolor{AppRankTwo} 0.6842 & \cellcolor{AppRankOne} 0.7368 & \cellcolor{AppRankPeak} 0.7895 & 0.4895 & 0.5395 & 0.5605 & \cellcolor{AppRankThree} 0.6500 & \cellcolor{AppRankFour} 0.6316 & 0.5789 \\
Tunin & 0.3750 & 0.3750 & 0.3750 & 0.3125 & 0.3750 & 0.0625 & 0.3750 & \cellcolor{AppRankThree} 0.6250 & \cellcolor{AppRankThree} 0.6250 & \cellcolor{AppRankThree} 0.6250 & \cellcolor{AppRankOne} 0.7500 & \cellcolor{AppRankTwo} 0.6875 & \cellcolor{AppRankThree} 0.6250 & \cellcolor{AppRankThree} 0.6250 & \cellcolor{AppRankPeak} 0.8125 & 0.6156 & 0.5469 & 0.5375 & \cellcolor{AppRankFour} 0.6219 & 0.5625 & \cellcolor{AppRankThree} 0.6250 \\
UCI & \cellcolor{AppRankPeak} 0.8623 & \cellcolor{AppRankFour} 0.7515 & \cellcolor{AppRankThree} 0.7814 & 0.6108 & 0.6617 & 0.3593 & 0.7246 & \cellcolor{AppRankTwo} 0.8563 & \cellcolor{AppRankTwo} 0.8563 & \cellcolor{AppRankTwo} 0.8563 & \cellcolor{AppRankTwo} 0.8563 & 0.1437 & 0.4371 & 0.5808 & 0.7006 & 0.5382 & 0.4976 & 0.5879 & 0.5961 & \cellcolor{AppRankOne} 0.8593 & \cellcolor{AppRankTwo} 0.8563 \\
vehic & 0.7059 & 0.4471 & 0.6353 & 0.5765 & 0.3882 & 0.3059 & 0.5882 & \cellcolor{AppRankFour} 0.7412 & \cellcolor{AppRankFour} 0.7412 & \cellcolor{AppRankFour} 0.7412 & \cellcolor{AppRankFour} 0.7412 & \cellcolor{AppRankFour} 0.7412 & 0.4235 & \cellcolor{AppRankTwo} 0.8235 & \cellcolor{AppRankPeak} 0.9412 & 0.5759 & 0.6347 & \cellcolor{AppRankThree} 0.7924 & \cellcolor{AppRankOne} 0.8918 & 0.7294 & \cellcolor{AppRankFour} 0.7412 \\
vowel & \cellcolor{AppRankFour} 0.8990 & 0.5051 & 0.8485 & 0.7071 & 0.8081 & 0.5960 & 0.6869 & \cellcolor{AppRankThree} 0.9091 & \cellcolor{AppRankThree} 0.9091 & \cellcolor{AppRankThree} 0.9091 & \cellcolor{AppRankThree} 0.9091 & 0.7879 & 0.8889 & \cellcolor{AppRankPeak} 0.9293 & \cellcolor{AppRankFour} 0.8990 & 0.5692 & 0.7460 & 0.8828 & \cellcolor{AppRankTwo} 0.9096 & \cellcolor{AppRankFour} 0.8990 & \cellcolor{AppRankOne} 0.9191 \\
wisco & \cellcolor{AppRankTwo} 0.6000 & \cellcolor{AppRankPeak} 0.7500 & 0.5000 & \cellcolor{AppRankFour} 0.5500 & \cellcolor{AppRankFour} 0.5500 & \cellcolor{AppRankFour} 0.5500 & 0.5000 & 0.4500 & \cellcolor{AppRankTwo} 0.6000 & 0.4500 & \cellcolor{AppRankFour} 0.5500 & 0.4000 & \cellcolor{AppRankFour} 0.5500 & 0.3500 & \cellcolor{AppRankFour} 0.5500 & 0.4821 & \cellcolor{AppRankThree} 0.5564 & 0.5385 & 0.5385 & \cellcolor{AppRankFour} 0.5500 & \cellcolor{AppRankOne} 0.6500 \\
\midrule
\rowcolor{AppSummary}
\textbf{Mean} & \textbf{0.5804} & \textbf{0.5930} & \textbf{0.5765} & \textbf{0.5599} & \textbf{0.5870} & \textbf{0.5313} & \textbf{0.5519} & \textbf{0.6255} & \textbf{0.6580} & \textbf{0.6925} & \cellcolor{AppRankOne}\textbf{0.7246} & \textbf{0.5546} & \textbf{0.5894} & \textbf{0.6647} & \cellcolor{AppRankTwo}\textbf{0.7219} & \textbf{0.5353} & \textbf{0.5985} & \textbf{0.6527} & \cellcolor{AppRankFour}\textbf{0.6974} & \cellcolor{AppRankThree}\textbf{0.7021} & \cellcolor{AppRankPeak}\textbf{0.7542} \\
\rowcolor{AppSummary}
\textbf{Rank i} & \textbf{9.25} & \textbf{8.49} & \textbf{9.04} & \textbf{9.68} & \textbf{8.81} & \textbf{9.81} & \textbf{9.96} & \textbf{8.30} & \textbf{7.17} & \textbf{5.87} & \cellcolor{AppRankOne}\textbf{4.42} & \textbf{9.30} & \textbf{8.87} & \textbf{6.47} & \cellcolor{AppRankThree}\textbf{4.74} & \textbf{11.21} & \textbf{8.75} & \textbf{6.40} & \cellcolor{AppRankTwo}\textbf{4.55} & \cellcolor{AppRankFour}\textbf{5.02} & \cellcolor{AppRankPeak}\textbf{3.11} \\
\bottomrule
\end{tabular}
}
\endgroup
\end{table}

\noindent\textbf{Zero-shot and few-shot regression.} NMAE is reported on the shared unseen tasks.
\begin{table}[H]
\centering
\caption{Per-task NMAE ($\downarrow$) on ZeroTabBench and FewTabBench, with supervised few-shot comparisons. Darker blue cells indicate lower NMAE. General-purpose LLM columns and TabXLLM's zero-shot column report ZeroTabBench results; TabXLLM's 32-shot column reports FewTabBench results. \textcolor{FigBlueDark}{\textbf{TabPFN}} and \textcolor{FigBlueDark}{\textbf{XGBoost}} use labeled target-task samples for supervised few-shot fitting.}
\label{full_table:regression_few_zero}
\begingroup
\fontsize{5.266897}{5.793587}\selectfont
\setlength{\tabcolsep}{1.053379pt}
\renewcommand{\arraystretch}{1.15}
\setlength{\heavyrulewidth}{0.421352pt}
\setlength{\lightrulewidth}{0.263345pt}
\setlength{\cmidrulewidth}{0.158007pt}
\setlength{\aboverulesep}{0.907054pt}
\setlength{\belowrulesep}{1.473978pt}
\setlength{\cmidrulekern}{2.633449pt}
\begin{tabular}{lccccccccccccccccc}
\toprule
\multirow{2}{*}{\vspace{-2.106759pt}Dataset} & \multirow{2}{*}{\vspace{-2.106759pt}Qwen-2.5-7B} & \multirow{2}{*}{\vspace{-2.106759pt}Qwen-2.5-32B} & \multirow{2}{*}{\vspace{-2.106759pt}Qwen3-8B} & \multirow{2}{*}{\vspace{-2.106759pt}Qwen3-32B} & \multirow{2}{*}{\vspace{-2.106759pt}Qwen3-235B} & \multirow{2}{*}{\vspace{-2.106759pt}DS-R1} & \multirow{2}{*}{\vspace{-2.106759pt}DS-0528} & \multicolumn{4}{c}{\textbf{XGBoost}} & \multicolumn{4}{c}{\textbf{TabPFN}} & \multicolumn{2}{c}{\textcolor{AppAccent}{\textbf{TabXLLM-8B}}} \\
\cmidrule(lr){9-12} \cmidrule(lr){13-16} \cmidrule{17-18}
& &  &  &  &  &  &  & \textbf{4 shot} & \textbf{8 shot} & \textbf{16 shot} & \textbf{32 shot} & \textbf{4 shot} & \textbf{8 shot} & \textbf{16 shot} & \textbf{32 shot} & \textcolor{AppAccent}{\textbf{ZS transfer}} & \textcolor{AppAccent}{\textbf{32 shot}}\\
\midrule

\rowcolors{2}{AppStripe}{white}
ames & 0.0839 & 0.0997 & 0.0919 & \cellcolor{AppRankFour} 0.0578 & 0.0635 & \cellcolor{AppRankFive} 0.0605 & 0.1009 & 0.2751 & 0.0926 & 0.0777 & 0.0858 & 0.1003 & 0.0706 & \cellcolor{AppRankThree} 0.0541 & \cellcolor{AppRankTwo} 0.0432 & 0.0868 & \cellcolor{AppRankOne} 0.0335 \\
aucti & 0.1831 & 0.1838 & \cellcolor{AppRankFour} 0.1804 & \cellcolor{AppRankFive} 0.1816 & 0.1819 & 0.1826 & 0.1846 & 0.3189 & 0.2286 & 0.1876 & 0.3318 & 0.1931 & 0.1879 & \cellcolor{AppRankThree} 0.1512 & \cellcolor{AppRankOne} 0.0732 & 0.1829 & \cellcolor{AppRankTwo} 0.0828 \\
Bosto & 0.1346 & 0.1121 & 0.1506 & 0.1360 & \cellcolor{AppRankOne} 0.0619 & \cellcolor{AppRankTwo} 0.0675 & 43.1375 & 0.2525 & 0.1170 & 0.1077 & 0.0794 & 0.1856 & 0.1127 & \cellcolor{AppRankFive} 0.0782 & \cellcolor{AppRankFour} 0.0756 & 0.1251 & \cellcolor{AppRankThree} 0.0687 \\
cars & 0.3299 & 0.2764 & 0.3575 & 0.2941 & 0.3314 & 0.2907 & 0.4177 & 0.0974 & \cellcolor{AppRankFive} 0.0905 & \cellcolor{AppRankFour} 0.0867 & \cellcolor{AppRankOne} 0.0469 & 0.1790 & 0.1578 & 0.1051 & \cellcolor{AppRankThree} 0.0681 & 0.1567 & \cellcolor{AppRankTwo} 0.0584 \\
colre & 0.2892 & 0.2378 & 0.3233 & 1.0247 & 1.1183 & 0.4831 & 0.8425 & 0.2406 & \cellcolor{AppRankFive} 0.1890 & \cellcolor{AppRankThree} 0.1710 & 0.2072 & 0.3113 & 0.2052 & \cellcolor{AppRankFour} 0.1733 & \cellcolor{AppRankTwo} 0.1420 & 0.2426 & \cellcolor{AppRankOne} 0.0906 \\
concr & 0.2237 & 0.1704 & 0.1961 & 0.2316 & \cellcolor{AppRankTwo} 0.1106 & \cellcolor{AppRankThree} 0.1139 & 0.1685 & 0.2326 & 0.1630 & 0.1661 & \cellcolor{AppRankFive} 0.1230 & 0.1829 & 0.1567 & \cellcolor{AppRankFour} 0.1194 & \cellcolor{AppRankOne} 0.0851 & 0.1685 & 0.1499 \\
CPMP & 0.3852 & 0.3860 & 0.3854 & \cellcolor{AppRankFive} 0.3835 & 0.3845 & \cellcolor{AppRankFour} 0.3793 & 0.3856 & 0.4157 & 0.6947 & 0.5518 & 0.4200 & 0.3977 & 0.3880 & \cellcolor{AppRankTwo} 0.3705 & \cellcolor{AppRankThree} 0.3780 & 0.3855 & \cellcolor{AppRankOne} 0.2864 \\
emplo & 0.1665 & 0.1804 & \cellcolor{AppRankFive} 0.0599 & 0.1034 & 0.0697 & \cellcolor{AppRankThree} 0.0418 & 0.1888 & 0.1216 & 0.1249 & 0.0746 & 0.0642 & 0.1048 & 0.0720 & \cellcolor{AppRankFour} 0.0506 & \cellcolor{AppRankTwo} 0.0285 & 0.0860 & \cellcolor{AppRankOne} 0.0209 \\
healt & 0.1836 & 0.1512 & 0.1537 & \cellcolor{AppRankFive} 0.1247 & 0.1517 & 0.1478 & 0.1656 & 0.4697 & 0.2409 & 0.1843 & \cellcolor{AppRankFour} 0.1023 & 0.2098 & 0.1501 & \cellcolor{AppRankThree} 0.1005 & \cellcolor{AppRankOne} 0.0828 & 0.1503 & \cellcolor{AppRankTwo} 0.0947 \\
house & 0.0937 & 0.0675 & \cellcolor{AppRankFour} 0.0571 & 0.0578 & \cellcolor{AppRankTwo} 0.0405 & 0.0774 & 0.0697 & 0.2050 & 0.0715 & 0.0660 & \cellcolor{AppRankFive} 0.0574 & 0.0890 & 0.0691 & \cellcolor{AppRankThree} 0.0510 & \cellcolor{AppRankOne} 0.0381 & 0.0632 & 0.0651 \\
Lisbo & 0.1661 & 0.3716 & 0.3182 & 0.1666 & \cellcolor{AppRankFive} 0.1269 & 0.1269 & 0.6827 & 0.1881 & 0.2410 & 0.2204 & 0.1948 & 0.1744 & 0.1298 & \cellcolor{AppRankThree} 0.1125 & \cellcolor{AppRankTwo} 0.0918 & \cellcolor{AppRankFour} 0.1267 & \cellcolor{AppRankOne} 0.0570 \\
lowbw & 0.1755 & 0.1520 & 0.2667 & 0.1732 & \cellcolor{AppRankFour} 0.1393 & \cellcolor{AppRankTwo} 0.1268 & 0.2929 & 0.1541 & 0.1797 & 0.1601 & 0.1683 & 0.2024 & \cellcolor{AppRankFive} 0.1490 & \cellcolor{AppRankThree} 0.1275 & \cellcolor{AppRankOne} 0.1171 & 0.1759 & 0.1592 \\
mauna & 0.2041 & \cellcolor{AppRankFive} 0.0333 & 0.0573 & 0.0352 & \cellcolor{AppRankOne} 0.0244 & \cellcolor{AppRankTwo} 0.0246 & 0.0832 & 0.1577 & 0.1278 & 0.0779 & 0.0696 & 0.1126 & 0.0445 & \cellcolor{AppRankThree} 0.0298 & \cellcolor{AppRankFour} 0.0303 & 0.0508 & 0.0531 \\
newto & 0.4029 & 1.6552 & 0.8940 & 0.3716 & 0.3384 & 0.3670 & 0.5957 & 0.2934 & \cellcolor{AppRankTwo} 0.2184 & 0.2649 & 0.3309 & 0.2807 & \cellcolor{AppRankFive} 0.2521 & \cellcolor{AppRankThree} 0.2307 & \cellcolor{AppRankOne} 0.2001 & \cellcolor{AppRankFour} 0.2429 & 0.2857 \\
NHANE & 0.2972 & 0.2798 & 0.2676 & 0.2925 & 0.2644 & 0.2652 & 0.2933 & 0.4441 & 0.3740 & 0.3218 & 0.3794 & 0.2833 & \cellcolor{AppRankFour} 0.2602 & \cellcolor{AppRankThree} 0.2502 & \cellcolor{AppRankTwo} 0.2299 & \cellcolor{AppRankFive} 0.2636 & \cellcolor{AppRankOne} 0.1861 \\
place & \cellcolor{AppRankFour} 0.1765 & 0.5358 & 0.5702 & 0.4590 & 0.4478 & 0.4549 & 0.3526 & 0.1953 & 0.2105 & 0.4233 & \cellcolor{AppRankFive} 0.1866 & 0.2181 & 0.1899 & \cellcolor{AppRankThree} 0.1740 & \cellcolor{AppRankTwo} 0.1653 & 0.1925 & \cellcolor{AppRankOne} 0.1618 \\
pm10 & 0.1888 & 0.1992 & 0.1909 & 0.1893 & 0.1819 & 0.1883 & 0.2176 & 0.2090 & \cellcolor{AppRankFive} 0.1736 & 0.1809 & 0.2337 & 0.2220 & \cellcolor{AppRankFour} 0.1681 & \cellcolor{AppRankThree} 0.1591 & \cellcolor{AppRankTwo} 0.1457 & 0.1891 & \cellcolor{AppRankOne} 0.1202 \\
seoul & 0.1747 & 0.1223 & 0.1412 & \cellcolor{AppRankFour} 0.1181 & \cellcolor{AppRankOne} 0.1088 & \cellcolor{AppRankThree} 0.1110 & 0.1333 & 0.1644 & 0.2444 & 0.5734 & 0.1514 & 0.1531 & 0.1690 & 0.1306 & \cellcolor{AppRankTwo} 0.1095 & 0.1326 & \cellcolor{AppRankFive} 0.1219 \\
std & \cellcolor{AppRankTwo} 0.2026 & 1.2481 & 0.2262 & 0.3650 & 0.9962 & 1.2384 & 0.7261 & 0.2380 & \cellcolor{AppRankFive} 0.2125 & 0.5603 & 0.2984 & 0.2429 & \cellcolor{AppRankFour} 0.2114 & 0.2146 & \cellcolor{AppRankThree} 0.2045 & 0.2238 & \cellcolor{AppRankOne} 0.2005 \\
weath & \cellcolor{AppRankFive} 0.0325 & \cellcolor{AppRankFive} 0.0325 & \cellcolor{AppRankFive} 0.0325 & 0.0642 & \cellcolor{AppRankFour} 0.0324 & 0.0325 & \cellcolor{AppRankFive} 0.0325 & 0.1638 & 0.1558 & 0.0610 & 0.0612 & 0.1360 & 0.0371 & \cellcolor{AppRankThree} 0.0264 & \cellcolor{AppRankTwo} 0.0207 & \cellcolor{AppRankFive} 0.0325 & \cellcolor{AppRankOne} 0.0125 \\
yacht & 0.1502 & 0.1112 & 0.1396 & 0.1475 & 0.1673 & 0.1106 & 0.1425 & 0.3972 & 0.2245 & 0.2645 & \cellcolor{AppRankFive} 0.0695 & 0.1998 & \cellcolor{AppRankFour} 0.0434 & \cellcolor{AppRankThree} 0.0400 & \cellcolor{AppRankOne} 0.0207 & 0.1113 & \cellcolor{AppRankTwo} 0.0216 \\
\midrule
\rowcolor{AppSummary}
\textbf{Mean} & \textbf{0.2021} & \textbf{0.3146} & \textbf{0.2409} & \textbf{0.2370} & \textbf{0.2544} & \textbf{0.2329} & \textbf{2.3435} & \textbf{0.2492} & \textbf{0.2083} & \textbf{0.2277} & \textbf{0.1744} & \textbf{0.1990} & \cellcolor{AppRankFour}\textbf{0.1536} & \cellcolor{AppRankThree}\textbf{0.1309} & \cellcolor{AppRankTwo}\textbf{0.1119} & \cellcolor{AppRankFive}\textbf{0.1614} & \cellcolor{AppRankOne}\textbf{0.1110} \\
\rowcolor{AppSummary}
\textbf{Rank i} & \textbf{11.05} & \textbf{10.76} & \textbf{10.81} & \textbf{9.86} & \cellcolor{AppRankFour}\textbf{7.24} & \cellcolor{AppRankFour}\textbf{7.24} & \textbf{13.10} & \textbf{13.14} & \textbf{11.33} & \textbf{10.33} & \textbf{9.19} & \textbf{12.52} & \cellcolor{AppRankFive}\textbf{7.43} & \cellcolor{AppRankThree}\textbf{3.71} & \cellcolor{AppRankOne}\textbf{2.00} & \textbf{8.19} & \cellcolor{AppRankTwo}\textbf{3.10} \\
\bottomrule
\end{tabular}
\endgroup
\end{table}




\FloatBarrier
\Needspace{120pt}
\section{Detailed Information of Datasets}\label{appendix:dataset_details}
To provide a clear overview of the datasets used in our experiments, we summarize the detailed information of the 139 OpenML datasets in this section.
Each dataset includes its abbreviation, full name, experimental setting, domain classification, and OpenML reference.
Datasets marked \texttt{zero\_few\_shot} in \Cref{tab:setting and domain} constitute the tasks shared by ZeroTabBench and FewTabBench; the remaining datasets support fully trained experiments.

In the experimental result tables (\Cref{full_table:classification_fully_trained}, \Cref{full_table:regression_fully_trained}, \Cref{full_table:classification_few_zero}, \Cref{full_table:regression_few_zero}), we use abbreviated dataset names for brevity.
\Cref{tab:abbreviation} presents the full names of all datasets used in the experiments.
\Cref{tab:setting and domain} summarizes the experimental settings and corresponding domain classifications for each dataset,
while \Cref{tab:url} provides the OpenML IDs and URLs for reference and reproducibility.
\begingroup
\renewcommand{\arraystretch}{1.03}
\setlength{\LTpre}{6pt}
\setlength{\LTpost}{6pt}
{\small
\begin{longtable}{ll}
\caption{Abbreviations and full names of the OpenML datasets used in our experiments.}
\label{tab:abbreviation}\\
\toprule
\makecell[l]{\textbf{Dataset}\\\textbf{Abbreviation}} & \textbf{Full dataset name} \\
\midrule
\endfirsthead
\multicolumn{2}{l}{\small\itshape (continued)} \\
\toprule
\makecell[l]{\textbf{Dataset}\\\textbf{Abbreviation}} & \textbf{Full dataset name} \\
\midrule
\endhead
\midrule
\multicolumn{2}{r}{\small\itshape Continued on next page} \\
\endfoot
\bottomrule
\endlastfoot
1000 & 1000\_Cameras\_Dataset \\*
AAPL & AAPL\_stock\_price\_2021\_2022 \\*
AVIDa & AVIDa\_SARS\_CoV\_2 \\*
Amazo & Amazon\_employee\_access\_seed\_0\_nrows\_2000\_nclasses\_10\_ncols\_100\_stratify\_True \\*
Bank & Bank\_Customer\_Churn \\
Bosto & Boston\_house\_price\_data \\
CPMP & CPMP\_2015\_runtime\_regression \\
Carpe & CarpenterFdaData \\
Click & Click\_prediction\_small\_seed\_0\_nrows\_2000\_nclasses\_10\_ncols\_100\_stratify\_True \\
Emplo & Employee \\
Fraud & Fraud\_Detection\_Updated \\
Goodr & Goodreads\_Computer\_Books \\
HMEQ & HMEQ\_Data \\
Lisbo & Lisbon\_House\_Prices \\
Loan & Loan\_Predication \\
MIP & MIP\_2016\_regression \\
Marke & Marketing\_Campaign \\
NATIC & NATICUSdroid \\
NHANE & NHANES\_age \\
PostP & PostPartum\_Depression \\
SPECT & SPECTF \\
Skin & Skin\_Cancer\_PAD\_UFES\_20 \\
Speed & SpeedDating \\
Tour & Tour\_and\_Travels\_Customer\_Churn\_Prediction \\
Tunin & TuningSVMs \\
UCI & UCI\_churn \\
WMO & WMO\_Hurricane\_Survival\_Dataset \\
ada & ada\_prior \\
adult & adult\_seed\_0\_nrows\_2000\_nclasses\_10\_ncols\_100\_stratify\_True \\
airli & airlines\_seed\_0\_nrows\_2000\_nclasses\_10\_ncols\_100\_stratify\_True \\
ames & ames\_housing \\
analc & analcatdata\_apnea1 \\
annea & anneal \\
arrhy & arrhythmia \\
arsen & arsenic\_female\_bladder \\
art & art\_daily\_nojump4 \\
aucti & auction\_verification \\
autoM & autoMpg \\
blast & blastchar \\
blood & blood\_transfusion\_service\_center \\
brazi & braziltourism \\
breas & breast\_cancer \\
brstw & breast\_w \\
bwin & bwin\_amlb \\
calif & california\_seed\_0\_nrows\_2000\_nclasses\_10\_ncols\_100\_stratify\_True \\
cars & cars \\
chole & cholesterol (classification) \\
chole & cholesterol (regression) \\
churn & churn \\
cleve & cleve \\
coffe & coffee\_distribution\_across\_94\_counties \\
coil2 & coil2000\_insurance\_policies \\
colic & colic \\
colre & colrec \\
commu & communities\_and\_crime\_binary \\
compa & compas\_two\_years \\
concr & concrete\_compressive\_strength \\
confe & conference\_attendance \\
datas & dataset\_credit\_g \\
datat & datatrieve \\
depre & depression\_2020 \\
dgf & dgf\_test \\
diabe & diabetes \\
diffe & differentiated\_thyroid\_cancer\_recurrence \\
dis & dis \\
discl & disclosure\_x\_noise \\
doa & doa\_bwin \\
e1684 & e1684 \\
emplo & employee\_salaries \\
eye & eye\_movements \\
flags & flags \\
haber & haberman \\
hayes & hayes\_roth \\
healt & healthcare\_insurance\_expenses \\
hepat & hepatitis \\
house & house\_prices \\
hypot & hypothyroid \\
ibm & ibm\_employee\_attrition \\
imdb & imdb\_genre\_prediction \\
ipums & ipums\_la\_98\_small \\
iris & iris \\
irish & irish \\
jungl & jungle\_chess\_2pcs\_endgame\_elephant\_elephant \\
kc1 & kc1 \\
kc2 & kc2 \\
kc3 & kc3 \\
kdd & kdd\_el\_nino\_small \\
kick & kick\_seed\_0\_nrows\_2000\_nclasses\_10\_ncols\_100\_stratify\_True \\
lowbw & lowbwt \\
lungc & lungcancer\_GSE31210 \\
mauna & mauna\_loa\_atmospheric\_co2 \\
mc1 & mc1 \\
mc2 & mc2 \\
meta & meta \\
munic & munich\_rent\_index\_1999 \\
mw1 & mw1 \\
newto & newton\_hema (classification) \\
newto & newton\_hema (regression) \\
no2 & no2 \\
nwtco & nwtco \\
page & page\_blocks \\
pakis & pakistan\_hunger\_data \\
pc1 & pc1 \\
pc2 & pc2 \\
pc3 & pc3 \\
pc4 & pc4 \\
phary & pharynx \\
place & places \\
plasm & plasma\_retinol (classification) \\
plasm & plasma\_retinol (regression) \\
pm10 & pm10 (classification) \\
pm10 & pm10 (regression) \\
polis & polish\_companies\_bankruptcy \\
polle & pollen \\
prnn & prnn\_crabs \\
profb & profb \\
quake & quake \\
regim & regime\_alimentaire \\
road & road\_safety\_seed\_0\_nrows\_2000\_nclasses\_10\_ncols\_100\_stratify\_True \\
segme & segment \\
seism & seismic\_bumps \\
seoul & seoul\_bike\_sharing\_demand\_cat \\
sf & sf\_police\_incidents\_seed\_0\_nrows\_2000\_nclasses\_10\_ncols\_100\_stratify\_True \\
solar & solar\_flare \\
spamb & spambase \\
spect & spectrometer \\
std & std \\
taiwa & taiwanese\_bankruptcy\_prediction \\
telco & telco\_customer\_churn \\
tempe & temperature\_emissions\_environmental\_trends\_2000\_2024 \\
tic & tic\_tac\_toe \\
triaz & triazines \\
tumor & tumor \\
vehic & vehicle \\
vowel & vowel \\
weath & weather\_ankara \\
wilt & wilt \\
wisco & wisconsin \\
yacht & yacht\_hydrodynamics \\
\end{longtable}
}

\begin{longtable}{lll}
\caption{Experimental settings and domain categories of the datasets.}\label{tab:setting and domain}\\
\toprule
\textbf{Dataset abbreviation} & \textbf{experiment\_setting} & \textbf{domain} \\
\midrule
\endfirsthead
\multicolumn{3}{l}{\small\itshape (continued)} \\
\toprule
\textbf{Dataset abbreviation} & \textbf{experiment\_setting} & \textbf{domain} \\
\midrule
\endhead
\midrule
\multicolumn{3}{r}{\small\itshape Continued on next page} \\
\endfoot
\bottomrule
\endlastfoot
1000 & regression\_fully\_trained & software \\*
AAPL & regression\_fully\_trained & finance \\*
AVIDa & classification\_fully\_trained & medical \\*
Amazo & classification\_fully\_trained & finance \\*
Bank & classification\_zero\_few\_shot & finance \\
Bosto & regression\_zero\_few\_shot & business \\
CPMP & regression\_zero\_few\_shot & software \\
Carpe & regression\_fully\_trained & medical \\
Click & classification\_fully\_trained & finance \\
Emplo & classification\_zero\_few\_shot & business \\
Fraud & classification\_fully\_trained & finance \\
Goodr & regression\_fully\_trained & education \\
HMEQ & classification\_zero\_few\_shot & finance \\
Lisbo & regression\_zero\_few\_shot & business \\
Loan & classification\_fully\_trained & finance \\
MIP & regression\_fully\_trained & finance \\
Marke & classification\_fully\_trained & business \\
NATIC & classification\_zero\_few\_shot & software \\
NHANE & regression\_zero\_few\_shot & medical \\
PostP & classification\_zero\_few\_shot & medical \\
SPECT & classification\_zero\_few\_shot & medical \\
Skin & classification\_zero\_few\_shot & medical \\
Speed & classification\_fully\_trained & business \\
Tour & classification\_zero\_few\_shot & finance \\
Tunin & classification\_zero\_few\_shot & education \\
UCI & classification\_zero\_few\_shot & finance \\
WMO & classification\_fully\_trained & generic \\
ada & classification\_fully\_trained & generic \\
adult & classification\_zero\_few\_shot & finance \\
airli & classification\_zero\_few\_shot & business \\
ames & regression\_zero\_few\_shot & business \\
analc & classification\_zero\_few\_shot & medical \\
annea & classification\_zero\_few\_shot & science \\
arrhy & classification\_zero\_few\_shot & medical \\
arsen & classification\_fully\_trained & medical \\
art & classification\_fully\_trained & science \\
aucti & regression\_zero\_few\_shot & business \\
autoM & classification\_zero\_few\_shot & science \\
blast & classification\_zero\_few\_shot & business \\
blood & classification\_fully\_trained & medical \\
brazi & classification\_zero\_few\_shot & business \\
breas & classification\_fully\_trained & medical \\
brstw & classification\_fully\_trained & medical \\
bwin & classification\_fully\_trained & finance \\
calif & classification\_zero\_few\_shot & generic \\
cars & regression\_zero\_few\_shot & business \\
chole & classification\_zero\_few\_shot & medical \\
chole & regression\_fully\_trained & medical \\
churn & classification\_zero\_few\_shot & finance \\
cleve & classification\_zero\_few\_shot & medical \\
coffe & regression\_fully\_trained & business \\
coil2 & classification\_fully\_trained & finance \\
colic & classification\_zero\_few\_shot & medical \\
colre & regression\_zero\_few\_shot & software \\
commu & classification\_zero\_few\_shot & science \\
compa & classification\_zero\_few\_shot & generic \\
concr & regression\_zero\_few\_shot & software \\
confe & classification\_fully\_trained & business \\
datas & classification\_fully\_trained & finance \\
datat & classification\_fully\_trained & education \\
depre & classification\_fully\_trained & medical \\
dgf & classification\_zero\_few\_shot & generic \\
diabe & classification\_zero\_few\_shot & medical \\
diffe & classification\_zero\_few\_shot & medical \\
dis & classification\_fully\_trained & medical \\
discl & regression\_fully\_trained & generic \\
doa & classification\_fully\_trained & finance \\
e1684 & regression\_fully\_trained & medical \\
emplo & regression\_zero\_few\_shot & business \\
eye & classification\_zero\_few\_shot & science \\
flags & classification\_zero\_few\_shot & generic \\
haber & classification\_fully\_trained & medical \\
hayes & classification\_zero\_few\_shot & education \\
healt & regression\_zero\_few\_shot & medical \\
hepat & classification\_zero\_few\_shot & medical \\
house & regression\_zero\_few\_shot & business \\
hypot & classification\_zero\_few\_shot & medical \\
ibm & classification\_zero\_few\_shot & finance \\
imdb & classification\_fully\_trained & education \\
ipums & classification\_fully\_trained & generic \\
iris & classification\_fully\_trained & science \\
irish & classification\_fully\_trained & science \\
jungl & classification\_zero\_few\_shot & software \\
kc1 & classification\_fully\_trained & software \\
kc2 & classification\_zero\_few\_shot & software \\
kc3 & classification\_fully\_trained & software \\
kdd & classification\_zero\_few\_shot & science \\
kick & classification\_fully\_trained & finance \\
lowbw & regression\_zero\_few\_shot & medical \\
lungc & classification\_zero\_few\_shot & medical \\
mauna & regression\_zero\_few\_shot & science \\
mc1 & classification\_fully\_trained & software \\
mc2 & classification\_fully\_trained & software \\
meta & classification\_fully\_trained & education \\
munic & regression\_fully\_trained & finance \\
mw1 & classification\_fully\_trained & software \\
newto & classification\_zero\_few\_shot & medical \\
newto & regression\_zero\_few\_shot & medical \\
no2 & classification\_zero\_few\_shot & science \\
nwtco & regression\_fully\_trained & medical \\
page & classification\_zero\_few\_shot & software \\
pakis & regression\_fully\_trained & business \\
pc1 & classification\_fully\_trained & software \\
pc2 & classification\_fully\_trained & software \\
pc3 & classification\_fully\_trained & software \\
pc4 & classification\_fully\_trained & software \\
phary & classification\_zero\_few\_shot & medical \\
place & regression\_zero\_few\_shot & science \\
plasm & classification\_fully\_trained & medical \\
plasm & regression\_fully\_trained & medical \\
pm10 & classification\_zero\_few\_shot & science \\
pm10 & regression\_zero\_few\_shot & science \\
polis & classification\_fully\_trained & finance \\
polle & classification\_fully\_trained & science \\
prnn & classification\_zero\_few\_shot & science \\
profb & classification\_fully\_trained & business \\
quake & classification\_fully\_trained & science \\
regim & classification\_fully\_trained & business \\
road & classification\_zero\_few\_shot & science \\
segme & classification\_zero\_few\_shot & software \\
seism & classification\_fully\_trained & generic \\
seoul & regression\_zero\_few\_shot & science \\
sf & classification\_fully\_trained & business \\
solar & classification\_fully\_trained & science \\
spamb & classification\_zero\_few\_shot & software \\
spect & classification\_zero\_few\_shot & science \\
std & regression\_zero\_few\_shot & medical \\
taiwa & classification\_fully\_trained & finance \\
telco & classification\_zero\_few\_shot & finance \\
tempe & regression\_fully\_trained & science \\
tic & classification\_fully\_trained & generic \\
triaz & classification\_zero\_few\_shot & science \\
tumor & regression\_fully\_trained & medical \\
vehic & classification\_zero\_few\_shot & science \\
vowel & classification\_zero\_few\_shot & science \\
weath & regression\_zero\_few\_shot & science \\
wilt & classification\_fully\_trained & science \\
wisco & classification\_zero\_few\_shot & medical \\
yacht & regression\_zero\_few\_shot & science \\
\end{longtable}

\renewcommand{\arraystretch}{1.03919}
\begin{longtable}{lll}
\caption{OpenML IDs and URLs of the datasets for reference.}\label{tab:url}\\
\toprule
\textbf{Dataset abbreviation} & \textbf{openml\_ID} & \textbf{openml\_URL} \\
\midrule
\endfirsthead
\multicolumn{3}{l}{\small\itshape (continued)} \\
\toprule
\textbf{Dataset abbreviation} & \textbf{openml\_ID} & \textbf{openml\_URL} \\
\midrule
\endhead
\midrule
\multicolumn{3}{r}{\small\itshape Continued on next page} \\
\endfoot
\bottomrule
\endlastfoot
1000 & 43714 & https://www.openml.org/d/43714 \\*
AAPL & 43878 & https://www.openml.org/d/43878 \\*
AVIDa & 46800 & https://www.openml.org/d/46800 \\*
Amazo & 44708 & https://www.openml.org/d/44708 \\*
Bank & 46911 & https://www.openml.org/d/46911 \\
Bosto & 43465 & https://www.openml.org/d/43465 \\
CPMP & 41928 & https://www.openml.org/d/41928 \\
Carpe & 46159 & https://www.openml.org/d/46159 \\
Click & 44763 & https://www.openml.org/d/44763 \\
Emplo & 46925 & https://www.openml.org/d/46925 \\
Fraud & 46359 & https://www.openml.org/d/46359 \\
Goodr & 43785 & https://www.openml.org/d/43785 \\
HMEQ & 46431 & https://www.openml.org/d/46431 \\
Lisbo & 43660 & https://www.openml.org/d/43660 \\
Loan & 43595 & https://www.openml.org/d/43595 \\
MIP & 43070 & https://www.openml.org/d/43070 \\
Marke & 46940 & https://www.openml.org/d/46940 \\
NATIC & 46969 & https://www.openml.org/d/46969 \\
NHANE & 46946 & https://www.openml.org/d/46946 \\
PostP & 46758 & https://www.openml.org/d/46758 \\
SPECT & 337 & https://www.openml.org/d/337 \\
Skin & 46630 & https://www.openml.org/d/46630 \\
Speed & 40536 & https://www.openml.org/d/40536 \\
Tour & 45545 & https://www.openml.org/d/45545 \\
Tunin & 41976 & https://www.openml.org/d/41976 \\
UCI & 44232 & https://www.openml.org/d/44232 \\
WMO & 43607 & https://www.openml.org/d/43607 \\
ada & 1037 & https://www.openml.org/d/1037 \\
adult & 44723 & https://www.openml.org/d/44723 \\
airli & 44528 & https://www.openml.org/d/44528 \\
ames & 43926 & https://www.openml.org/d/43926 \\
analc & 767 & https://www.openml.org/d/767 \\
annea & 989 & https://www.openml.org/d/989 \\
arrhy & 1017 & https://www.openml.org/d/1017 \\
arsen & 949 & https://www.openml.org/d/949 \\
art & 43051 & https://www.openml.org/d/43051 \\
aucti & 44958 & https://www.openml.org/d/44958 \\
autoM & 831 & https://www.openml.org/d/831 \\
blast & 46280 & https://www.openml.org/d/46280 \\
blood & 46913 & https://www.openml.org/d/46913 \\
brazi & 957 & https://www.openml.org/d/957 \\
breas & 13 & https://www.openml.org/d/13 \\
brstw & 15 & https://www.openml.org/d/15 \\
bwin & 45717 & https://www.openml.org/d/45717 \\
calif & 44362 & https://www.openml.org/d/44362 \\
cars & 46914 & https://www.openml.org/d/46914 \\
chole & 798 & https://www.openml.org/d/798 \\
chole & 204 & https://www.openml.org/d/204 \\
churn & 40701 & https://www.openml.org/d/40701 \\
cleve & 40710 & https://www.openml.org/d/40710 \\
coffe & 46748 & https://www.openml.org/d/46748 \\
coil2 & 46916 & https://www.openml.org/d/46916 \\
colic & 25 & https://www.openml.org/d/25 \\
colre & 46145 & https://www.openml.org/d/46145 \\
commu & 43891 & https://www.openml.org/d/43891 \\
compa & 42192 & https://www.openml.org/d/42192 \\
concr & 44959 & https://www.openml.org/d/44959 \\
confe & 41538 & https://www.openml.org/d/41538 \\
datas & 46532 & https://www.openml.org/d/46532 \\
datat & 1075 & https://www.openml.org/d/1075 \\
depre & 46876 & https://www.openml.org/d/46876 \\
dgf & 42882 & https://www.openml.org/d/42882 \\
diabe & 37 & https://www.openml.org/d/37 \\
diffe & 46605 & https://www.openml.org/d/46605 \\
dis & 40713 & https://www.openml.org/d/40713 \\
discl & 704 & https://www.openml.org/d/704 \\
doa & 45711 & https://www.openml.org/d/45711 \\
e1684 & 46157 & https://www.openml.org/d/46157 \\
emplo & 42125 & https://www.openml.org/d/42125 \\
eye & 43946 & https://www.openml.org/d/43946 \\
flags & 1012 & https://www.openml.org/d/1012 \\
haber & 43 & https://www.openml.org/d/43 \\
hayes & 974 & https://www.openml.org/d/974 \\
healt & 46931 & https://www.openml.org/d/46931 \\
hepat & 55 & https://www.openml.org/d/55 \\
house & 42165 & https://www.openml.org/d/42165 \\
hypot & 1000 & https://www.openml.org/d/1000 \\
ibm & 43893 & https://www.openml.org/d/43893 \\
imdb & 46667 & https://www.openml.org/d/46667 \\
ipums & 1002 & https://www.openml.org/d/1002 \\
iris & 969 & https://www.openml.org/d/969 \\
irish & 451 & https://www.openml.org/d/451 \\
jungl & 40999 & https://www.openml.org/d/40999 \\
kc1 & 1067 & https://www.openml.org/d/1067 \\
kc2 & 1063 & https://www.openml.org/d/1063 \\
kc3 & 1065 & https://www.openml.org/d/1065 \\
kdd & 839 & https://www.openml.org/d/839 \\
kick & 44758 & https://www.openml.org/d/44758 \\
lowbw & 203 & https://www.openml.org/d/203 \\
lungc & 1412 & https://www.openml.org/d/1412 \\
mauna & 41187 & https://www.openml.org/d/41187 \\
mc1 & 1056 & https://www.openml.org/d/1056 \\
mc2 & 1054 & https://www.openml.org/d/1054 \\
meta & 757 & https://www.openml.org/d/757 \\
munic & 46772 & https://www.openml.org/d/46772 \\
mw1 & 1071 & https://www.openml.org/d/1071 \\
newto & 784 & https://www.openml.org/d/784 \\
newto & 492 & https://www.openml.org/d/492 \\
no2 & 886 & https://www.openml.org/d/886 \\
nwtco & 46165 & https://www.openml.org/d/46165 \\
page & 1021 & https://www.openml.org/d/1021 \\
pakis & 46732 & https://www.openml.org/d/46732 \\
pc1 & 1068 & https://www.openml.org/d/1068 \\
pc2 & 1069 & https://www.openml.org/d/1069 \\
pc3 & 1050 & https://www.openml.org/d/1050 \\
pc4 & 1049 & https://www.openml.org/d/1049 \\
phary & 738 & https://www.openml.org/d/738 \\
place & 509 & https://www.openml.org/d/509 \\
plasm & 915 & https://www.openml.org/d/915 \\
plasm & 511 & https://www.openml.org/d/511 \\
pm10 & 750 & https://www.openml.org/d/750 \\
pm10 & 522 & https://www.openml.org/d/522 \\
polis & 46950 & https://www.openml.org/d/46950 \\
polle & 871 & https://www.openml.org/d/871 \\
prnn & 446 & https://www.openml.org/d/446 \\
profb & 470 & https://www.openml.org/d/470 \\
quake & 772 & https://www.openml.org/d/772 \\
regim & 42172 & https://www.openml.org/d/42172 \\
road & 44453 & https://www.openml.org/d/44453 \\
segme & 958 & https://www.openml.org/d/958 \\
seism & 45562 & https://www.openml.org/d/45562 \\
seoul & 46328 & https://www.openml.org/d/46328 \\
sf & 44773 & https://www.openml.org/d/44773 \\
solar & 40702 & https://www.openml.org/d/40702 \\
spamb & 44 & https://www.openml.org/d/44 \\
spect & 954 & https://www.openml.org/d/954 \\
std & 46156 & https://www.openml.org/d/46156 \\
taiwa & 46962 & https://www.openml.org/d/46962 \\
telco & 42178 & https://www.openml.org/d/42178 \\
tempe & 46731 & https://www.openml.org/d/46731 \\
tic & 50 & https://www.openml.org/d/50 \\
triaz & 788 & https://www.openml.org/d/788 \\
tumor & 46171 & https://www.openml.org/d/46171 \\
vehic & 994 & https://www.openml.org/d/994 \\
vowel & 1016 & https://www.openml.org/d/1016 \\
weath & 42368 & https://www.openml.org/d/42368 \\
wilt & 40983 & https://www.openml.org/d/40983 \\
wisco & 753 & https://www.openml.org/d/753 \\
yacht & 42370 & https://www.openml.org/d/42370 \\
\end{longtable}
\endgroup

\begingroup
\setlength{\parskip}{2pt}
\tcbset{
  before skip=4pt,after skip=4pt,boxsep=0.1mm,
  before upper={\setlength{\parskip}{2pt}\setlength{\parindent}{0pt}}
}
\noindent\begin{minipage}{\linewidth}
\section{Reasoning Example}
In this subsection, we illustrate the reasoning process of TabXLLM. An example of TabXLLM's reasoning process for a classification task is shown below.
\vspace{3pt}
\begin{tcolorbox}[
  enhanced,
  unbreakable,
  colback=FigBlue!8,
  colframe=FigBlueDark,
  colbacktitle=FigBlueDark,
  coltitle=white,
  title={An example of TabXLLM's reasoning process on a classification task},
  fonttitle=\bfseries\small,
  fontupper=\small\setlength{\baselineskip}{10.9325pt},
  boxrule=0.9pt,
  arc=2mm,
  left=1.2mm,right=1.2mm,top=1.5667159pt,bottom=1.5737120pt,
  toptitle=0.6mm,bottomtitle=0.6mm,
  before title={\rule[-3.5pt]{0pt}{12pt}},after title={\rule[-3.5pt]{0pt}{12pt}}
]
\begin{tcolorbox}[
  unbreakable,
  colback=white,
  colframe=FigBlue!70!FigInk,
  colbacktitle=FigBlue!20,
  coltitle=FigBlueDark,
  title={System},
  fonttitle=\bfseries\small,
  fontupper=\small\setlength{\baselineskip}{10.9325pt},
  boxrule=0.5pt,
  arc=1mm,
  left=1mm,right=1mm,top=0.3667348pt,bottom=-0.5825069pt,
  toptitle=0.4mm,bottomtitle=0.4mm,
  before title=\vphantom{System},after title=\vphantom{System},before skip=0.6856201pt,after skip=0.6856201pt
]
\textbf{Role:} You are an expert in tabular prediction. When addressing tabular prediction tasks, you first conduct analysis and reasoning in your mind before presenting the prediction results to the user.
\end{tcolorbox}

\begin{tcolorbox}[
  unbreakable,
  colback=white,
  colframe=FigBlue!70!FigInk,
  colbacktitle=FigBlue!20,
  coltitle=FigBlueDark,
  title={User},
  fonttitle=\bfseries\small,
  fontupper=\small\setlength{\baselineskip}{10.9325pt},
  boxrule=0.5pt,
  arc=1mm,
  left=1mm,right=1mm,top=0.3654701pt,bottom=-0.5375069pt,
  toptitle=0.4mm,bottomtitle=0.4mm,
  before title=\vphantom{System},after title=\vphantom{System},before skip=2.7600514pt,after skip=2.7600514pt
]
\textbf{Input:} The state is 2.0. The account\_length is 110.0. The area\_code is 408.0. The phone\_number is 2398.0.\\
The international\_plan is 0.0. The voice\_mail\_plan is 0.0. The number\_vmail\_messages is 0.0.\\
The total\_day\_minutes is 55.3. The total\_day\_calls is 102.0. The total\_day\_charge is 9.4.\\
The total\_eve\_minutes is 164.7. The total\_eve\_calls is 124.0. The total\_eve\_charge is 14.0.\\
The total\_night\_minutes is 200.7. The total\_night\_calls is 108.0. The total\_night\_charge is 9.03.\\
The total\_intl\_minutes is 10.2. The total\_intl\_calls is 5.0. The total\_intl\_charge is 2.75.\\
The number\_customer\_service\_calls is 1.0.

\vspace{0.9337422pt}
\textbf{Task:} {\color{FigPeachDark}\textbf{Will this customer churn (leave the telecom service)? yes or no?}}

\vspace{0.8257422pt}
\textbf{Output Format:} Show your analysis and reasoning in {\color{FigPurpleDark}\texttt{<think> </think>}} tags. And return the final answer in {\color{FigPurpleDark}\texttt{<answer> </answer>}} tags.
\end{tcolorbox}

\begin{tcolorbox}[
  unbreakable,
  colback=white,
  colframe=FigBlue!70!FigInk,
  colbacktitle=FigBlue!20,
  coltitle=FigBlueDark,
  title={Assistant},
  fonttitle=\bfseries\small,
  fontupper=\small\setlength{\baselineskip}{10.9325pt},
  boxrule=0.5pt,
  arc=1mm,
  left=1mm,right=1mm,top=0.3621477pt,bottom=-0.5870675pt,
  toptitle=0.4mm,bottomtitle=0.4mm,
  before title=\vphantom{System},after title=\vphantom{System},before skip=2.7633637pt,after skip=2.7633637pt
]
\textbf{Reasoning:} {\color{FigPurpleDark}\texttt{<think>}}\\
To determine whether this customer will churn, I examine the profile from several complementary perspectives: service usage, billing pressure, customer support interactions, and long-term retention signals. Churn is usually associated with repeated service issues, unusual cost burden, or very short account age. In this example, most of the evidence points in the opposite direction.\par

\vspace{0.9337422pt}
\textbf{Step-by-step assessment:}
\begin{enumerate}[leftmargin=1.2em,itemsep=2pt,topsep=0.8257301pt,parsep=0pt,partopsep=0pt
]
  \item \textbf{Account longevity.} The account length is 110 months, which is a strong indicator of customer stability and lower churn risk.
  \setlength{\itemsep}{0.9337402pt}\item \textbf{Customer support burden.} The number of customer service calls is only 1, so a major dissatisfaction signal is absent.
  \setlength{\itemsep}{0.9337402pt}\item \textbf{Daytime usage profile.} The customer has regular daytime usage, and the associated charge remains modest.
  \setlength{\itemsep}{0.9337402pt}\item \textbf{Evening and night behavior.} The evening and night usage values are also within a normal range, without signs of unusual cost pressure or disengagement.
  \setlength{\itemsep}{0.9337402pt}\item \textbf{International activity.} The customer does not have an international plan, but the international usage is not excessive, so the related charge is unlikely to create serious billing frustration.
  \setlength{\itemsep}{0.9337402pt}\item \textbf{Voice mail and feature adoption.} The customer does not use a voice mail plan, which appears to reflect preference rather than dissatisfaction.
  \setlength{\itemsep}{0.9337402pt}\item \textbf{Overall cost structure.} Looking across day, evening, night, and international charges together, the total spending pattern appears moderate rather than extreme. Customers facing unusually high bills are more likely to leave, but that pattern is not visible here.
  \setlength{\itemsep}{0.9337402pt}\item \textbf{Absence of classic red flags.} Common churn indicators include repeated support calls, abnormal charges, very short tenure, or unstable usage patterns. This customer does not exhibit any of those major warning signs.
\end{enumerate}

\vspace{0.1080100pt}
\textbf{Synthesis:} The strongest evidence comes from the long account history and the very low number of customer service calls. The usage profile is steady, the charges are not alarming, and there is no obvious mismatch between services used and services available. Overall, this customer appears more likely to stay than to leave.\par

\vspace{0.9337422pt}
\textbf{Conclusion:} Based on the customer's usage patterns, cost structure, service interaction history, and long tenure, the most likely prediction is that this customer will \emph{not} churn. {\color{FigPurpleDark}\texttt{</think>}}

\begin{tcolorbox}[
  colback=white,
  colframe=FigYellow!65!FigInk,
  colbacktitle=FigYellow!55,
  coltitle=FigInk,
  title={Final Answer},
  fonttitle=\bfseries\small,
  fontupper=\small\setlength{\baselineskip}{10.9325pt},
  boxrule=0.45pt,
  arc=1mm,
  left=0.9mm,right=0.9mm,top=0.2598490pt,bottom=0.7139485pt,
  toptitle=0.4mm,bottomtitle=0.4mm,
  before title=\vphantom{System},after title=\vphantom{System},before skip=2.8081867pt,after skip=2.8081867pt
]
{\color{FigPurpleDark}\texttt{<answer>}}no{\color{FigPurpleDark}\texttt{</answer>}}
\end{tcolorbox}
\end{tcolorbox}
\end{tcolorbox}

\end{minipage}
\par
\noindent\begin{minipage}{\linewidth}
An example of TabXLLM's reasoning process for a regression task is shown below.
\vspace{3pt}
\begin{tcolorbox}[
  enhanced,
  unbreakable,
  colback=FigBlue!8,
  colframe=FigBlueDark,
  colbacktitle=FigBlueDark,
  coltitle=white,
  title={An example of TabXLLM's reasoning process on a regression task},
  fonttitle=\bfseries\small,
  fontupper=\small\setlength{\baselineskip}{11.9428pt},
  boxrule=0.9pt,
  arc=2mm,
  left=1.2mm,right=1.2mm,top=2.1357583pt,bottom=2.1431158pt,
  toptitle=0.6mm,bottomtitle=0.6mm,
  before title={\rule[-3.5pt]{0pt}{12pt}},after title={\rule[-3.5pt]{0pt}{12pt}}
]
\begin{tcolorbox}[
  unbreakable,
  colback=white,
  colframe=FigBlue!70!FigInk,
  colbacktitle=FigBlue!20,
  coltitle=FigBlueDark,
  title={System},
  fonttitle=\bfseries\small,
  fontupper=\small\setlength{\baselineskip}{11.9428pt},
  boxrule=0.5pt,
  arc=1mm,
  left=1mm,right=1mm,top=0.9361186pt,bottom=-0.0130930pt,
  toptitle=0.4mm,bottomtitle=0.4mm,
  before title=\vphantom{System},after title=\vphantom{System},before skip=1.6922349pt,after skip=1.6922349pt
]
\textbf{Role:} You are an expert in tabular prediction. When addressing tabular prediction tasks, you first conduct analysis and reasoning in your mind before presenting the prediction results to the user.
\end{tcolorbox}

\begin{tcolorbox}[
  unbreakable,
  colback=white,
  colframe=FigBlue!70!FigInk,
  colbacktitle=FigBlue!20,
  coltitle=FigBlueDark,
  title={User},
  fonttitle=\bfseries\small,
  fontupper=\small\setlength{\baselineskip}{11.9428pt},
  boxrule=0.5pt,
  arc=1mm,
  left=1mm,right=1mm,top=0.9353657pt,bottom=0.0318969pt,
  toptitle=0.4mm,bottomtitle=0.4mm,
  before title=\vphantom{System},after title=\vphantom{System},before skip=3.3289775pt,after skip=3.3289775pt
]
\textbf{Input:} The book\_Title is Schaum's Outline of Computer Architecture. The Author\_Name is Nick Carter.\\
The ratings\_count is 24.0. The Publish\_year is 2001. The Edition is 2. The Pages\_no is 306.\\
The Book\_language is English. The Reviews is 2.

\vspace{0.4928515pt}
\textbf{Task:} {\color{FigPeachDark}\textbf{Based on a computer science book's title, author, ratings count, publication year, edition number, page count, language, and number of reviews, what is its average rating on Goodreads?}}

\vspace{0.3848415pt}
\textbf{Output Format:} Show your analysis and reasoning in {\color{FigPurpleDark}\texttt{<think> </think>}} tags. And return the final answer in {\color{FigPurpleDark}\texttt{<answer> </answer>}} tags.
\end{tcolorbox}

\begin{tcolorbox}[
  unbreakable,
  colback=white,
  colframe=FigBlue!70!FigInk,
  colbacktitle=FigBlue!20,
  coltitle=FigBlueDark,
  title={Assistant},
  fonttitle=\bfseries\small,
  fontupper=\small\setlength{\baselineskip}{11.9428pt},
  boxrule=0.5pt,
  arc=1mm,
  left=1mm,right=1mm,top=0.9307083pt,bottom=-0.0176637pt,
  toptitle=0.4mm,bottomtitle=0.4mm,
  before title=\vphantom{System},after title=\vphantom{System},before skip=3.3335947pt,after skip=3.3335947pt
]
\textbf{Reasoning:} {\color{FigPurpleDark}\texttt{<think>}}\\
To estimate the average Goodreads rating for this book, I consider the title, author information, edition number, publication year, length, and amount of reader feedback. In a regression setting like this, the goal is to infer a plausible value on the rating scale by combining multiple weak signals.\par

\vspace{0.4928415pt}
\textbf{Step-by-step assessment:}
\begin{enumerate}[leftmargin=1.2em,itemsep=2pt,topsep=0.3848596pt,parsep=0pt,partopsep=0pt
]
  \item \textbf{Title semantics.} The title ``Schaum's Outline of Computer Architecture'' strongly suggests that the book is an educational outline or study guide.
  \setlength{\itemsep}{0.4928496pt}\item \textbf{Domain and audience.} Computer architecture is a technical subject, and readers often reward accuracy, structure, and practical usefulness.
  \setlength{\itemsep}{0.4928496pt}\item \textbf{Author signal.} The named author provides some credibility, although by itself the author name is not enough to imply an exceptionally high score. It is more reasonable to view this as a mild positive signal than as decisive evidence for a very high rating.
  \setlength{\itemsep}{0.4928496pt}\item \textbf{Edition number.} The book is in its second edition, which is usually a favorable sign because it suggests the material was useful enough to justify revision.
  \setlength{\itemsep}{0.4928496pt}\item \textbf{Publication year.} The book was published in 2001. For a foundational technical topic, this is not so old that the content would automatically be viewed as obsolete.
  \setlength{\itemsep}{0.4928496pt}\item \textbf{Length and scope.} The page count of 306 suggests moderate coverage: substantial enough to be useful, but not overwhelmingly long.
  \setlength{\itemsep}{0.4928496pt}\item \textbf{Ratings count and reviews.} The book has only 24 ratings and 2 reviews. Low feedback volume means the average score may be noisy, but it still suggests some positive reader engagement.
  \setlength{\itemsep}{0.4928395pt}\item \textbf{Expected score range.} Combining the educational nature of the book, the second edition, moderate length, and limited but nonzero engagement, the most plausible estimate is a rating that is clearly positive but not outstandingly high. In other words, the metadata supports a score above average, but not enough evidence is present to push the estimate confidently into an exceptional range.
\end{enumerate}

\vspace{0.1079900pt}
\textbf{Synthesis:} I would not expect the book to fall near the low end of the scale, because the metadata suggests a legitimate and useful technical resource. The second edition, moderate page count, and technical focus all support a favorable evaluation. At the same time, the small number of ratings and the specialized nature of the subject make an extremely high score less certain. This points to a value in the high 3s rather than the mid 4s.\par

\vspace{0.4928415pt}
\textbf{Conclusion:} Considering the title, domain, edition information, moderate page count, and limited but positive-seeming feedback footprint, a reasonable estimate for the average Goodreads rating is about 3.9. {\color{FigPurpleDark}\texttt{</think>}}

\begin{tcolorbox}[
  colback=white,
  colframe=FigYellow!65!FigInk,
  colbacktitle=FigYellow!55,
  coltitle=FigInk,
  title={Final Answer},
  fonttitle=\bfseries\small,
  fontupper=\small\setlength{\baselineskip}{11.9428pt},
  boxrule=0.45pt,
  arc=1mm,
  left=0.9mm,right=0.9mm,top=0.9771955pt,bottom=1.2833524pt,
  toptitle=0.4mm,bottomtitle=0.4mm,
  before title=\vphantom{System},after title=\vphantom{System},before skip=4.5792138pt,after skip=4.5792138pt
]
Given these considerations, the average rating can be estimated to be around {\color{FigPurpleDark}\texttt{<answer>}}3.9{\color{FigPurpleDark}\texttt{</answer>}}.
\end{tcolorbox}
\end{tcolorbox}
\end{tcolorbox}

\end{minipage}
\par
\endgroup

\clearpage
\section*{NeurIPS Paper Checklist}

\begin{enumerate}

\item {\bf Claims}
    \item[] Question: Do the main claims made in the abstract and introduction accurately reflect the paper's contributions and scope?
    \item[] Answer: \answerYes{}.
    \item[] Justification: The abstract and introduction state the main contribution, PRPO, and the intended scope: strengthening LLMs for tabular prediction through structural-prior RL post-training. The claims are supported by the theoretical analysis in Section~\ref{sec:theory} and the main experiments in Section~\ref{experiment}.

\item {\bf Limitations}
    \item[] Question: Does the paper discuss the limitations of the work performed by the authors?
    \item[] Answer: \answerNo{}.
    \item[] Justification: The paper does not include a dedicated discussion of the limitations of the work.

\item {\bf Theory assumptions and proofs}
    \item[] Question: For each theoretical result, does the paper provide the full set of assumptions and a complete (and correct) proof?
    \item[] Answer: \answerYes{}.
    \item[] Justification: The statements, assumptions, and proofs are given in Section~\ref{sec:theory} and Appendix~\ref{appendix:prpo_proof}. They cover advantage mixing, independent-rollout correct-answer coverage, the clipped surrogate gradient, and conditional stationarity with an explicit gradient-bias term.

    \item {\bf Experimental result reproducibility}
    \item[] Question: Does the paper fully disclose all the information needed to reproduce the main experimental results of the paper to the extent that it affects the main claims and/or conclusions of the paper (regardless of whether the code and data are provided or not)?
    \item[] Answer: \answerYes{}.
    \item[] Justification: The paper describes the dataset construction, train/test settings, evaluation settings, baselines, reward functions, PRPO algorithm, and hyperparameters in Sections~\ref{method}--\ref{experiment} and Appendices~\ref{appendixA}--\ref{appendix:full_quantitative_results}. OpenML dataset identifiers and URLs are listed in Appendix~\ref{appendix:dataset_details}.

\item {\bf Open access to data and code}
    \item[] Question: Does the paper provide open access to the data and code, with sufficient instructions to faithfully reproduce the main experimental results, as described in supplemental material?
    \item[] Answer: \answerYes{}.
    \item[] Justification: The training and inference code, the TabXLLM-8B model weights, and the training, validation, and test data are publicly released; the links are given at the end of the Abstract. OpenML IDs and URLs for all datasets are listed in Appendix~\ref{appendix:dataset_details}.

\item {\bf Experimental setting/details}
    \item[] Question: Does the paper specify all the training and test details (e.g., data splits, hyperparameters, how they were chosen, type of optimizer) necessary to understand the results?
    \item[] Answer: \answerYes{}.
    \item[] Justification: The experimental setup, dataset split protocol, model families, baseline settings, prompts, reward definitions, and PRPO hyperparameters are described in Section~\ref{experiment} and Appendices~\ref{appendixA}, \ref{appendix:text template tabr1}, \ref{appendix:text template qwen deepseek}, \ref{appendix:baseline_settings}, and \ref{appendix:full_quantitative_results}.

\item {\bf Experiment statistical significance}
    \item[] Question: Does the paper report error bars suitably and correctly defined or other appropriate information about the statistical significance of the experiments?
    \item[] Answer: \answerNo{}.
    \item[] Justification: The paper reports aggregate results and complete per-dataset results, but it does not include formal confidence intervals, significance tests, or error bars for the main PRPO experiments due to the high cost of repeated RL post-training runs.

\item {\bf Experiments compute resources}
    \item[] Question: For each experiment, does the paper provide sufficient information on the computer resources (type of compute workers, memory, time of execution) needed to reproduce the experiments?
    \item[] Answer: \answerNo{}.
    \item[] Justification: The paper reports the hardware type and memory used for the main PRPO training runs (8 NVIDIA H100-80G GPUs) and provides training hyperparameters in Appendix~\ref{appendixA}, but it does not provide wall-clock time or total compute for every individual experiment.
    
\item {\bf Code of ethics}
    \item[] Question: Does the research conducted in the paper conform, in every respect, with the NeurIPS Code of Ethics \url{https://neurips.cc/public/EthicsGuidelines}?
    \item[] Answer: \answerYes{}.
    \item[] Justification: The research uses publicly available tabular datasets and open or publicly accessible model baselines, and we are not aware of any deviation from the NeurIPS Code of Ethics.

\item {\bf Broader impacts}
    \item[] Question: Does the paper discuss both potential positive societal impacts and negative societal impacts of the work performed?
    \item[] Answer: \answerNo{}.
    \item[] Justification: The paper does not include a dedicated discussion of potential positive or negative societal impacts of the work. This study focuses on general-purpose tabular prediction without examining a specific deployment scenario.
    
\item {\bf Safeguards}
    \item[] Question: Does the paper describe safeguards that have been put in place for responsible release of data or models that have a high risk for misuse (e.g., pre-trained language models, image generators, or scraped datasets)?
    \item[] Answer: \answerNA{}.
    \item[] Justification: The released assets are a tabular prediction model post-trained from an open-weight base model and data derived from public OpenML datasets. They do not pose a high risk of misuse that would require controlled-access safeguards.

\item {\bf Licenses for existing assets}
    \item[] Question: Are the creators or original owners of assets (e.g., code, data, models), used in the paper, properly credited and are the license and terms of use explicitly mentioned and properly respected?
    \item[] Answer: \answerNo{}.
    \item[] Justification: The paper cites the datasets, models, and software frameworks used, and Appendix~\ref{appendix:dataset_details} provides OpenML identifiers and URLs. However, the paper does not explicitly enumerate the licenses and terms of use for every existing dataset, model, and software asset.

\item {\bf New assets}
    \item[] Question: Are new assets introduced in the paper well documented and is the documentation provided alongside the assets?
    \item[] Answer: \answerYes{}.
    \item[] Justification: We release TabXLLM-8B and the derived training, validation, and test data on Hugging Face (links in the Abstract). Their construction is documented in Section~\ref{experiment} and Appendices~\ref{appendixA}--\ref{appendixB}, and each repository includes a model or dataset card.

\item {\bf Crowdsourcing and research with human subjects}
    \item[] Question: For crowdsourcing experiments and research with human subjects, does the paper include the full text of instructions given to participants and screenshots, if applicable, as well as details about compensation (if any)? 
    \item[] Answer: \answerNA{}.
    \item[] Justification: The paper does not involve crowdsourcing experiments or research with human subjects.

\item {\bf Institutional review board (IRB) approvals or equivalent for research with human subjects}
    \item[] Question: Does the paper describe potential risks incurred by study participants, whether such risks were disclosed to the subjects, and whether Institutional Review Board (IRB) approvals (or an equivalent approval/review based on the requirements of your country or institution) were obtained?
    \item[] Answer: \answerNA{}.
    \item[] Justification: The paper does not involve crowdsourcing experiments or research with human subjects, so IRB approval or equivalent review is not applicable.

\item {\bf Declaration of LLM usage}
    \item[] Question: Does the paper describe the usage of LLMs if it is an important, original, or non-standard component of the core methods in this research? Note that if the LLM is used only for writing, editing, or formatting purposes and does \emph{not} impact the core methodology, scientific rigor, or originality of the research, declaration is not required.
    \item[] Answer: \answerYes{}.
    \item[] Justification: LLMs are a core component of the method and experiments. The paper describes the base model, PRPO post-training procedure, prompt templates, reward design, and evaluated LLM baselines in Sections~\ref{method}--\ref{experiment} and the appendices.

\end{enumerate}

\end{document}